\pdfoutput=1

\documentclass[11pt]{article}

\usepackage[preprint]{acl}

\usepackage{times}
\usepackage{latexsym}

\usepackage[T1]{fontenc}

\usepackage[utf8]{inputenc}

\usepackage{microtype}

\usepackage{inconsolata}

\usepackage{graphicx}

\usepackage{enumitem}
\usepackage{microtype}
\usepackage{hyperref}
\usepackage{url}
\usepackage{booktabs}
\definecolor{darkblue}{rgb}{0, 0, 0.5}
\hypersetup{colorlinks=true, citecolor=darkblue, linkcolor=darkblue, urlcolor=darkblue}

\usepackage{hyperref}
\usepackage{url}
\usepackage{xspace}
\usepackage{wrapfig}
\usepackage{float}
\usepackage{natbib}
\usepackage{xcolor}
\usepackage{amsmath}
\usepackage{amsfonts}
\usepackage{algorithm}
\usepackage{algorithmicx}
\usepackage{algpseudocode}
\usepackage{spverbatim}
\usepackage{soul}
\usepackage{tablefootnote}
\usepackage{multirow}
\usepackage[textsize=scriptsize]{todonotes}
\usepackage{colortbl}
\usepackage{yfonts}
\usepackage{amssymb}
\usepackage{wasysym}
\usepackage{longtable}
\usepackage{todonotes}
\usepackage{makecell}
\usepackage{subcaption}
\usepackage{tablefootnote}

\usepackage{tabularx} 
\usepackage{fontawesome}
\usepackage{array}    

\definecolor{darkgreen}{rgb}{0.0, 0.5, 0.13} 
\definecolor{gold}{rgb}{0.83, 0.69, 0.52}

\definecolor{lightred}{HTML}{e99090}

\newcommand{\name}{\textsc{\small  NoCha}}

\newcommand{\claude}{\textsc{\small Claude-3-Opus}}
\newcommand{\sonnet}{\textsc{\small Claude-3.5-Sonnet}}

\newcommand{\gptturbo}{\textsc{\small GPT-4-Turbo}}
\newcommand{\gpto}{\textsc{\small GPT-4o}}
\newcommand{\geminipro}{\textsc{\small Gemini Pro 1.5}}
\newcommand{\geminiflash}{\textsc{\small Gemini Flash 1.5}}
\newcommand{\comrplus}{\textsc{\small Command R+}}
\newcommand{\comr}{\textsc{\small Command R}}
\newcommand{\phimodel}{\textsc{\small Phi-3-mini}}
\newcommand{\gemma}{\textsc{\small Gemma-10M}}
\newcommand{\longllama}{\textsc{\small LongLLaMA}}
\newcommand{\bm}{\textsc{\small BM25+GPT-4o}}
\newcommand{\mixtral}{\textsc{\small Mixtral-8x22B}}
\newcommand{\qwen}{\textsc{\small Qwen-2-72B}}

\newcommand\blankfootnote[1]{%
  \let\thefootnote\relax\footnotetext{#1}%
  \let\thefootnote\svthefootnote%
}

\title{\includegraphics[height=1em]{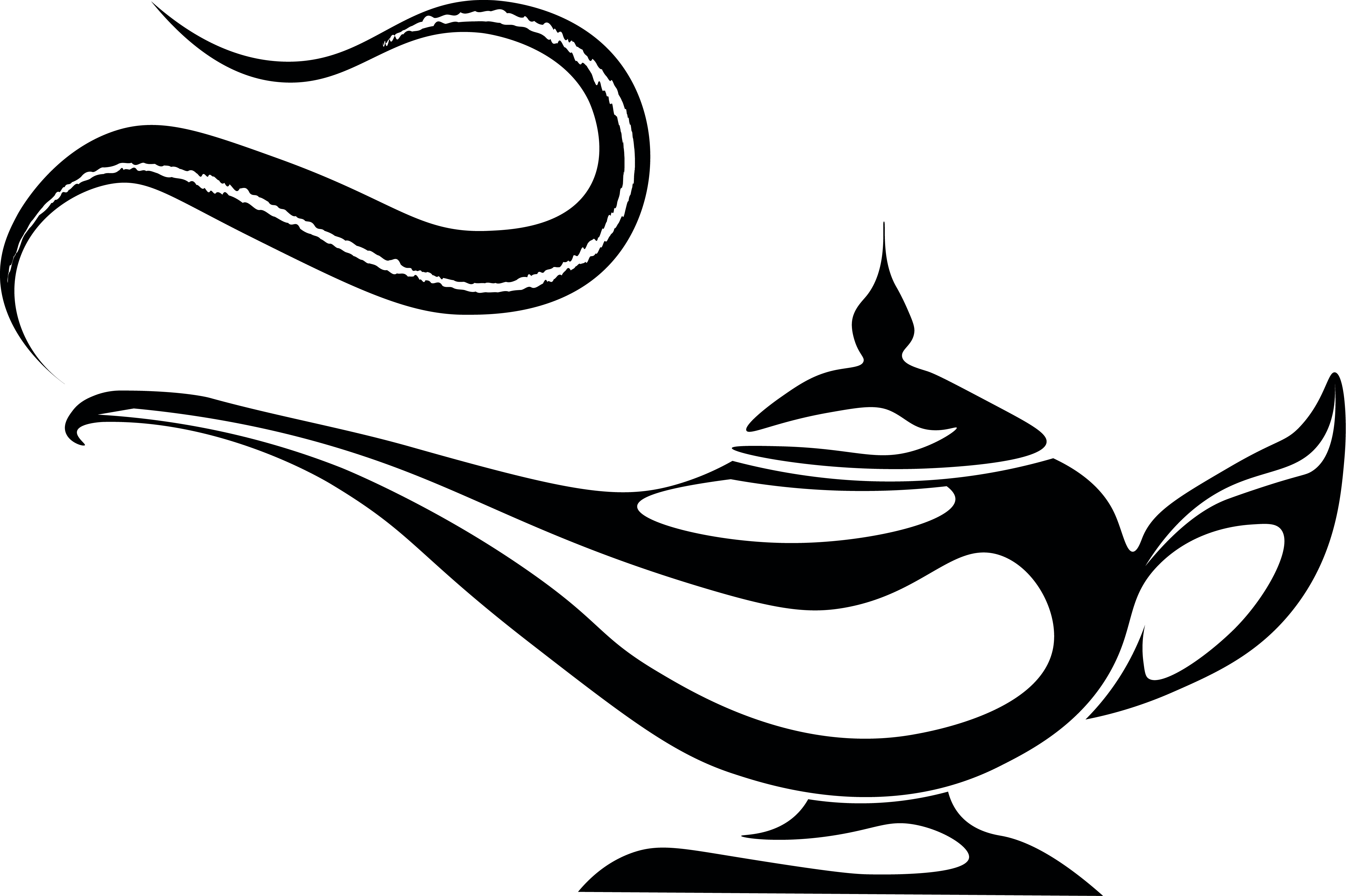} \emph{One Thousand and One Pairs}:\\ A ``novel'' challenge for long-context language models
\\  
    }

\author{Marzena Karpinska\textsuperscript{\faMoonO} \quad
    Katherine Thai\textsuperscript{\faMoonO} \quad
    Kyle Lo\textsuperscript{\faMagic} \quad
    Tanya Goyal\textsuperscript{\faStarO}  \quad 
    Mohit Iyyer\textsuperscript{\faMoonO} \\
    \textsuperscript{\faMoonO}UMass Amherst \quad
    \textsuperscript{\faMagic}Allen Institute for AI \quad
    \textsuperscript{\faStarO}Cornell University\\
    \small{\texttt{\{mkarpinska,kbthai,miyyer\}@umass.edu} \quad 
    \texttt{kylel@allenai.org} \quad \texttt{tanyagoyal@cornell.edu}}
}

\begin{document}

\maketitle

\begin{abstract}

Synthetic long-context LLM benchmarks (e.g., ``needle-in-the-haystack'') test only surface-level retrieval capabilities, but how well can long-context LLMs retrieve, synthesize, and reason over information across book-length inputs? We address this question by creating \name, a dataset of \emph{1,001} \textit{minimally different} pairs 
of true and false claims about \emph{67} recently-published English fictional books, written by human readers of those books. In contrast to existing long-context benchmarks, our annotators confirm that the largest share of pairs in \name\ require \emph{global reasoning over the entire book} to verify. 
Our experiments show that while human readers easily perform this task, it is enormously challenging for all ten long-context LLMs that we evaluate: no open-weight model performs above random chance (despite their strong performance on synthetic benchmarks), while \gpto\ achieves the highest accuracy at 55.8\%. Further analysis reveals that (1) on average, models perform much better on pairs that require only sentence-level retrieval vs. global reasoning; (2) model-generated explanations for their decisions are often inaccurate even for correctly-labeled claims; and (3) models perform substantially worse on speculative fiction books that contain extensive world-building. The methodology proposed in \name\ allows for the evolution of the benchmark dataset and the easy analysis of future models.

\end{abstract}

\section{Introduction}
\label{sec:intro}

The context size of large language models has increased by multiple orders of magnitude over the last year: for instance, Google's \geminipro{} can process millions of input tokens at once. But can models \textit{truly} utilize and reason over their claimed context? 
Existing long-context evaluation methods such as finding the "needle in the haystack" (NIAH) \citep{kamradt2023needle} measure surface-level retrieval capabilities, but do not necessarily assess performance on the more challenging task of synthesizing distant and underlying information as we show in \S\ref{sec:analysis}.

We bridge this gap by introducing \name\ (A \textbf{No}vel \textbf{Cha}llenge), in which LLMs are prompted to verify claims written about recently-published fiction books. Claim verification has been extensively studied in the context of shorter documents \citep{thorne-etal-2018-fever, wadden-etal-2020-fact, fabbri-etal-2022-qafacteval}, but its application to book-length fictional texts presents unique challenges. Firstly, the task necessitates reasoning over both explicit information directly stated in the text and implicit information inferred from the narrative, which is often distributed throughout the entire document. 
Secondly, the use of recent, fictional context prevents the model from relying solely on parametric knowledge, necessitating the comprehension and interpretation of the long text.

\begin{figure*}[tbp]
  \includegraphics[width=1\linewidth]{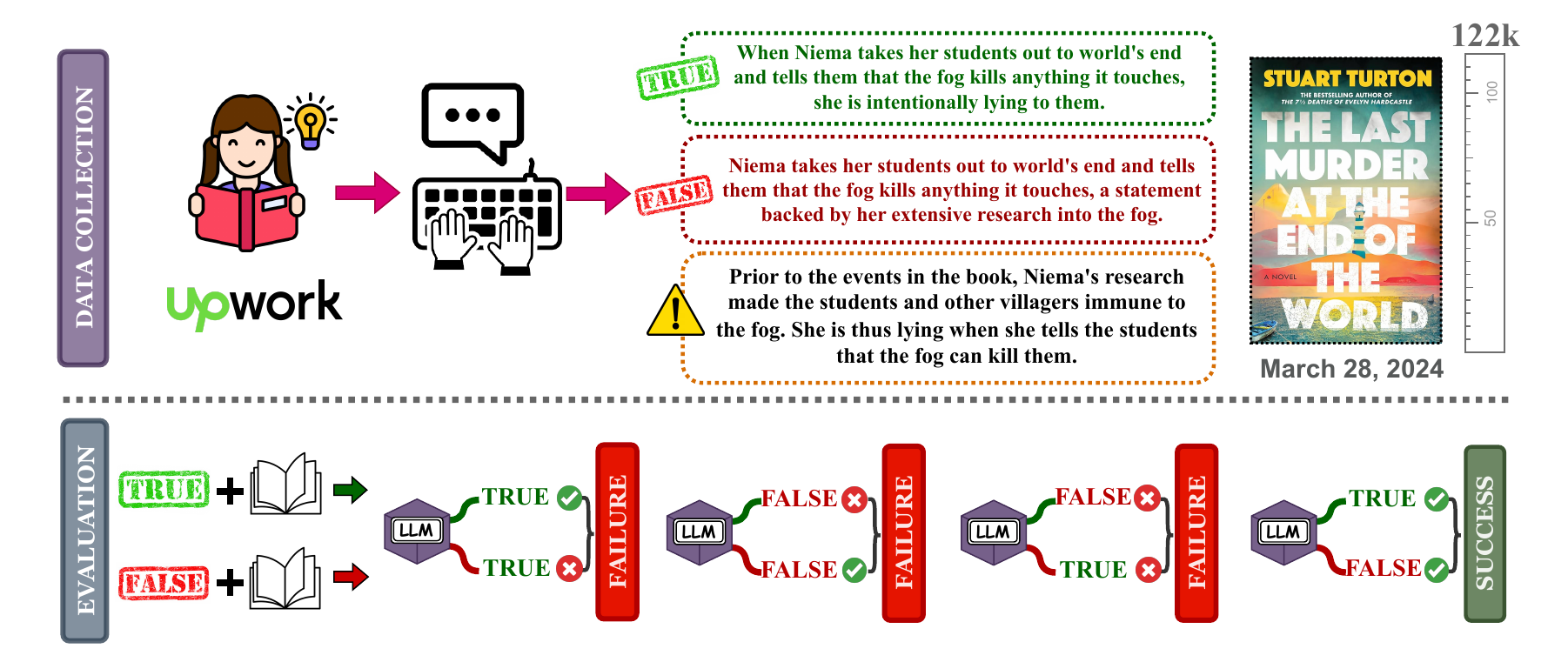} 
  \caption{Overview of \name's data collection and evaluation pipeline. Readers create true/false claim pairs for books published between 2023 and 2024 with written justifications for the false labels. Each model is given the full book as context and evaluates one claim at a time. We measure \emph{pair accuracy}, where the model must identify the true claim as true and the false claim as false to receive credit. This approach helps guard against label bias while also better assessing the true understanding of the text, as both claims pertain to the same events or parts of the story. Our books range from 49k to 336k tokens, and each model is tested on a subset of books that fit within its context.}
  \label{fig:datacollection}
\end{figure*}

Our data collection process aims to balance efficiency and quality. Rather than pre-selecting a set of books, we follow the approach of \citet{kim2024fables} and ask annotators to self-report recently published novels they have read. The annotators then create true/false \textbf{narrative minimal pairs} that isolate a single narrative phenomenon present in their novels. Each false claim differs from the true claim in its pair \textit{only} by the inclusion of false information regarding the same event or entity (see \autoref{fig:datacollection}). This approach offers two key advantages: (1) it minimizes the chances that the model is ``correct for the wrong reason,'' as it must accurately predict \textit{both} labels in the pair, 
and (2) it simplifies the process of quality control, as the false claim can be easily verified against its true counterpart, making it easier to identify claims that are either too similar or overly subjective. \name\  contains 1,001 narrative minimal pairs for 67 books, created at a total cost of \$3,330 USD.

Accurately labeling the minimal pairs in \name\ often requires not only information retrieval from the context, but also synthesis, reasoning, and inference over information from multiple parts of the document. 
Experiments on 5 openly-available and 6 closed-source models show that: (1) though all models struggle on \name, no open-weight model performs above random chance, while \gpto\ is the best-performing model overall at 55.8\% accuracy; (2) on average, all models perform much better on pairs that require sentence-level retrieval than global reasoning (59.8\% vs 41.6\%), though their  performance on these pairs in \name\ is much lower (59.8\%) than on NIAH (100\%) as reported in \citet{sun2024ruler}; (3) model-generated explanations for their decisions are often inaccurate, even for correctly labeled claims; and (4) models perform substantially worse on books with extensive world-building, as these require more sophisticated reasoning.

To summarize, our contributions are threefold:
\begin{enumerate}[leftmargin=*]
    \item \faFileTextO\ \textbf{\textsc{Data:}} We introduce \name, a dataset of 1,001 narrative minimal pairs about \textit{recently} published fictional books, designed to evaluate the long-context language models' ability to process and reason about long narratives.
    \item \faSearch\ \textbf{\textsc{Analysis:}} We use \name\ to conduct a comprehensive analysis of 5 openly available and 6 closed-source models, identifying where models struggle, thereby providing valuable insights improving long-context models.\footnote{In this paper, we define long-context models as those with a claimed context window of at least 128k tokens.}
    \item \faGraduationCap\ \textbf{\textsc{Methodology:}} We contribute our data collection and evaluation methodology, which balances efficiency and quality, minimizing the possibility of models receiving credit for correct predictions made without full utilization of the context.
\end{enumerate}

\section{Data \& Methods}
\label{sec:data_methods}
In this section, we first motivate \name's design---narrative minimal pairs written by readers of books---and then describe our data collection process and evaluation methodology.

\paragraph{Issues with existing long-context benchmarks:} 
A popular method to evaluate long-context models is the ``needle-in-a-haystack'' task \citep[][NIAH]{kamradt2023needle}, which involves injecting a sentence-level piece of information at different depths of a document. NIAH was recently extended by \textsc{Ruler}~\citep{sun2024ruler} to include various types of needles and other synthetic tasks. While this approach allows control over the exact position of the evidence, it has several limitations: (1) evidence scope is heavily restricted by the length and complexity of the needle, (2) the needle is unrelated to the continuous context, making it easier to retrieve,\footnote{The context in \citet{sun2024ruler} may match the needle's topic, but it is not a continuous, logically connected text.} and (3) the task is synthetic, making it a poor proxy for real-world tasks. 
Recently, more realistic long-context benchmarks have emerged. The most similar to \name\ are \textsc{L-Eval} \citep{chenxin2024leval}, \textsc{NovelQA} \citep{wang2024novelqa}, and $\infty$\textit{bench} \citep{zhang2024inftybench}, which are all literary QA tasks. However, all three use novels that are publicly available online, meaning they are likely part of LLMs' pretraining data along with the numerous books, articles, and summaries written about them.\footnote{While $\infty$\textit{bench} swaps core entities in the texts, we found that prompting \gptturbo{} with excerpts from the "falsified" novels results in the model recognizing the original novel and identifying the changed entities.} They also lack the complex global coverage of many of our claims; for instance, while 35\% of \textsc{NovelQA}  reportedly consists of multi-hop questions, these involve simpler tasks such as aggregation (e.g., \textit{"How many times has Bran jumped off and ran?"}) that do not require reasoning over \textit{implicit} information in the text.

\subsection{Data collection}
\label{subsec:data_collection}

To address these limitations, we adopt a design that includes (1) human-written examples for claim verification, an important real-world task; (2) recently-published texts to mitigate data contamination issues \cite{jacovi-etal-2023-stop}; (3) fictional texts to prevent over-reliance on parametric knowledge vs. in-context information \citep{sun2024ruler}; and (4) minimal pairs to guard against models being ``right for the wrong reasons'' (see \autoref{fig:pair_failure} for example) while also enabling easy verification to ensure dataset quality.



\paragraph{Collecting a  corpus of recently-published fiction:} 
We collect the 67 books in \name\ following the approach of~\citet{kim2024fables}, who created the FABLES dataset to evaluate summarization faithfulness with similar objectives in mind (reducing data contamination and reliance solely on parametric knowledge). Concretely, we select books that are: (1) fictional narratives, (2) published in 2023 or 2024,\footnote{We will not release the full \name\ dataset because (1) the books are copyrighted, and (2) we want to prevent model providers from training on this data and compromising the benchmark. Instead, we additionally annotate a smaller subset of \textit{classic books} that are out of copyright. This subset will serve as a publicly available sample of the data. The authors commit to updating \name\ with new books and evaluating new long-context models.} and (3) self-reported as having been read by our annotators.
Our resulting dataset comprises 63 new books (33 published in 2023 and 30 in 2024) and four classic novels (see \autoref{tab:list_of_novels} for full list of the books). The mean length of books in our dataset is \textbf{127k} tokens and \textbf{98.5k} words (see \autoref{tab:basic_summary_stats_dataset} for statistics).\footnote{All token counts provided in this paper are based on \texttt{tiktoken} (\url{https://github.com/openai/tiktoken}) with the \texttt{cl100k} encoding and word counts are determined by splitting the text on \texttt{whitespaces}, unless stated otherwise.}

\paragraph{Annotators:} We recruited 18 annotators via \href{https://www.upwork.com/}{Upwork} and 5 volunteer annotators (convenience sample) who reported having read books published in 2023 or 2024.\footnote{The volunteer group included three of the authors, each of whom read three books of their choice and performed the annotations to better understand the difficulty of the task and the time required to complete it.} All annotators were required to read the guidelines and sign a consent form before starting the task. They were compensated \$3--\$5 USD per pair of claims, based on the agreed rate and the number of annotations, resulting in an hourly rate of approximately \$18--\$30 USD.\footnote{The annotators were usually able to create about 6-10 claim pairs per hour.} Overall, we collected 1001 claim pairs (approximately 10-15 per book) at a cost of \$2.8k USD. The protocol was reviewed and deemed \textit{Not Human Subjects Research} by the Institutional Review Board. See \S\ref{app_sec:claim_collection} for details on the annotators and the recruitment process.

\begin{table}[t!]
\centering
\small
\resizebox{0.48\textwidth}{!}{%
\begin{tabular}{lcc|cc|c}
\toprule
 & \multicolumn{2}{c}{\textbf{Books } \includegraphics[height=1.1em]{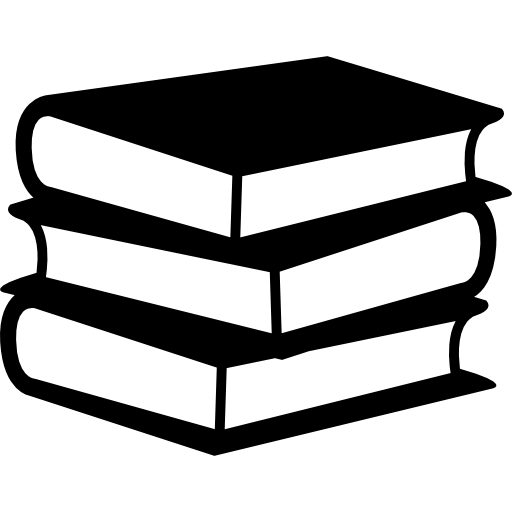}} & \multicolumn{3}{c}{\textbf{Claim\textsubscript{\{\textit{Pairs}\}} } \includegraphics[height=1em]{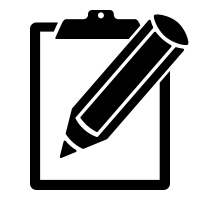}} \\
 & \multicolumn{2}{c}{(\textit{n=67})} & \multicolumn{3}{c}{(\textit{n=2002})\textsubscript{\{\textit{1001}\}}} \\
\cmidrule(lr){2-3} \cmidrule(lr){4-6}
 & \textsc{Tokens} & \textsc{Words} & \textsc{Tokens} & \textsc{Words} & \textsc{\# Claim/Book}\\
\midrule
\textsc{Mean} & 127,324 & 98,587 & 23.22 & 18.26 & 14.94 \\
\textsc{St. Dev.} & 52,561 & 39,506 & 7.62 & 6.49 & 8.37 \\
\textsc{Max} & 336,288 & 257,445 & 63 & 57 & 46 \\
\textsc{Min} & 49,156 & 38,023 & 5 & 4  & 4 \\
\bottomrule
\end{tabular}
}
\caption{Summary statistics for \name. }
\label{tab:basic_summary_stats_dataset}
\end{table}

\paragraph{Collecting true/false pairs:} To collect true/false pairs, annotators were instructed to first write a true statement about the events or characters in the book, and then create a corresponding false statement addressing the same aspect of the book such that the verification of one claim as "true" necessitates the verification of the other as "false". We trained annotators by having them write a true counterpart to a false model-generated claim.\footnote{Model-generated claims were selected from FABLES~\cite{kim2024fables} as well as generated by  \claude{} and \gptturbo{}.} Both types of pairs—those written from scratch and those where only one claim was created by the annotator—were included in the final dataset after quality control.\footnote{Around 30\% of the pairs in the data contain one model generated claim. In some cases, these claims were edited in order to decontextualize the claim.} The annotators were also asked to write a short explanation as to why these claims were true or false based on the content of the book. We closely monitored this process, providing feedback and seeking clarifications from the annotators.

\paragraph{Quality control:} To ensure the quality of the claims, we hired annotators who had read the same books to validate 152 claims from six books, one new annotator per book. The annotators were paid \$1.66--\$2.77 USD per claim based on their requested rate, with the total annotation costing \$285 USD. Overall, the annotators agreed on 148 out of 152 labels (Krippendorff's $\alpha=0.960$) (see \S\ref{app_sec:claim_collection} for details on this process). This high agreement also allow us to conclude that \textbf{human readers are very strong performers on \name's claim verification task (97.4\%).}
Finally, after collecting all pairs, two of the authors reviewed all instances, resolving unclear cases with the annotators and between themselves if a claim appeared to be subjective or incorrect.

\paragraph{Evidence scope:} To assess the reasoning required to validate a claim pair, we obtained additional annotations on 121 claim pairs from 8 books. These annotations determine whether verification is possible based on (1) one or two sentences (similar to NIAH), (2) a single contiguous passage, or (3) global reasoning over the full book. Annotators were compensated \$2 USD per pair, totaling \$242 USD for the annotations. Overall, our annotations suggest that 12.4\% of pairs can be validated with reasoning over one or two sentences, 39.7\% require reasoning over a longer contiguous passage, and \textbf{the largest fraction, 47.9\%, necessitate global reasoning for correct verification.}

\section{Experiments}
\label{sec:experiments}

\begin{table}[t]
\centering
\footnotesize
\resizebox{0.48\textwidth}{!}{%
\begin{tabular}{l c l c  c}
\toprule
\textsc{Model} & \textsc{Context} & \textsc{Avail.} & \textsc{Checkpoints}& \textsc{\# Param}  \\
\midrule
\href{https://platform.openai.com/docs/models}{\gpto} & 128k & \faLock & \texttt{gpt-4o-2024-05-13} & \includegraphics[height=1.1em]{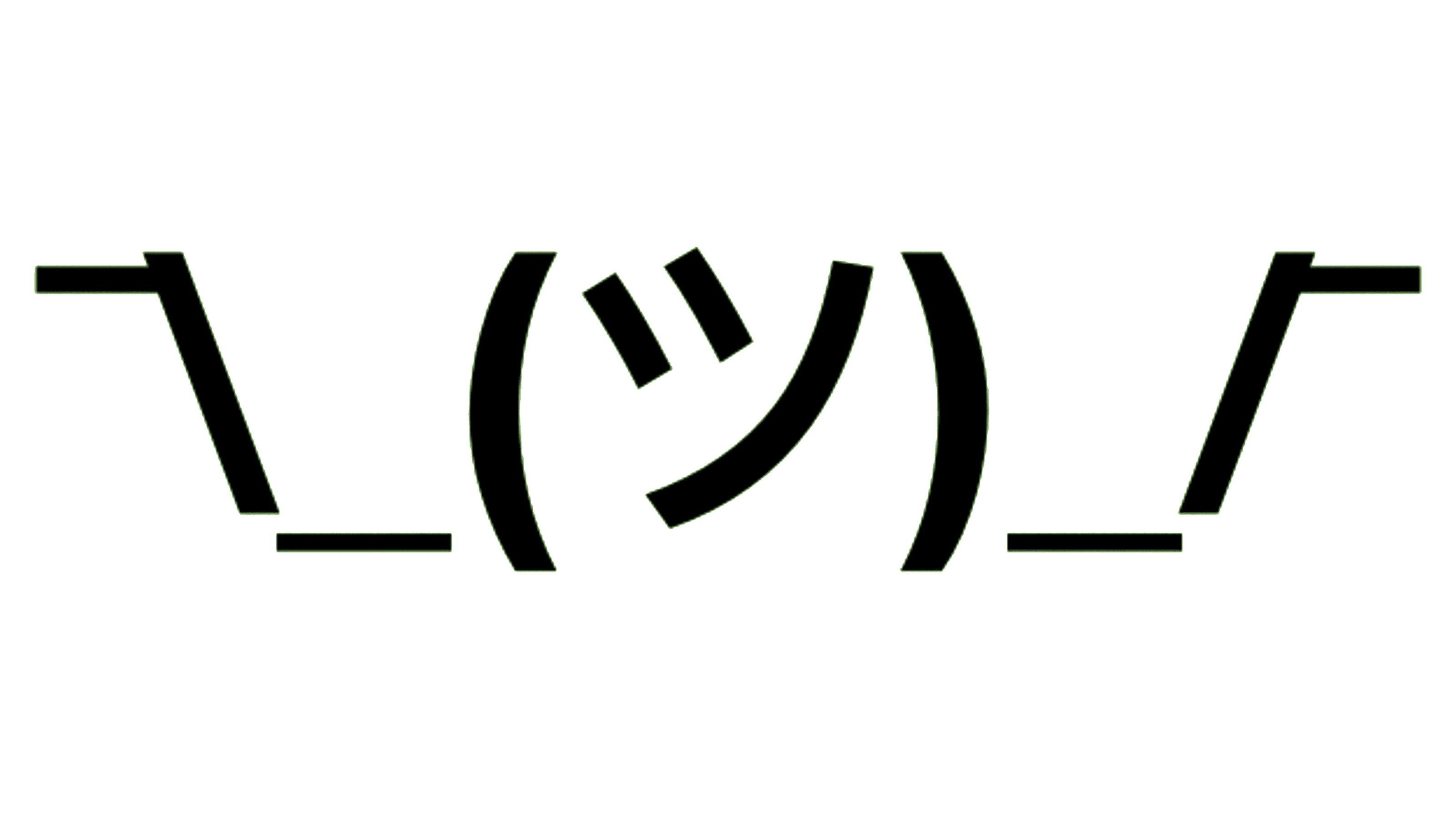} \\
\href{https://platform.openai.com/docs/models}{\gptturbo} & 128k & \faLock & \texttt{gpt-4-turbo-2024-04-09} & \includegraphics[height=1.1em]{figures/shrug_kaomoji.png} \\
\href{https://www.anthropic.com/news/claude-3-family}{\claude} & 200k & \faLock & \texttt{claude-3-opus-20240229} & \includegraphics[height=1.1em]{figures/shrug_kaomoji.png} \\
\href{https://www.anthropic.com/news/claude-3-5-sonnet}{\sonnet} & 200k & \faLock & \texttt{claude-3-5-sonnet@20240620} & \includegraphics[height=1.1em]{figures/shrug_kaomoji.png} \\
\href{https://cloud.google.com/vertex-ai/generative-ai/docs/model-reference/gemini}{\geminipro} & 1M & \faLock & \texttt{gemini-1.5-pro-preview-0514} & \includegraphics[height=1.1em]{figures/shrug_kaomoji.png}  \\
\href{https://cloud.google.com/vertex-ai/generative-ai/docs/model-reference/gemini}{\geminiflash} & 1M & \faLock & \texttt{gemini-1.5-flash-preview-0514} & \includegraphics[height=1.1em]{figures/shrug_kaomoji.png}  \\
\midrule
\href{https://huggingface.co/CohereForAI/c4ai-command-r}{\comr} & 128k & \faUnlock & \texttt{c4ai-command-r-v01} &35B  \\
\href{https://huggingface.co/CohereForAI/c4ai-command-r-plus}{\comrplus} & 128k & \faUnlock & \texttt{c4ai-command-r-plus} & 104B  \\
\href{https://huggingface.co/mustafaaljadery/gemma-2B-10M}{\gemma{}} & 10M & \faUnlock & \texttt{gemma-2b-10m} &2B \\
\href{https://huggingface.co/microsoft/Phi-3-mini-128k-instruct}{\phimodel{}} & 128k & \faUnlock & \texttt{Phi-3-mini-128k-instruct} &3.8B  \\
\href{https://huggingface.co/syzymon/long_llama_3b_instruct}{\longllama{}} & 256k & \faUnlock & \texttt{long\_llama\_3b\_instruct} &3B   \\
\bottomrule
\end{tabular}
}
\caption{Evaluated models: the upper row displays all closed-source models, while the lower row lists all open-weight models (see \S\ref{app_sec:methods_info} for details).}
\label{tab:models}
\end{table}

We test the long context reasoning capabilities of 5 open-weight models and 6 closed-source models (see \autoref{tab:models}) by prompting each model in a zero-shot manner to verify a single claim given the entire book content as evidence, similar to the summarization setup of \citet{kim2024fables}.
After pilot experiments with different prompts (see \S\ref{app_sec:methods_info} for a full description), we observe that the best-performing prompt asks the model to first explain its reasoning before making a final decision, similar to chain-of-thought prompting~\citep{wei2022chain}.

\paragraph{Model and prompt details:}
Each model was prompted with a claim and the entire book using the prompt template in \autoref{tab:prompt_template_claim_eval}.\footnote{We set the temperature to 0 and generate up to 800 tokens.} All \faLock\ \textbf{closed-source models} were accessed via the provider's API\footnote{We encounter several issues while generating the text. Most notably, \geminipro{} and \geminiflash{} refuse to process some of the books returning a \texttt{prohibited content} error (likely due to copyrights). Overall, both models refused to generate the label in about 48\% of cases significantly reducing the number of pairs tested for these models.} at a total cost of $\sim$\$8k USD.  Since we found that \faUnlock\ \textbf{open-weight models} struggle to follow the prompt's instructions, we also experiment a simplified prompt that just asked for a true or false decision (see \autoref{tab:prompt_template_claim_eval_simple}).\footnote{Outputs generated using the simplified prompt are denoted by the subscript \textit{simple}.}\footnote{We encountered some generation issues with \longllama{} which failed to produce outputs when prompted with the prompt in  \autoref{tab:prompt_template_claim_eval}. Hence, we only report the results with the simplified prompt for this model.} Finally, to measure whether retrieval-augmented language models perform better or worse than the long-context setting, we also  experiment with a \faDatabase~\textbf{\bm{}} configuration, in which  {\textsc{\small BM25}} \citep{Robertson1995Okapi} is used to rank the most relevant $k \in \{5,25,50\}$ excerpts from the book.\footnote{Excerpts were an average of 285 words based on \texttt{whitespace} splitting with all paragraph breaks preserved.} We then prompt \gpto{} (our best performing model) with the retrieved excerpts as evidence using the prompt in \autoref{tab:prompt_template_claim_eval_bm25}.\footnote{50 excerpts equals at most 39\% of the total text of any book in our dataset; additional statistics on the total percentage of a book retrieved using this approach are available in \autoref{fig:percentage_retrieved_bm25}.  The total cost of this experiment was \$330 USD.}

\paragraph{Evaluation:} We report the overall \textsc{Pairwise Accuracy} for each model. Models get credit only if they label \textit{both} the true and false claim in a pair correctly; and no points otherwise. We report the number of correctly-verified pairs divided by the total number of pairs that the model labeled. See \S\ref{app_sec:methods_info} for details.

\section{Results \& analysis}
\label{sec:analysis}

\begin{table}[t]
\centering
\resizebox{0.48\textwidth}{!}{%
\begin{tabular}{lcc}
\toprule
\textsc{Model} & \textsc{\faLink\  Pair ACC}\textsubscript{(correct/total)} & \textsc{Common Set ACC} \\
\midrule
\gpto & \textbf{55.8} \textsubscript{(344/617)} & \textbf{58.2} \textsubscript{(206/354)} \\
\gptturbo & 40.2 \textsubscript{(248/617)} & 40.1 \textsubscript{(142/354)} \\
\claude & 49.4 \textsubscript{(463/937)} & 50.8 \textsubscript{(180/354)} \\
\sonnet & 41.0 \textsubscript{(384/937)} & 40.7 \textsubscript{(144/354)} \\
\geminipro & 48.1 \textsubscript{(247/514)} & 48.3 \textsubscript{(171/354)} \\
\geminiflash & 34.2 \textsubscript{(176/515)} & 35.0 \textsubscript{(124/354)}  \\
\midrule
\comr & {\color{purple}19.6} \textsubscript{(87/445)} & \textit{n/a} \\
\comr{}\textsubscript{simple} & {\color{purple}22.5} \textsubscript{(100/445)} & \textit{n/a} \\
\comrplus & {\color{purple}17.3} \textsubscript{(77/445)} & \textit{n/a} \\
\comrplus{}\textsubscript{simple} & {\color{purple}13.7} \textsubscript{(61/445)} & \textit{n/a} \\
\phimodel{} & {\color{purple}9.3} \textsubscript{(23/247)} & \textit{n/a} \\
\phimodel{}\textsubscript{simple} & {\color{purple}14.5} \textsubscript{(48/331)} & \textit{n/a} \\
\gemma{} & {\color{purple}3.9} \textsubscript{(39/1001)} & \textit{n/a} \\
\gemma{}\textsubscript{simple} & {\color{purple}7.5} \textsubscript{(75/1001)} & \textit{n/a} \\
\longllama{}\textsubscript{simple} & {\color{purple}4.9} \textsubscript{(61/937)} & \textit{n/a} \\
\midrule
\faDatabase\ \bm{} (\textit{k=5}) & 28.2 \textsubscript{(282/1001)} & 28.9 \textsubscript{(102/353)} \\
\faDatabase\ \bm{} (\textit{k=25}) & 44.1 \textsubscript{(441/1001)} & 46.5 \textsubscript{(164/353)} \\
\faDatabase\ \bm{} (\textit{k=50}) & 49.7 \textsubscript{(497/1001)} & 51.0 
 \textsubscript{(180/353)} \\
\midrule
\textsc{Random} & 25.0 \textsubscript{(250/1001)} & 25.0 \textsubscript{(88/353)} \\
\bottomrule
\end{tabular}
}
\caption{Percentage of claim pairs identified correctly by each model (see \autoref{tab:prompt_template_claim_eval} for the prompt; and \autoref{tab:prompt_template_claim_eval_bm25} for the prompt employed with {\textsc{\small BM25}}). ``\textsc{Common set}'' refers to claim pairs shared among the models. The subscript ``\textsc{simple}'' refers to the calls done with simplified prompt (\autoref{tab:prompt_template_claim_eval_simple}). Models performing below random are marked in {\color{purple}red}. 
This results include also four classic novels, which were most likely in the training data. For results excluding these novels see \autoref{tab:pair_accuracy_models_no_classics}.
}
\label{tab:pair_accuracy_models_with_classics}
\end{table}

\begin{table}[t]
\centering
\resizebox{0.48\textwidth}{!}{%
\begin{tabular}{lccc}
\toprule
\textsc{Model} & \makecell{\textsc{Ruler (\%)} \\ \textsc{Vanilla NIAH}} & \makecell{\textsc{Ruler (\%)} \\ \textsc{NIAH Suite}}  & \textsc{\name\ (\%)} \\
\midrule
\gptturbo{} & 100.0 & 89.6 & 40.2 \\
\comr{} & 98.0 & 84.8 & 19.6 / 22.5\textsubscript{simple} \\
\bottomrule
\end{tabular}
}
\caption{Performance on NIAH does not translate to \name\ accuracy for \gptturbo{} and \comr{}. The table includes published results with 128k tokens from~\citet{sun2024ruler} for both the standard NIAH (Table 11) and the NIAH suite, which averages results from 8 NIAH variants (Table 13). Additionally, we provide a comparison of performance on the entire \textsc{Ruler} benchmark versus \name\ for overlapping models in \autoref{tab:ruler_full_vs_1001} for reference.}
\label{tab:ruler_vs_1001}
\end{table}

\begin{table}[t]
\centering
\footnotesize
\resizebox{0.4\textwidth}{!}{%
\begin{tabular}{lcc}
\toprule
\textsc{Model} &  \makecell{\textsc{Ruler (\%)} \\ (128k)} &  \makecell{\textsc{\name\ (\%)} \\ ($\sim$128k)} \\
\midrule
\geminipro{} & 94.4 & 48.1 / 48.3\textsubscript{($\sim$128k)} \\
\gptturbo{} & 81.2 & 40.2 \\
\comr{} & 76.0 & 19.6 / 22.5\textsubscript{simple} \\
\comrplus{} & 63.1 & 17.3 / 13.7\textsubscript{simple} \\
\phimodel{} & 43.3 & 9.3 / 14.5\textsubscript{simple} \\
\bottomrule
\end{tabular}
}
\caption{Performance on \textsc{Ruler} \cite{sun2024ruler} compared to \name\ accuracy for overlapping models. The table includes results for the 128k tokens group from \url{https://github.com/hsiehjackson/RULER}.}
\label{tab:ruler_full_vs_1001}
\end{table}

\begin{figure*}[t]
\centering
  \includegraphics[width=0.9\linewidth]{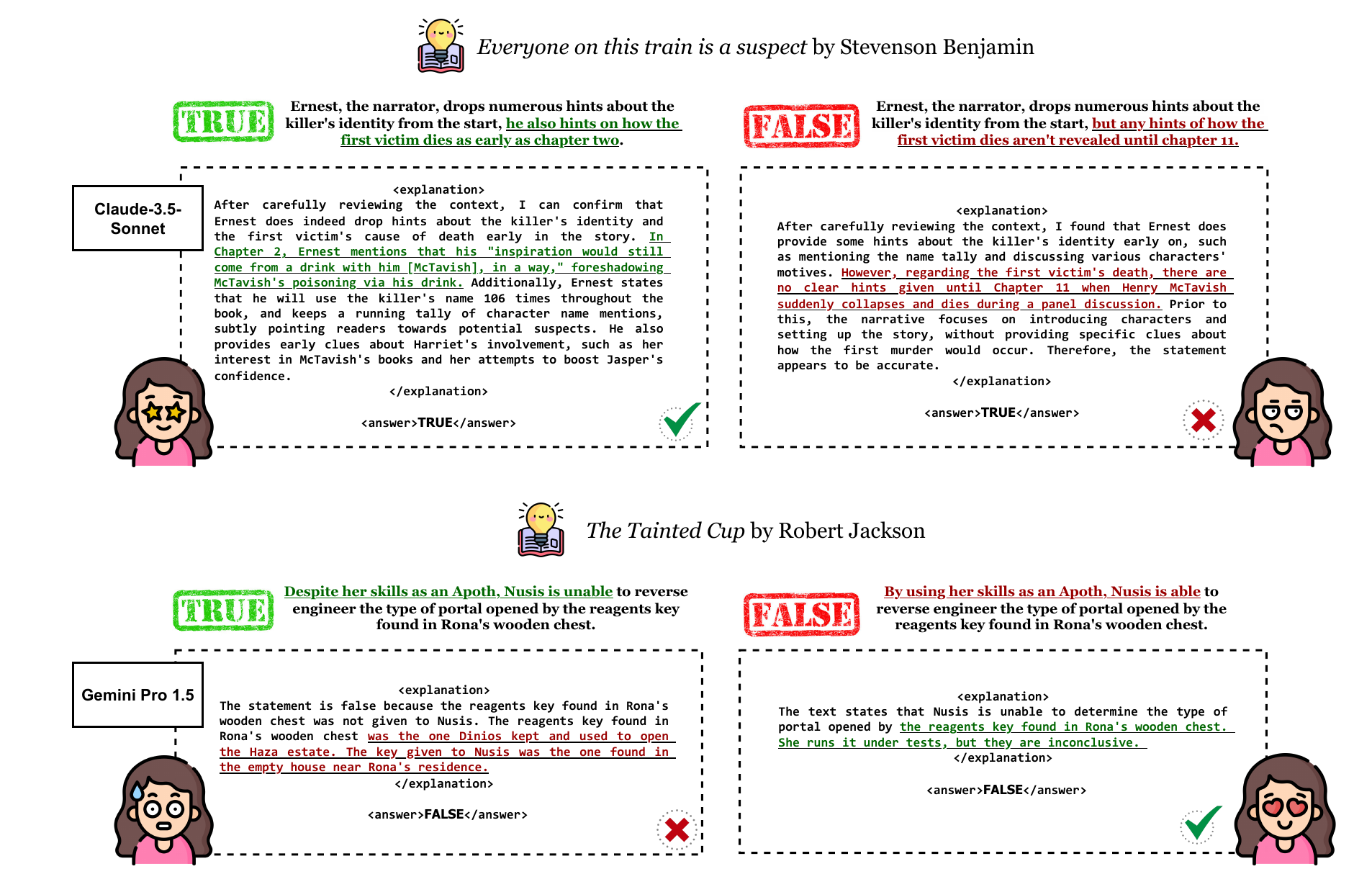} 
    \caption{Examples of claim pairs where the models failed to validate one of the claims in the pair. Employing narrative minimal pairs helps us avoid awarding the model for cases where it only \textit{appears} to produce the correct answer, when in fact the prediction was made without fully or efficiently utilizing the context. In the top example, the model first correctly identifies hints dropped by the author about how the first victim dies (when verifying the true claim),\tablefootnote{In fact, after the first victim dies, the narrator mentions: \textit{"All I can tell you is what I’ve been telling you so far: the truth. After all, I told you Henry McTavish would be poisoned, didn’t I? Well, not in those words, I suppose. But I did say the inspiration for this book would come from a drink with him."} It is also worth noting that the model incorrectly states that the first victim dies during the panel, when in fact McTavish dies the next day during breakfast.} then incorrectly claims that no such hints exist (when verifying the false claim). In the bottom example, the model first incorrectly claims that the key given to Nuna was not found in Rona's wooden chest (when verifying the true claim), then correctly raises no objections to the fact that the key was found in Rona's wooden chest (when verifying the false claim).}
  \label{fig:pair_failure}
\end{figure*}

\begin{table*}[h!]
    \centering
    \footnotesize
    \renewcommand{\arraystretch}{1.2}
    \resizebox{0.9\textwidth}{!}{%
    \begin{tabular}{p{0.33\textwidth} p{0.5\textwidth} p{0.45\textwidth}}
        \toprule
         \textsc{\textbf{Claim \& Gold Label}} & \textsc{\textbf{Model's Label \& Justification}} & \textsc{\textbf{Human Explanation}} \\
        \midrule
        \textit{Wyatt apologizes to Ernest because he believes
        McTavish left him a negative review.} \newline 
        
        \textit{Everyone on the Train is a Suspect} \newline 
        \makebox[110pt][r]{by Benjamin Stevenson}  \newline 

        \makebox[250pt][r]{\texttt{{\color{teal}\textbf{\Large--True}}} \newline} \newline 

        & \textit{Wyatt's apology to Ernest is mentioned in the context, but it is not because he believes McTavish left him a negative review. \textbf{\color{purple}Instead, Wyatt's apology is vague and not directly linked to the review.} Ernest himself is confused about the reason for the apology, and it is later revealed that Simone, Ernest's agent, was behind the negative review as part of a coded message to McTavish. Therefore, the statement that Wyatt apologizes to Ernest because he believes McTavish left him a negative review is incorrect.} \newline
        \newline
        --\gpto{}: \texttt{{\color{purple}\textbf{\Large False}}}
        
        & When Wyatt apologizes, Ernest does not initially understand the reason behind it. However, it is \textit{heavily} implied shortly after that Wyatt, McTavish's publisher, was apologizing for the bad review, which at that time everyone, including Wyatt thought was written by McTavish himself. \\
        \textit{Emily Sparrow was pregnant with John's child and disappeared in 1966.} \newline 
        
        \textit{How to Solve Your Own Murder} \newline
        \makebox[90pt][r]{by Kristen Perrin} \newline

        \makebox[250pt][r]{\texttt{{\color{teal}\textbf{\Large --True}}} \newline} \newline 
        
        & \textit{The text states that Emily Sparrow was pregnant, but \textbf{\color{purple}it is never explicitly stated who the father of the child is}. While it is implied that the father could be John, it is also implied that it could be Rutherford Gravesdown (Ford).} 
        \newline
        \newline
        --\geminipro{}: \texttt{{\color{purple}\textbf{\Large False}}}
        
        & Throughout the book, the reader is taken back to 1966, where it is revealed that Emily was pregnant, and the father's identity was uncertain—either Lord Gravesdown or John. The baby, Laura, was later adopted and had a daughter, Annie. In the latter part of the book, set in the present day, Annie meets John, who confesses that he is her grandfather, thus confirming that he was the father of Emily's baby.  \\
        \textit{None of the writers on the train went to the Gorge excursion as they had to stay to participate in a festival event.} \newline
        
        \textit{Everyone on the Train is a Suspect} \newline 
        \makebox[110pt][r]{by Benjamin Stevenson} \newline

        \makebox[250pt][r]{\texttt{{\color{purple}\textbf{\Large --False}}} \newline} \newline 
        
        & \textit{The statement is true based on the context. The passage states that while the non-festival tourists went on a scheduled day trip sailing down Katherine Gorge, "In front of our carriages were forty or so black fold-out chairs set up in the red dirt for the festival attendees." \textbf{\color{purple}This indicates that the writers stayed behind for a festival event} while the other passengers went on the Gorge excursion.} 
        \newline
        \newline
        --\claude{}: \texttt{{\color{teal}\textbf{\Large True}}}
        
        & While all writers invited to the festival stayed for the event, Juliet, who was also an accomplished writer and, in fact, we later learn she was originally invited as well, went on the trip. \\

        \textit{Lost Bride Manor is haunted by various spirits, including the benevolent ghost of Clover, Sonya's grandmother and Collin's mother, who communicates through music.} \newline
         
        \textit{Inheritance} \newline 
        \makebox[110pt][r]{by Nora Roberts} \newline 

        \makebox[250pt][r]{\texttt{{\color{teal}\textbf{\Large --True}}} \newline} \newline
        
        & \textit{The statement is mostly true, but contains one key inaccuracy. The context does indicate that Lost Bride Manor is haunted by various spirits, including Clover, who communicates through music. Clover is described as Sonya's grandmother and Collin's mother, which is correct. However, the statement characterizes Clover as "benevolent", but the context does not provide enough information to definitively label her as such. \textbf{\color{purple}While Clover seems to be helpful at times, playing music and aiding Sonya, the text does not explicitly state that she is entirely benevolent.} Given this small but significant discrepancy, the statement cannot be considered entirely true.} 
        \newline
        \newline
        --\sonnet{}: \texttt{{\color{purple}\textbf{\Large False}}}
        
        & Clover is a ghost haunting the manor where Sonya moves in. From the first day, Clover, who we learn is the ghost of Sonya's grandmother, helps Sonya by playing music to cheer her up, adding to her courage, and keeping her company. Clover also tries to protect Sonya from the evil spirit that also haunts the manor. Although it is not directly stated in the text, Clover's kind and well-meaning nature towards Sonya is evident through her actions. \\
        \midrule  
        \textit{Ernest, the narrator, drops numerous hints about the killer's identity from the start, but any hints of how the first victim dies aren't revealed until chapter 11.} \newline 
        
        \textit{Everyone on the Train is a Suspect} \newline 
        \makebox[110pt][r]{by Benjamin Stevenson} \newline

        \makebox[250pt][r]{\texttt{{\color{purple}\textbf{\Large --False}}} \newline} \newline 
        
         & \textit{To determine the accuracy of the statement, I reviewed the context for any early hints about the killer's identity and the method of the first victim's death. The prologue and early chapters contain several clues about the characters' motives and potential involvement in the murder. Specifically, \textbf{\color{purple}the prologue includes a detailed description of a potential murder scene, mentioning "the bloodied handprint on the bathroom door" and "the crimson, dripping tip of the murder weapon."} This indicates that hints about the method of the first victim's death are indeed revealed before chapter 11.}
        \newline
        \newline
        --\gpto{}: \texttt{{\color{purple}\textbf{\Large False}}}       \newline
         & It is true that Ernest, the narrator, drops some hints about how the first victim dies from the start. However, the passage cited by the model is irrelevant. It is taken from the Prologue, where Ernest, a mystery novel writer, talks about elements that typically appear in a mystery novel's prologue. He mentions a "bloodied handprint" and "the crimson, dripping tip of the murder weapon," which are common in mystery novels prologes. However, in this novel, the first victim was poisoned, and Ernest hints at this by saying that the inspiration for this book "came from a drink" with the victim. \\
      
         \textit{Levi promises Grace that he will kill anyone who tries to hurt her, and Grace calls him to help her dispose of Norman's body.} \newline 
         
        \textit{Safe and Sound} \newline 
        \makebox[110pt][r]{by Laura McHugh} \newline

        \makebox[250pt][r]{\texttt{{\color{purple}\textbf{\Large --False}}} \newline} \newline
        
         & \textit{In the context, Levi does promise Grace that he would kill anyone who tried to hurt her again. However, when Grace kills Norman,…she calls Tyson, Levi's brother. The passage states: "She…wanted to make it look like she couldn't have survived. So she could just go missing."… \textbf{\color{purple}It then says "Tyson grabbed me, clamped his hands around my neck, and didn't let go. `I dumped a body for you.'"} This shows that it was Tyson, not Levi, who Grace called to help dispose of Norman's body.}
        \newline
        \newline
        --\gpto{}: \texttt{{\color{purple}\textbf{\Large False}}}        \newline
         & The model's justification begins correctly, but the first passage cited is unrelated to the claim and clearly a lie told by Tyson in the context of the story. Additionally, the text "Tyson grabbed me, clamped his hands around my neck, and didn't let go." never appears in the book. Tyson does say "I dumped a body for you." to Grace---which proves the claim is false---but he does not choke her before saying so or at any other point in the book.\\

         \textit{While on Sigrid's boat, Seventeen admits to Kat that he had the same underwater dream that she had.} \newline 
         
        \textit{Assassin Eighteen} \newline 
        \makebox[110pt][r]{by John Brownlow} 

        \makebox[250pt][r]{\texttt{{\color{teal}\textbf{\Large --True}}} \newline} \newline
        
         & \textit{The statement is true based on the context provided. In Part 8, \textbf{\color{purple}after Kat is rescued from the water by Rachel and Vilmos in the Zodiac}, she describes her near-death experience underwater where she felt she and Seventeen were entwined as one. Seventeen then admits "I had the same dream", confirming he experienced the same vision while he was drowning. }
        \newline
        \newline
        --\claude{}: \texttt{{\color{teal}\textbf{\Large True}}}
         & It is true that Seventeen admits he had the same dream in Part 8 of the book. However, Kat was not rescued by Rachel and Vilmos in the Zodiac during this scene. That happens much later in the book, towards the end of Part 11. This conversation happened when Kat and Seventeen were rescued by Sigrid. \\

         \textit{Sonya befriends Trey's sister Anna, a potter, and his cousin Owen Poole, who becomes involved in solving the mystery of the lost brides.} \newline 
        
        \textit{Inheritance} \newline 
        \makebox[110pt][r]{by Nora Roberts} \newline 
        
         \makebox[250pt][r]{\texttt{{\color{purple}\textbf{\Large --False}}} \newline} \newline 
         & \textit{The statement is mostly true, but contains one key inaccuracy. The context shows that Sonya does befriend Trey's sister Anna, who is a potter, and helps redesign her website. Sonya also meets and interacts with Owen Poole, \textbf{\color{purple}Trey's cousin, who becomes involved in investigating the haunted manor and the mystery of the lost brides. However, the statement is incorrect in saying that Owen "becomes involved in solving the mystery of the lost brides."} While Owen does help investigate strange occurrences at the manor and even enters the mirror with Sonya at the end, he is not described as actively trying to solve the mystery of the lost brides specifically. Sonya is the main character focused on that mystery. Therefore, while most of the statement is accurate, this one detail makes it not entirely true.}
        \newline
        \newline
        --\sonnet{}: \texttt{{\color{purple}\textbf{\Large False}}}
         & It is true that Sonya befriends Trey's sister, Anna, and Owen. However, Owen is not Trey's cousin; he is Sonya's cousin, whom she did not know about. Owen and Trey then try to help Sonya solve the mystery of the lost brides. \\
         
        \bottomrule
    \end{tabular}
    }
   \caption{\textbf{Top row:} Example of a claim where the model generated incorrect label, along with the model-generated label and justification, and human-written explanations for the correct label. \textbf{Bottom row:} Example of claim correctly labeled by the model but with wrong justifications. The explanations were provided by the authors, based on examples from books they read. Tags have been removed from the generated content for clarity. 
   }
    \label{tab:generation_example}
\end{table*}

\autoref{tab:pair_accuracy_models_with_classics} reports results for individual models as well as for a common set of claims shared among closed-source models.\footnote{Open-weights models are excluded as this would significantly reduce the number of shared pairs to just over 100.} Overall, \gpto{} exhibits the highest accuracy at 55.8\% (58.2\% for the common set), though it still misidentifies at least one claim in the pair 44.2\% of the time (see \autoref{tab:generation_example} for example failure cases). The second-best performing model is \claude{} with an overall accuracy of 49.4\%, followed by \geminipro{} with an accuracy of 48.1\%. \sonnet{}, \gptturbo{} and \geminiflash{} perform worse, with accuracies of 41.0\%, 40.2\% and 34.2\%, respectively.\footnote{See \autoref{fig:all_models_failed} for example of claim on which all models failed to generate the accurate label.} All open-weight models perform below random (25\%), ranging from 22.5\% for \comr{} (with a simplified prompt) to as low as 3.9\% for \gemma{}. Because of their poor accuracies, we exclude open-weights models from the following analysis; see \S\ref{app_sec:results_info} for details.

\paragraph{Good performance on ``needle in the haystack'' does not imply high \name\ accuracy:} In \autoref{tab:ruler_vs_1001}, we compare the performance of \comr{} and \gptturbo{} on \name\ with the results reported in \citet[][\textsc{Ruler}]{sun2024ruler}, a more complex variant of NIAH, to estimate how performance on \textsc{Ruler} translates to our task. Despite both models achieving high performance on \textsc{Ruler} (84.8\% and 89.6\%, respectively, for the longest tested context of 128k),
they underperform on our task, with \gptturbo{} achieving only 40.2\% accuracy and \comr{} performing \textit{below} random at 19.6\%.\footnote{Note that \textsc{Ruler} reports the results for \texttt{gpt-4-1106-preview} checkpoints for \gptturbo{}, while the current work uses \texttt{gpt-4-turbo-2024-04-09} checkpoints, presumably a more advanced version of that model.} This shows that synthetic retrieval-focused benchmarks are insufficient to evaluate global long-document reasoning.

\begin{table}[t]
\centering
\resizebox{0.48\textwidth}{!}{%
\begin{tabular}{lccc}
\toprule
\textsc{Scope} & \textsc{Correct} & \textsc{Incorrect} & \textsc{Accuracy (\%)} \\
\midrule
\faBook\ \textsc{Global} & 104 & 146 & 41.6 \\
\faFileTextO\ \textsc{Passage} & 101 & 111 & 47.6 \\
\faAlignLeft\ \textsc{Sentence} & 49 & 33 & 59.8 \\
\bottomrule
\end{tabular}
}
\caption{Overall accuracy on claim pairs of \faLock\ closed-source models on a subset of data annotated for evidence scope (see \S\ref{sec:data_methods} for the annotation process). Pairs requiring global reasoning to verify are most difficult for models.
\autoref{fig:scope_annot_acc_by_model} shows results by model.
}
\label{tab:scope_acc_total}
\end{table}

\paragraph{\name\ claims that require global reasoning are particularly difficult to verify:} \autoref{tab:scope_acc_total} contains  further analysis of model accuracy on a subset of \name\ annotated for the scope of evidence (see \S\ref{subsec:data_collection} for details). Overall, models perform worst for claim pairs requiring global reasoning (41.6\%), followed by reasoning over a longer passage (47.6\%), and, finally, sentence-level evidence (59.8\%). While performance on sentence-level evidence is higher than in the other two setups, it is still much lower than the ``needle-in-a-haystack'' performance reported in \citet{sun2024ruler}. This indicates that \name\ claim pairs with sentence-level evidence are still much harder to solve than NIAH, possibly due to NIAH's out-of-context evidence injection.

\begin{figure}[t]
  \includegraphics[width=1\linewidth]{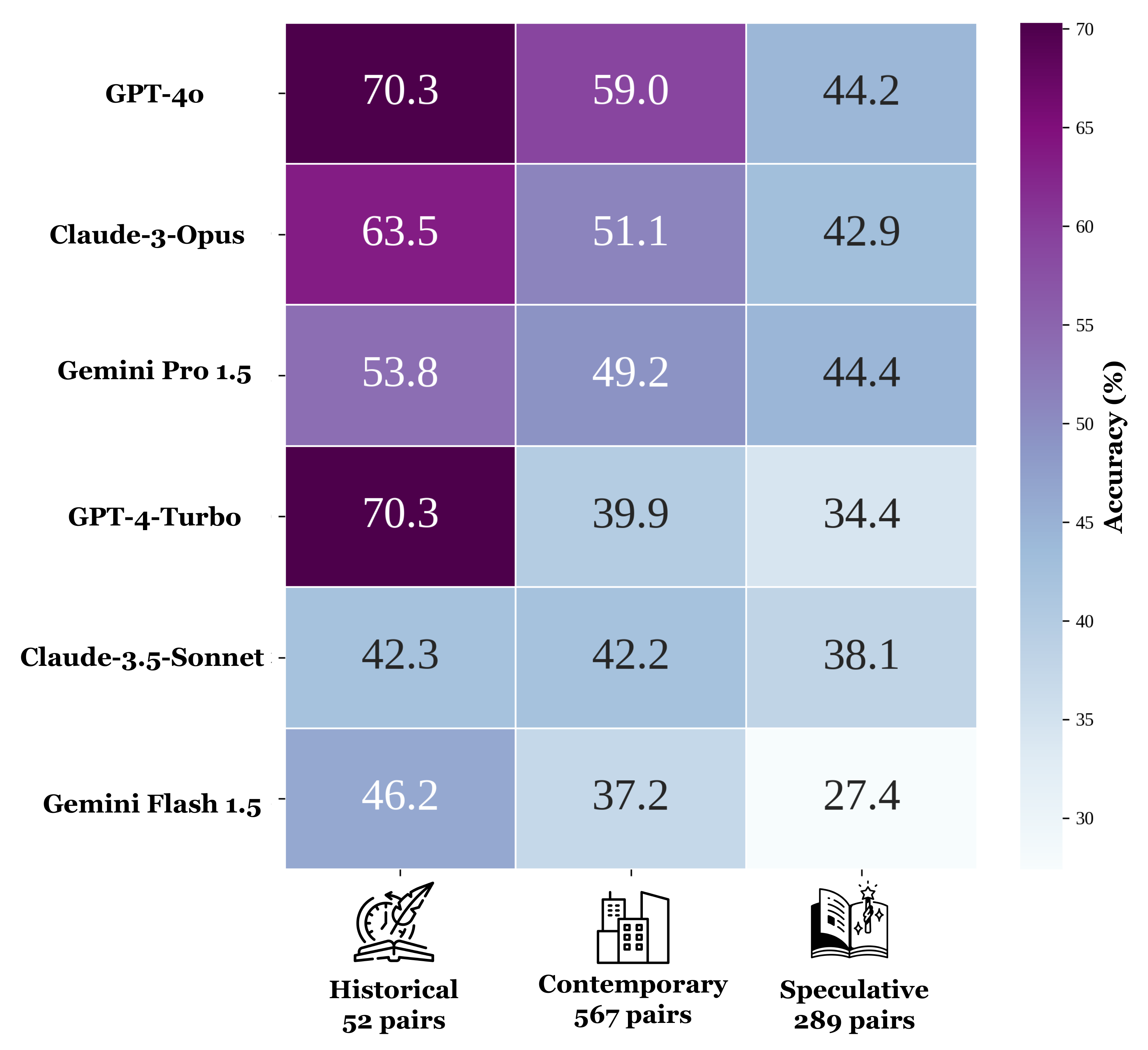} 
  \caption{Performance of closed-source models on different types of novels. Note that two novels were excluded from this analysis as they could not clearly be classified in one of these categories. We provide the total number of claims in each category for reference however these numbers will vary slightly between the models due to the context-limitation and \geminipro{}/\geminiflash{} refusals.}
  \label{fig:genre_acc_heatmap}
\end{figure}

\paragraph{Claims about speculative fiction are harder to verify:} Are texts with extensive world-building more difficult for models to process than those that take place in a fictional but realistic world? We categorized 65 out of 67\footnote{One book was a nonfiction collection of essays and another split the plot between modern day and the past, rendering it a combination of historical and contemporary fiction.} of the books in our dataset into one of three broad categories:     
\begin{itemize}
    \item \faHistory\ \textbf{Historical}:  works set in our world before World War II, without any unrealistic elements.
    \item \faBuildingO\ \textbf{Contemporary}: works set in our world after World War II, without any unrealistic elements.
    \item \faMagic\ \textbf{Speculative}: works set in an alternate version of our world, containing both realistic and unrealistic elements, or in a completely invented universe.
\end{itemize}
The accuracy across all six closed-source models is 56.4\% for historical fiction, 46.8\% for contemporary fiction, and 38.8\% for speculative fiction. Figure \ref{fig:genre_acc_heatmap} shows that this pattern holds for each individual model, with accuracy being highest for historical fiction, followed by contemporary fiction, and lowest for speculative fiction. These results support the intuition that texts set in a realistic version of our world require less ``work'' from models to reason over than texts set in a universe that is defined within the text, perhaps because the models can rely more on their parametric knowledge.\footnote{We acknowledge that there may be confounding factors, such as the relative difficulty of writing challenging claims for different books types and the high average length of speculative fiction books (148k tokens)---though we note that all models perform better on historical (avg. length 133k tokens) than the shorter contemporary books (avg. length 115k tokens).}

\begin{figure}[t]
  \includegraphics[width=1\linewidth]{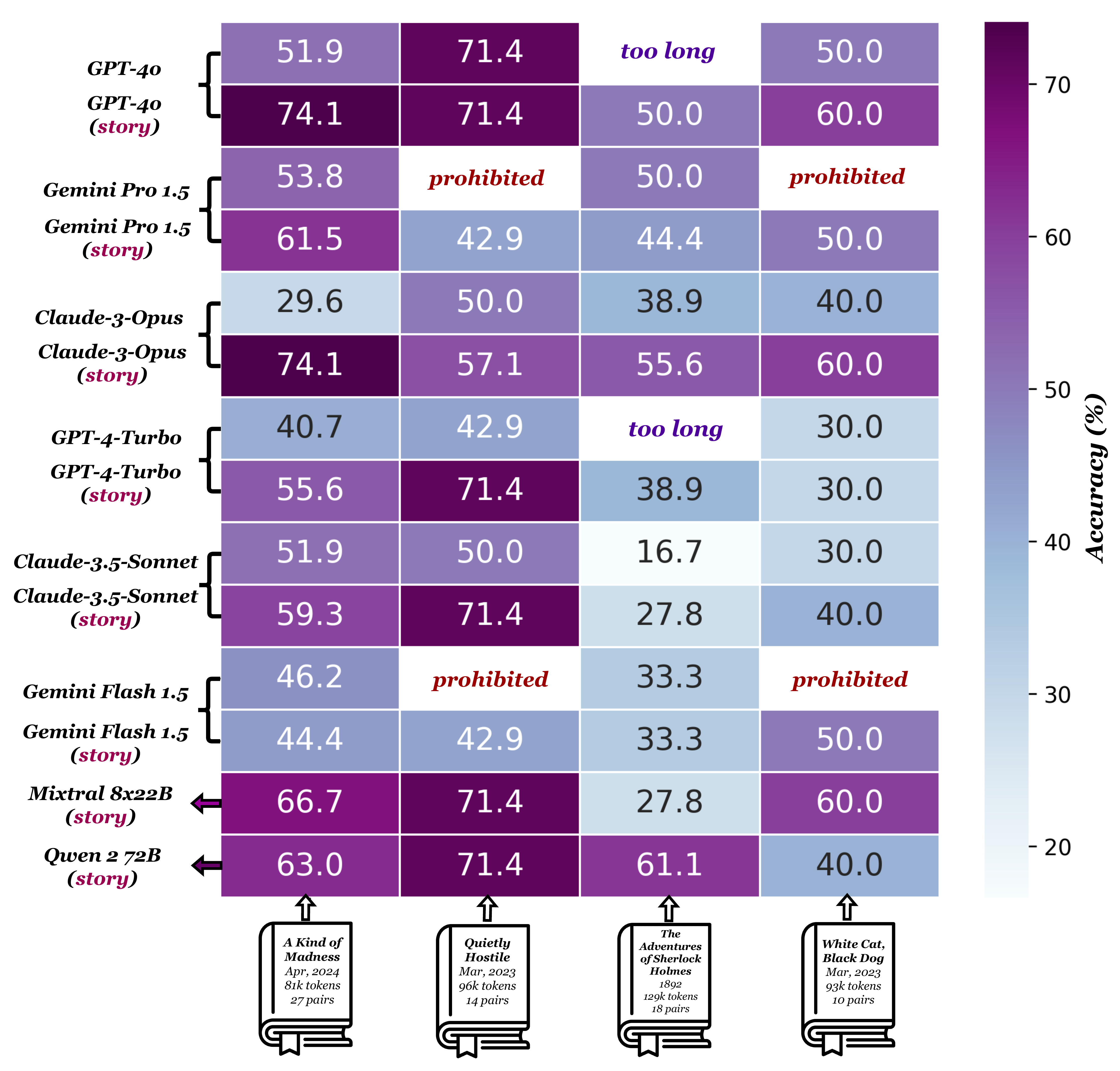} 
  \caption{Model accuracy on claim pairs for different stories within collections. Accuracy is shown for (1) using the entire collection as context when prompting about a story, and (2) using only the individual story ("story") for the same set of claims. For \gpto{} and \gptturbo{}, one book was too long, so only the "story" context performance is presented. \geminipro{} and \geminiflash{} refused to process two books but handled the stories within, so only "story" context performance is available. We also provide performance of \mixtral{} (65k) and \qwen{} (32k)\tablefootnote{Here we employ the 32k, which can also be extended to 128k using \textsc{YaRN} \cite{peng2023yarnefficientcontextwindow}.} on story-level input, for comparison.}
  \label{fig:stories_acc_heatmap}
\end{figure}

\paragraph{Irrelevant context confuses models:} If we know that a claim can be verified by a short span of the book in isolation, does providing the rest of the book affect its accuracy? \autoref{fig:stories_acc_heatmap} shows model performance on four \emph{short story collections}, where the model is either given just the short story relevant to the claim ($\sim$21k tokens)\footnote{Mean=8.5k, Min=702, Std.Dev.=4,355.} or the entire collection ($\sim$129k tokens). By prompting the model with the same claims, we partially control for confounding variables such as the inherent difficulty of the claim. While the Gemini models are relatively robust to added context, \claude{}'s pair accuracy drops as much as 44.5\% absolute when given the collection vs. the story, while \sonnet{}, \gptturbo{} and \gpto{} exhibit smaller but still substantial drops.\footnote{Notably, all models perform suboptimally on the short stories, suggesting that reasoning tasks may be challenging, even with shorter contexts.} More generally, it is unclear whether claims about longer books (with potentially more irrelevant context) are harder to verify; see \S\ref{app_sec:results_info} for more details.

\begin{table}[t]
\centering
\footnotesize
\resizebox{0.4\textwidth}{!}{%
\begin{tabular}{lc}
\toprule
\textsc{Model} & \textsc{Incorrect \% \textsubscript{(\textit{incorrect/total})}} \\ 
\midrule
\gpto & 21.7 \textsubscript{(15/69)} \\
\gptturbo & 45.0 \textsubscript{(27/60)} \\ 
\claude & 16.9 \textsubscript{(13/77)}  \\ 
\geminipro & 28.3 \textsubscript{(13/46)}  \\ 
\geminiflash & 65.9 \textsubscript{(27/41)}  \\ 
\bottomrule
\end{tabular}
}
\caption{The percentage of incorrect justifications provided for correct label by model which generated it in the analyzed subset of data.\tablefootnote{We do not analyze \sonnet{}'s explanations because the model was run on June 22, after the completion of this analysis. Nonetheless, we have observed mistakes in its explanation, with an example provided in \autoref{tab:generation_example}.} } 
\label{tab:evidence_eval}
\end{table}

\paragraph{Model-generated explanations are not always accurate:} 
Do models correctly explain why a claim is true or false? Three of this paper's authors examined explanations generated for \name\ books that they read, covering a subset of 7 books (293 claims).
\autoref{tab:evidence_eval} shows that  \textit{no} model consistently produces accurate explanations for every correctly-labeled claim, indicating a potential reliance on flawed or incomplete reasoning (see \autoref{tab:generation_example} for examples). \claude{} demonstrated the highest explanation accuracy, with 16.9\% of its explanations containing errors, followed by \gpto{} and \geminipro{} with 21.7\% and 28.3\% incorrect explanations respectively. \gptturbo{} and \geminiflash{} were the worst performing models with incorrect explanations reaching 45.0\% and 65.9\% respectively.\footnote{As we run \sonnet{} on June 22, we exclude it from this analysis.} This is problematic when considering that users tend to rely on model explanations when verifying claims, even when the explanation is incorrect \citep{si2024large}. Further discussion on citations of evidence from the source found in model justifications can be found in \S\ref{app_sec:results_info}.

\begin{table}[t]
\centering
\resizebox{0.48\textwidth}{!}{%
\begin{tabular}{lccc}
\toprule
 & \multicolumn{3}{c}{\bm{}} \\
\cmidrule(lr){2-4}
\textsc{Scope} & \textit{k=5} (\%) & \textit{k=25} (\%) & \textit{k=50}  (\%) \\
\midrule
\faBook\ \textsc{Global} & 25.9 & 29.3 & 41.4 \\
\faFileTextO\ \textsc{Passage} & 22.9 & 43.8 & 45.8 \\
\faAlignLeft\ \textsc{Sentence}  & 46.7 & 66.7 & 73.3 \\
\bottomrule
\end{tabular}
}
\caption{Performance of the \bm{} pipeline on the subset of data annotated for evidence scope.}
\label{tab:bm25_vs_evidence_scope}
\end{table}

\paragraph{Can {\textsc{\small BM25}} help prioritize important context?} We observe that our {\textsc{\small BM25}}-assisted \gpto{} approach with $k = 50$ excerpts performs better than all models except for \gpto{} with the full book. For $k=50$, an average of only 17\% of a book was retrieved by \textsc{\small{BM25}} and fed to \gpto{} in ranked order. Retrieval-based methods struggle on global reasoning due to out-of-order chunks, but they are effective for claims that require sentence- and, to a much lesser extent, passage-level retrieval to verify; thus, there is likely an upper bound to their \name\ accuracy (see \autoref{tab:bm25_vs_evidence_scope}).

\begin{table}[t]
\centering
\resizebox{0.9\columnwidth}{!}{%
\begin{tabular}{lcc}
\toprule
\textsc{Model} & \textsc{\color{teal}True}\textsubscript{(correct/total)} & \textsc{\color{purple}False}\textsubscript{(correct/total)} \\
\midrule
\gpto & 77.5\textsubscript{(478/617)} & 75.9\textsubscript{(468/617)}  \\
\gptturbo & 57.2\textsubscript{(353/617)}  & 78.8\textsubscript{(486/617)}  \\
\claude & 82.2\textsubscript{(770/937)}  & 64.7\textsubscript{(606/937)}  \\
\sonnet & 55.3\textsubscript{(518/937)}  & 83.5\textsubscript{(782/937)}  \\
\geminipro & 59.7\textsubscript{(307/514)}  & 83.7\textsubscript{(431/515)}  \\
\geminiflash & 51.7\textsubscript{(266/515)}  & 78.1\textsubscript{(402/515)}  \\
\midrule
\comr{} & 59.3\textsubscript{(264/445)} & 52.8\textsubscript{(235/445)} \\
\comr{}\textsubscript{simple} & 74.4\textsubscript{(331/445)}  & 44.3\textsubscript{(197/445)}  \\
\comrplus{} & 58.7\textsubscript{(261/445)}  & 49.2\textsubscript{(219/445)}  \\
\comrplus{}\textsubscript{simple} & 34.6\textsubscript{(154/445)}  & 75.3\textsubscript{(335/445)} \\
\phimodel{} & 34.8\textsubscript{(88/253)}  & 46.5\textsubscript{(118/254)}  \\
\phimodel{}\textsubscript{simple} & 69.5\textsubscript{(230/331)} & 32.0\textsubscript{(106/331)} \\
\gemma & 20.2\textsubscript{(202/1001)}  & 27.8\textsubscript{(278/1001)}  \\
\gemma{}\textsubscript{simple} & 77.5\textsubscript{(776/1001)} & 18.7\textsubscript{(187/1001)}  \\
\longllama\textsubscript{simple} & 63.8\textsubscript{(598/937)}  & 19.7\textsubscript{(185/937)} \\
\midrule
\bm{} (\textit{k=5}) & 33.7\textsubscript{(337/1001)} & 90.0\textsubscript{(901/1001)}  \\
\bm{} (\textit{k=25}) & 57.5\textsubscript{(576/1001)}  & 81.7\textsubscript{(818/1001)}  \\
\bm{} (\textit{k=50}) & 65.7\textsubscript{(658/1001)}  & 79.6\textsubscript{(797/1001)}  \\
\midrule
\textsc{Random} & 50.0\textsubscript{(500/1001)}  & 50.0\textsubscript{(50/1001)}  \\
\bottomrule
\end{tabular}
}
\caption{Model accuracy on True and False claims for all data (see \autoref{tab:prompt_template_claim_eval} for the prompt; and \autoref{tab:prompt_template_claim_eval_bm25} for the prompt employed with BM25). ``\textsc{Common set}'' refers to claim pairs shared among the models. The subscript ``\textsc{simple}'' refers to the calls made with the simplified prompt (\autoref{tab:prompt_template_claim_eval_simple}).}
\label{tab:model_acc_on_true_and_false}
\end{table}

\paragraph{Models have different predilections for predicting {\color{teal}True} vs {\color{purple}False}:} Our pairs were designed so that validating one claim should enable validation of the other. However, we observe in \autoref{tab:model_acc_on_true_and_false} that some models tend to predict one label much more frequently than another. This tendency was particularly evident in \sonnet{}, \geminipro{}, \geminiflash{}, and \gptturbo{}, which had strong preferences for predicting False, and is in line with the observation reported for \geminipro{} in \citet{levy2024flenqa}. In contrast, \claude{} exhibited much higher accuracy on True labels (82.2\%) compared to False (64.7\%). \gpto{} was the only balanced model among the closed-source models, with accuracies of 77.5\% for True and 75.9\% for False.

\section{Related Work}
\label{sec:rel_work}

\paragraph{Evaluation of long-context models:}  Several benchmarks with inputs ranging from at least 8k up to 1M tokens have recently been introduced to evaluate long-context language models. Some of these benchmarks use synthetic tasks that can be generated programmatically or by LLMs, such as NIAH  \cite{tay2021long, sun-etal-2021-long, kamradt2023needle, li2023loogle, sun2024ruler, levy2024flenqa,liu2024lostmiddle,Lee2024LongContext, laban2024summaryhaystackchallengelongcontext, vodrahalli2024michelangelolongcontextevaluations}. Others contain "realistic" tasks that typically require human annotators to devise, like traditional QA, summarization, and claim verification \citep{shaham-etal-2023-zeroscrolls, yushi2023longbench, zican2024bamboo, wang2024leavedocumentbehindbenchmarking} sometimes created by utilizing existing datasets \cite{hudson-al-moubayed-2022-muld, li2024longcontextllmsstrugglelong}. The existing long-context evaluation benchmarks with literary tasks all consist of publicly available English \citep{hudson-al-moubayed-2022-muld, zhang2024inftybench,wang2024novelqa,
chenxin2024leval} or Chinese language books \citep{yu-etal-2024-lfed-literary}. 
All but \citet{chenxin2024leval} contain generative or multiple choice questions, including true/false questions, but unlike \name, these are not minimal pairs.

\paragraph{Claim verification:} Our work relates to prior work on claim verification, where claims are verified against a whole knowledge datastore \citep{thorne-etal-2018-fever, wadden-etal-2020-fact, schuster-etal-2021-get} or single evidence documents \cite{maynez-etal-2020-faithfulness, falke2019ranking}. While earlier methods primarily relied on task-specific natural language inference \cite{laban2022summac, honovich2022true} or question answering models \cite{fabbri2022qafacteval, wang2020asking} for claim verification, more recent work has explored using LLMs to evaluate the factuality of long-form model-generated text \citep{min-2023-factscore, wei2024longform, manakul2023selfcheckgpt}. 

\paragraph{Minimal pairs:} Minimal pairs, or contrast sets, are pairs of test set instances where a slight but impactful difference between the instances affects the gold label. \citet{gardner-etal-2020-evaluating} proposed contrast sets as a method for evaluating models on their intended tasks by manually perturbing test set instances. Our narrative minimal pairs were inspired by the \textsc{\small{DEMETR}} \citep{karpinska-etal-2022-demetr} dataset of minimals pairs for evaluating machine translation metrics and the BLiMP \citep{warstadt-etal-2020-blimp-benchmark} benchmark of minimal pairs for evaluating English grammatical phenomena.


\section{Conclusion and Discussion}

We introduce \name, a claim verification dataset designed to evaluate long context LLMs in a realistic task setting. Our design ensures that most claims necessitate global reasoning over extended contexts for verification and cannot be easily ``gamed'' by relying solely on parametric knowledge or generating correct answers through incorrect reasoning. Our experiments with 11 different long-context LLMs (5 open-weight, 6 closed-source) demonstrate that all models struggle on \name. Furthermore, we reveal a substantial competency gap with human readers who can very easily perform this task. 

Importantly, our results show that models that are ``state-of-the-art'' according to synthetic benchmarks like NIAH actually perform very poorly on our meticulously designed dataset. Nevertheless, we argue that complex synthetic datasets (such as \textsc{Ruler}) are useful and complementary to our realistic dataset; they allow for much higher flexibility such as easily adjusting for different context lengths or analyzing the lost-in-the-middle phenomenon. We encourage researchers to use a holistic approach and consider \textit{both} synthetic and realistic tasks when evaluating long-context language models. 
\section*{Limitations}

The scope of our work is limited to novels published in English and the task of claim verification. It is unclear how the models' performance would translate to other languages, domains, or realistic tasks, and we leave that for future work.

We also acknowledge that this study and the methodology proposed are inherently and possibly prohibitively expensive due to the hiring of expert annotators and the thousands of LLM API calls. Because using the model developer's API is often the only way to access a closed-source model, the extent of our evaluation was limited to whether or not the closed-source models provided a response to our prompts. This resulted in a common set of only 354 pairs that all closed-source models labeled, making it difficult to truly compare the models to each other. Finally, while we commit to periodically updating the dataset and evaluating the models ourselves, we are unable to release the entire dataset due to copyright restrictions and to prevent model providers from training on it.
\section*{Ethical Considerations}

The data collection protocol was reviewed by the 
University of Massachusetts, Amherst 
Institutional Review Board and received a \textit{Not Human Subjects Research} determination
(IRB: \#5587). 
All annotators consented to the use and publication of their annotations, which we will release for the portions of data that is \textit{not} under copyright. Additionally, we ensured annotators received fair compensation for their contributions and respected their preferred rates. All copyrighted books were purchased using the funds that supported this work. We will \textit{not} release the copyrighted portion of the data. Instead, we commit to consistently updating the dataset with newly published novels and evaluating the models ourselves. 
\section*{Acknowledgements}
We would like to extend our gratitude to the Upwork annotators for their dedicated efforts, and to the anonymous reviewers for their time and valuable feedback. We are also grateful to the members of the UMass NLP lab for their insightful input, with special thanks to Chau Pham, Nader Akoury, Yekyung Kim, and Yapei Chang. Our deep appreciation goes to Byron Wallace for suggesting the project name and to Simeng Sun for the discussions that inspired the initial concept of this work.  This project was partially supported by awards IIS-2202506 and IIS-2312949 from the National Science Foundation (NSF) as well as an award from Adobe. 

\bibliography{custom}

\appendix
\section{Note on dataset versions}

Results on \name\ are being updated as new models are released. The most recent results can be found at \href{https://novelchallenge.github.io/}{\path{https://novelchallenge.github.io/}}. We are also actively collecting new claims/books and plan to update the benchmark periodically. While working on the collection of new claims we put more emphasis on reasoning over longer context and work only with the annotators who proved to be the best at creating their claims in the initial study. At the time of writing about 10 new books were added to the collection.

\section{Issues with current long-context evaluation benchmarks}


\subsection{Evaluation may be conducted on the training data}
\paragraph{Data contamination:} One of the most prominent issues in current evaluation practices is the test data. A fundamental rule in evaluating any trained model is to test it on a separate withheld \textit{test} set, avoiding the instances the model was trained on \cite{van-der-goot-2021-need, kapoor_sayash_leak_and_repro2023}. However, this has become increasingly difficult as existing test sets have almost certainly been leaked into the training data, and most available LLMs are either entirely closed-source or only open-weights, meaning we do not have access to the training instances \cite{sainz-etal-2023-nlp, balloccu-etal-2024-leak}. While analyzing existing long-context benchmarks, though this issue is certainly not limited to long-context setups, we noticed that many either utilize existing datasets \cite[e.g.,][]{yushi2023longbench, chenxin2024leval} or use old texts due to copyright issues \cite[e.g.,][]{li2023loogle, wang2024novelqa}, both of which are likely included in the models' training data. Some researchers have tried to mitigate this issue by transforming the original texts, for instance, by replacing the named entities with others \cite{zhang2024inftybench, zican2024bamboo}. However, we observe that prompting models with such data (even fragments of books) still leads to recognition of the source materials, an outcome in line with results reported in \citet{chang-etal-2023-speak}.

\paragraph{Issues with closed-book tests:}
A popular method used to ensure that the training data do not affect the model's performance is to perform the \textit{closed-book} test \cite{roberts-etal-2020-much, ciosici-etal-2021-perhaps}, used, for instance, in \citet{wang2024novelqa}. Just like a student taking a test without access to the study book, the model is tested without access to the source text. The idea is that if the model has memorized the training data, it will be able to perform well even without access to the source text. However, as was recently pointed out, memorization does not have to be perfect to impact generation \cite{chen2024copybenchmeasuringliteralnonliteral}. For instance, even a model that fails a \textit{closed-book} test might have encountered the source text in its training data and may benefit from this during the test when presented with the task along with the source text as the context. Since little is known about memorization in LLMs, the safest evaluation is to test the model on newly produced data (e.g., novels published past the model's cutoff date). However, even this approach necessitates constant updates to the benchmark.

\subsection{Employing LLMs for evaluation and data creation}
\paragraph{Unreliability of LLM-based evaluation:}  Another issue is the way in which model outputs are evaluated. While multiple-choice questions (including True/False evaluations) are relatively straightforward to verify if all plausible answers are correctly marked, they are also likely to be easier for models to answer correctly. Consequently, researchers have developed datasets requiring models to generate long-form answers or perform complex tasks involving long contexts, such as summarization \cite[inter alia]{hudson-al-moubayed-2022-muld, yushi2023longbench, zhang2024inftybench}. These setups are more challenging for the models but also harder to evaluate. Currently, no reliable automatic metric exists for verifying long-form outputs, leading some researchers to use language models, like \gpto{}, for evaluation \cite[e.g.,][]{wang2024leavedocumentbehindbenchmarking}. This method has several flaws. 

First, the model typically evaluates the produced answer against a gold standard, which may penalize valid answers that include additional, relevant information not covered by the gold answer. Second, inconsistent instructions given to the evaluator model can result in variability in its judgments. For example, when the model is asked to assign a score between 0 and 100, with 100 indicating a perfect response \cite[as in][]{wang2024leavedocumentbehindbenchmarking}, intermediate scores may fluctuate due to inconsistent deductions for similar errors across different evaluations. Even when using a simpler scale, such as a 5-point Likert scale, the model’s evaluations can still be unreliable. Unlike a human expert, a language model does not "build experience" through repeated evaluations of similar outputs and cannot refine its understanding or calibration of the scale over time. Human evaluators, in contrast, assess responses not only in relation to a predefined standard but also by drawing on their experience with a variety of previous responses, making their judgments more consistent and dependable. This inherent limitation in language model-based evaluation underscores the difficulty in achieving reliable automated assessment for complex tasks.

\paragraph{Creating test data with language models:} Since employing human annotators to create test instances (e.g., questions about the documents) is time- and resource-intensive (i.e., the annotators have to read long texts),\footnote{Reading an average book takes  at least 8-10 hours.} some researchers use language models to generate these instances. However, this approach has its own issues. For instance, using models such as \gpto{} to generate questions and answers \cite[as in][]{wang2024leavedocumentbehindbenchmarking} may result in flawed test data or test data that is inherently easier for a model from the same family to solve.\footnote{In fact, on a benchmark where Q\&A pairs were generated by \gpto{}, the model achieved one of the highest scores among all tested models \cite{wang2024leavedocumentbehindbenchmarking}.} This becomes even more complex when weaker models are used to generate test examples for relatively complex tasks. For instance, \citet{li2023loogle} employs \textsc{GPT-3.5-Turbo} to generate summaries for a cloze task, where the tested model is supposed to fill in masked named entities based on the source text. However, \textsc{GPT-3.5-Turbo} has been shown to be a poor summarizer, even when summarizing smaller chunks of text \cite{kim2024fables}.

\section{Dataset}
\label{app:dataset_info}

In this section of the appendix we provide more details on our corpus (\S\ref{app_subsec:corpus}) and human annotation efforts (\S\ref{app_sec:claim_collection}).

\subsection{Corpus}
\label{app_subsec:corpus}
\begin{figure}[tbp]
  \includegraphics[width=1\linewidth]{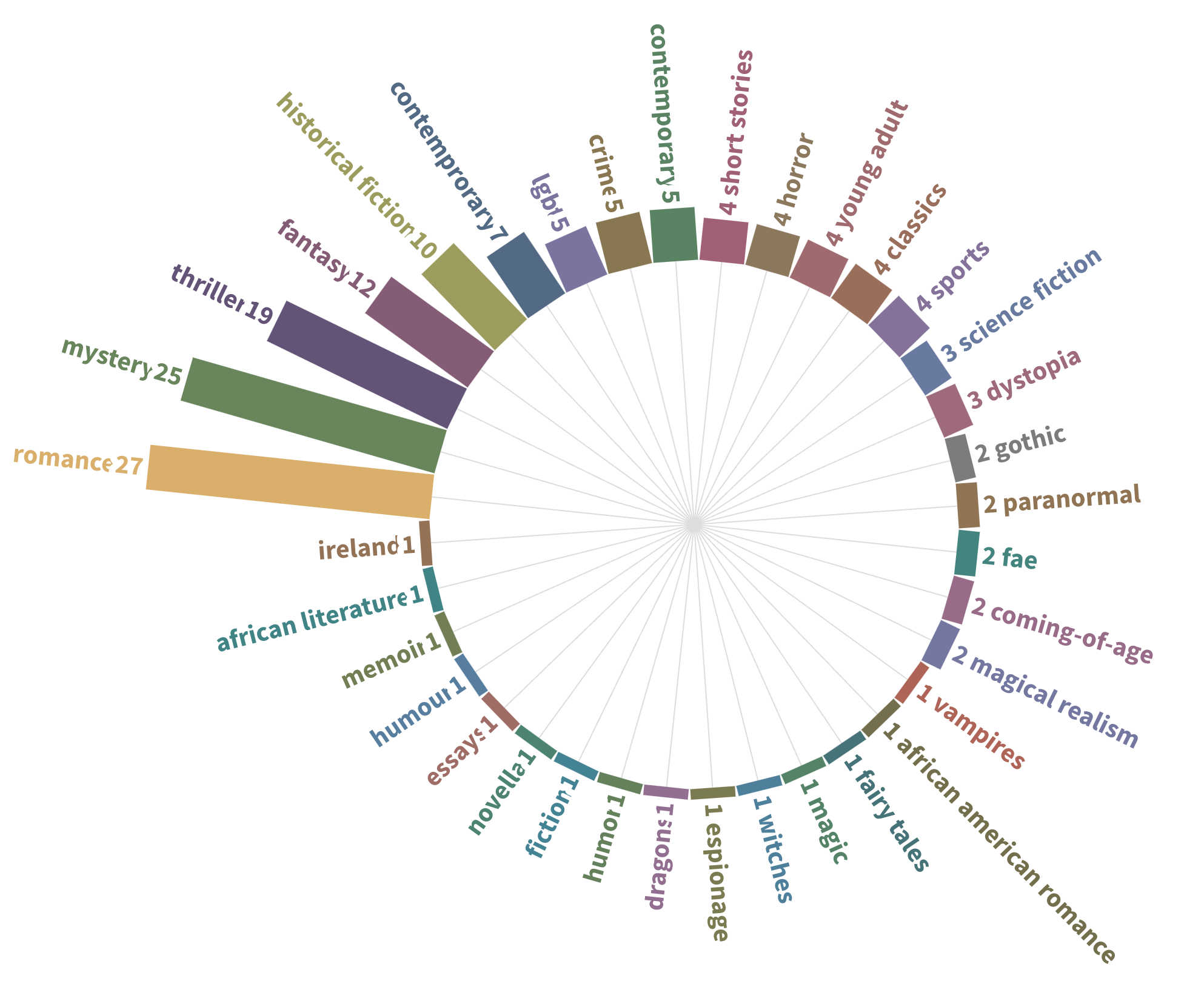} 
  \caption{Genre distribution in \name. As a book can belong to multiple genres, such as fantasy and romance, we allow up to three labels per book.}
  \label{fig:genre_distribution}
\end{figure}

This section provides detailed information about the corpus collected for this study. 
\autoref{tab:list_of_novels} lists all the novels included in the corpus,\footnote{All books in the corpus were purchased by the researchers.} with the genre distribution illustrated in \autoref{fig:genre_distribution}.\footnote{Please note that multiple genre tags are allowed per book, as a book can belong to more than one genre, such as both romance and mystery.} \autoref{tab:list_of_stories} lists all stories included in the story collections. Additionally, we provide the the statistics for books and claims by year of publication in \autoref{app_tab:summary_stats_dataset_by_year}. Finally, we report list the statistics for stories from story collections in \autoref{table:statistics_stories}.

\begin{table*}[h!]
\centering
\resizebox{0.53\width}{!}{%
\begin{tabular}{>{\raggedright\arraybackslash}p{6.5cm} 
                >{\raggedright\arraybackslash}p{3.5cm} 
                >{\centering\arraybackslash}p{1.1cm} 
                >{\raggedright\arraybackslash}p{6.8cm} 
                >{\raggedright\arraybackslash}p{3.3cm} 
                >{\raggedleft\arraybackslash}p{1.2cm} 
                >{\raggedleft\arraybackslash}p{1.2cm}
                >{\centering\arraybackslash}p{1.7cm}
                >{\raggedright\arraybackslash}p{1.2cm}}
                
\toprule
\textsc{Title} \faBook\ & \textsc{Author} \faUser\ & \textsc{Gender} & \textsc{Genre} & \textsc{Pub. Date} \faCalendar\ & \textsc{Tokens} & \textsc{Words} & \textsc{\# Pairs} \faLink\ & \textsc{Lang.} \\
\midrule
\textit{A Haunting on the Hill} & Elizabeth Hand & F & horror, Gothic, paranormal & Oct 3, 2023 & 116,411 & 87,391 &  20 & AmE \\
\textit{Assassin Eighteen} & John Brownlow & \textsc{M} & thriller, crime & Apr 23, 2024 & 126,011 & 96,830 &  15 & CanE \\
\textit{Come and Get It} & Kiley Reid & \textsc{F} & contemporary & Jan 30, 2024 & 141,484 & 104,373 &  30 & AmE \\
\textit{Curse of the Soul Collector} & Cara Blaine & \textsc{F} & fantasy & Sep 20, 2023 & 85,012 & 67,495  & 15 & AmE \\
\textit{Death Comes to Marlow} & Robert Thorogood & M & mystery, fiction, crime & Jan 5, 2023 & 113,544 & 87,844 &  15 & BrE \\
\textit{Divine Rivals} & Rebecca Ross & \textsc{F} & romance, young adult, historical fiction & Apr 4, 2023 & 135,800 & 106,371 & 8 & AmE \\
\textit{Everyone on the train is a suspect} & Benjamin Stevenson & M & mystery & Oct 17, 2023 & 108,673 & 82,296 & 7 & AusE \\
\textit{First Lie Wins} & Ashley Elston & F & thriller, mystery & Jan 2, 2024 & 117,547 & 97,151 &  15 & AmE \\
\textit{Fourth Wing} & Rebecca Yarros & F & fantasy, romance, dragons & May 2, 2023 & 242,683 & 183,789 &  9 & AmE \\
\textit{Funny Story} & Emily Henry & F & romance, contemporary & Apr 23, 2024 & 138,892 & 104,775 &  14 & AmE \\
\textit{Helpless} & Kelby C. Hughes & F & fantasy & Mar 25, 2024 & 65,677 & 53,381 & 15 & AmE \\
\textit{Home Is Where the Bodies Are} & Jeneva Rose & F & thriller, mystery, horror & Apr 30, 2024 & 97,390 & 76,866 & 14 & AmE \\
\textit{House of Flame and Shadow} & Sarah J. Maas & F & fantasy, romance, fae & Jan 30, 2024 & 334,581 & 248,192 & 15 & AmE \\
\textit{How to solve your own murder} & Kristen Perrin & F  & mystery, thriller, crime & Mar 26, 2024 & 130,414 & 104,156 &  10 & AmE\\
\textit{I Hope This Doesn't Find You} & Ann Liang & F & romance, young adult & Feb  6, 2024 & 105,934 & 81,432 & 9 & AusE \\
\textit{Inheritance} & Nora Roberts & F & paranormal, romance, mystery & Nov 21, 2023 & 170,979 & 127,511 & 40 & AmE \\
\textit{Leaving} & Roxana Robinson & F & romance, contemporary & Feb 13, 2024 & 133,317 & 101,039  & 10 & AmE \\
\textit{Long Island} & Colm Toibin & M & historical fiction, Ireland & May 7, 2024 & 103,361 & 84,339 & 10 & IrE \\
\textit{Love, Lies, and Cherry Pie} & Jackie Lou & F & contemporary, romance & May 7, 2024 & 112,179 & 86,994  & 15 & AmE \\
\textit{Monstrous Alterations} & Christopher Barzak & M & short stories, fantasy, LGBT & Sep 8, 2023 & 73,094 & 59,717 &  10 & AmE \\
\textit{Only for the week} & Natasha Bishop & F & African American romance & May 11, 2023 & 85,969 & 68,517 & 5 & AmE \\
\textit{Pet} & Catherine Chidgey & F & thriller, mystery & July 13, 2023 & 124,189 & 93,294 &  46 & NZE \\
\textit{Random in Death} & J. D. Robb & F & mystery, romance, crime & Jan  23, 2024 & 132,107 & 97,172 &  30 & AmE \\

\textit{Romantic Comedy} & Curtis Sittenfeld & F & romance, contemporary & Apr 4, 2023 & 115,004 & 88,775 &  12 & AmE \\
\textit{Roux} & Tamika Christy & F & romance, historical fiction, LGBT & Jan 9, 2024 & 121,364 & 89,221 &  15 & AmE \\
\textit{Ruthless Vows} & Rebecca Ross & F & fantasy, romance, young adult & Dec 26, 2023 & 161,337 & 127,090 &  15 & AmE \\
\textit{Safe and Sound}  & Laura McHugh & F & thriller, mystery & Apr 23, 2024 & 103,054 & 79,243 & 10 & AmE \\
\textit{Same Time Next Year} & Tessa Bailey & F
 & romance, sports, novella & Apr 18, 2023 & 49,156 & 38,023 &  35 & AmE \\
\textit{She Is a Haunting} & Trang Thanh Tran & F & horror, young adult, LGBT & Feb 28, 2023 & 105,378 & 80,291 & 41 & AmE \\
\textit{Six scorched roses} & Carissa Broadbent & F & fantasy, romance, vampires & Mar 21, 2023 & 51,779 & 40,062 &  6 & AmE \\

\textit{The Agency for Scandal} & Laura Wood & F & historical fiction, romance, mystery & Jan 5, 2023 & 115,382 & 90,741 &  18 & AmE \\
\textit{The Atonement Murders} & Jenifer Ruff & F & mystery, thriller & Apr 14, 2023 & 104,258 & 82,134 & 4 & AmE \\
\textit{The Beautiful and the Wild} & Peggy Townsend & F & thriller, mystery & Nov 7, 2023 & 92,424 & 75,908 & 16 & AmE \\
\textit{The book of love} & Kelly Link & F & mystery, magical realism, thriller & Feb 13, 2024 & 272,343 & 209,950 &  10 & AmE \\
\textit{The Bootleggers Daughter} & Nadine Nettman & F & mystery, thriller & May 1 2024 & 97,386 & 75,696 &  15 & AmE \\
The eye of the bedlam bride & Matt Diniman & M & fantasy, science fiction, humor & Jul 2, 2023 & 336,288 & 257,445 & 15 & AmE \\
\textit{The Future} & Naomi Alderman & F & science fiction, dystopia, fantasy & Nov 7, 2023 & 157,019 & 122,342 & 14 & BrE \\
\textit{The Glass Woman} & Alice Mcllroy & F & historical fiction, Gothic, mystery & Dec 19, 2023 & 99,113 & 78,795 & 15 & BrE \\
\textit{The Guest} & Emma Cline & F & mystery, thriller & May 16, 2023 & 89,042 & 68,735 & 20 & AmE \\
\textit{The Hanging City} & Charlie Holmberg & F & fantasy, romance, magic & Aug 1, 2023 & 140,622 & 106,182 &  8 & AmE \\
\textit{The Heiress} & Rachel Hawkins & F & mystery, thriller & Jan  9, 2024 & 94,406 & 73,954 & 15 & AmE \\
\textit{The Husbands} & Holly Gramazio & F & romance, magical realism & Apr 2, 2024
 & 130,017 & 100,432 &  15 & AusE \\
\textit{The Last Murder at the End of the World} & Stuart Turton & M & mystery, thriller, dystopia & May 21, 2024 & 122,364 & 91,280 &  15 & BrE \\
\textit{The Library of Borrowed Hearts} & Lucy Gilmore & F & contemporary, romance & Apr 30, 2024 & 132,205 & 105,259 & 15 & AmE \\
\textit{The Limits: A Novel} & Nell Freudenberger & F & contemporary & Apr 9, 2024 & 131,766 & 102,133 &  15 & AmE \\
\textit{The Marriage Act} & John Marrs & M & thriller, science fiction, dystopia & Jan 19, 2023 & 134,493 & 106,341 &  16 & BrE \\
\textit{The Promise of Tomorrow} & Mary Ellen Taylor & F & contemporary, romance & June 1 2024 & 116,035 & 89,682 &  15 & AmE \\
\textit{The Prospects} & KT Hoffman & F & romance, LGBT, sports & Apr 9, 2024 & 122,720 & 94,591 & 15 & AmE \\
\textit{The Resort} & Sarah Ochs & F & thriller, mystery & Feb 6, 2024 & 124,200 & 99,627 & 15 & AmE \\
\textit{The Spy Coast} & Tess Gerritsen & F & mystery, thriller, espionage & Nov 1, 2023 & 126,906 & 98,326 &  8 & AmE \\
\textit{The Tainted Cup} & Robert Jackson Benett & M & mystery, thriller & Feb 6, 2024 & 162,289 & 119,350 &  10 & AmE \\
\textit{The Teacher} & Freida McFadden & F  & thriller, mystery & Feb  6, 2024 & 110,260 & 88,104 &  15 & AmE \\

\textit{The White Lady} & Jacqueline Winspear & F & historical fiction, mystery & Mar 21, 2023 & 124,585 & 97,094 &  4 & BrE \\
\textit{Viciously Yours} & Jaime Applegate & F & fantasy, romance, fae & Jan 23, 2024 & 85,118 & 65,986 &  7 & AmE \\
\textit{Weyward} & Emilia Hart & F & historical fiction, fantasy, witches & Feb 2, 2023 & 126,501 & 98,842 & 9 & AusE \\
\textit{White Cat, Black Dog} & Kelly Link & F & short stories, horror, fairy tales & Mar 2, 2023 & 92,826 & 74,154 & 10 & AmE \\
\textit{Wildfire} & Hannah Grace & F & romance, sports, contemporary & Oct 3, 2023 & 138,441 & 109,641 &  5 & BrE \\
\textit{Yellowface} & R.F Kuang & F & thriller, contemporary, mystery & May 25, 2023 & 113164 & 87648 &  12 & AmE \\
\textit{You Should Be So Lucky} & Cat Sebastian & F & romance, LGBT, sports & May 7, 2024 & 140,422 & 109,764 & 15 & AmE \\
\textit{You, Again} & Kate Goldbeck & F & romance, contemporary & Sep 12, 2023 & 128,445 & 96,681 &  7 & AmE \\
\textit{Yours truly} & Abby Jimenez & F & romance, contemporary & April 11, 2023 & 134,609 & 105,968 & 10 & AmE \\
\midrule
\textit{Quietly Hostile} & Samantha Irby & F & essays, humour, memoir & May 16, 2023 & 95,842 & 76,424 & 14 & AmE \\
\midrule
\textit{Anne of Green Gables} & L. M. Montgomery & F & classics, coming-of-age, historical & Jan 1, 1908 & 129,908 & 102,366 & 15 & CanE \\
\textit{Little Women} & Louisa May Alcott & F & classics, coming-of-age, historical, romance & Sep 30, 1868 & 235,118 & 185,930 &  15 & AmE \\
\textit{The Adventures of Sherlock Holmes} & Arthur Conan Doyle & M & classics, short stories,  mystery, crime & Oct 14, 1892 & 129,293 & 104,434 &  18 & BrE \\
\textit{The Great Gatsby} & F. Scott Fitzgerald & M & classics, historical fiction, romance & Apr 10, 1925 & 61,689 & 48,187 &  15 & AmE \\
\bottomrule
\end{tabular}
}
\caption{List of novels included in \name. The upper row presents all recent fictional books included in the data. The middle row displays the essay collections, while the lower row shows classical books included in the data. The genre is provided as listed on \texttt{GoodReads} (\url{https://www.goodreads.com/}). The language is indicated based on the author's native language or, for non-native English authors, the primary language of the country where they spent most of their time. The token count is reported as per \texttt{tiktoken} tokenization with \texttt{cl100k} encoding while the word count was determined by a \texttt{whitespace} split.}
\label{tab:list_of_novels}
\end{table*}

\begin{table*}[h!]
\centering
\resizebox{0.5\width}{!}{%
\begin{tabular}{>{\raggedright\arraybackslash}p{0.7cm} 
                >{\raggedright\arraybackslash}p{6cm} 
                >{\raggedright\arraybackslash}p{3.5cm} 
                >{\raggedright\arraybackslash}p{7.5cm} 
                >{\raggedright\arraybackslash}p{1.4cm} 
                >{\raggedleft\arraybackslash}p{2cm} 
                >{\raggedleft\arraybackslash}p{2.8cm}
                >{\raggedright\arraybackslash}p{2.2cm}
                >{\raggedright\arraybackslash}p{2.4cm}}
                
\toprule
\textsc{ID} & \textsc{Collection} \faBook\ & \textsc{Author} \faUser\ & \textsc{Story Title} & \textsc{\# Pairs} & \textsc{Story Tokens} & \textsc{Collection Tokens} & \textsc{Location}  & \textsc{Pub. Date} \faCalendar\ \\
\midrule
1 & A Kind of Madness & Uche Okonkwo & Nwunye Belgium & 2 & 10299 & 80946 & \textit{beginning} & Apr 16, 2024 \\
2 & A Kind of Madness & Uche Okonkwo & Shadow & 3 & 8549 & 80946 & \textit{beginning} & Apr 16, 2024 \\
3 & A Kind of Madness & Uche Okonkwo & Debris & 3 & 1822 & 80946 & \textit{beginning} & Apr 16, 2024 \\
4 & A Kind of Madness & Uche Okonkwo & Long Hair & 3 & 3464 & 80946 & \textit{beginning} & Apr 16, 2024 \\
5 & A Kind of Madness & Uche Okonkwo & Animals & 3 & 8193 & 80946 & \textit{beginning} & Apr 16, 2024 \\
6 & A Kind of Madness & Uche Okonkwo & Milk and Oil & 3 & 11829 & 80946 & \textit{middle} & Apr 16, 2024 \\
7 & A Kind of Madness & Uche Okonkwo & The Harvest & 2 & 7145 & 80946 & \textit{middle} & Apr 16, 2024 \\
8 & A Kind of Madness & Uche Okonkwo & Eden & 2 & 7408 & 80946 & \textit{middle} & Apr 16, 2024 \\
9 & A Kind of Madness & Uche Okonkwo & The Girl Who Lied & 3 & 12834 & 80946 & \textit{end} & Apr 16, 2024 \\
10 & A Kind of Madness & Uche Okonkwo & Burning & 3 & 9524 & 80946 & \textit{end} & Apr 16, 2024 \\
1 & The Adventures of Sherlock Holmes & Arthur Conan Doyle & A Scandal in Bohemia & 2 & 10715 & 129293 & \textit{beginning} & Oct 14, 1892 \\
2 & The Adventures of Sherlock Holmes & Arthur Conan Doyle & The Red Headed League & 1 & 11369 & 129293 & \textit{beginning} & Oct 14, 1892 \\
3 & The Adventures of Sherlock Holmes & Arthur Conan Doyle & A Case of Identity & 1 & 8715 & 129293 & \textit{beginning} & Oct 14, 1892 \\
4 & The Adventures of Sherlock Holmes & Arthur Conan Doyle & The Boscombe Valley Mystery & 3 & 11748 & 129293 & \textit{beginning} & Oct 14, 1892 \\
5 & The Adventures of Sherlock Holmes & Arthur Conan Doyle & The Five Orange Pips & 2 & 9076 & 129293 & \textit{beginning} & Oct 14, 1892 \\
6 & The Adventures of Sherlock Holmes & Arthur Conan Doyle & The Man with the Twisted Lip & 1 & 11401 & 129293 & \textit{middle} & Oct 14, 1892 \\
7 & The Adventures of Sherlock Holmes & Arthur Conan Doyle & The Adventure of Blue Carbuncle & 1 & 9982 & 129293 & \textit{middle} & Oct 14, 1892 \\
8 & The Adventures of Sherlock Holmes & Arthur Conan Doyle & The Adventure of the Speckled Band & 2 & 12070 & 129293 & \textit{middle} & Oct 14, 1892 \\
9 & The Adventures of Sherlock Holmes & Arthur Conan Doyle & The Adventures of the Engineers Thumb & 1 & 10227 & 129293 & \textit{middle} & Oct 14, 1892 \\
10 & The Adventures of Sherlock Holmes & Arthur Conan Doyle & The Adventure of the Noble Bachelor & 1 & 10086 & 129293 & \textit{end} & Oct 14, 1892 \\
11 & The Adventures of Sherlock Holmes & Arthur Conan Doyle & The Adventure of the Beryl Coronet & 2 & 11776 & 129293 & \textit{end} & Oct 14, 1892 \\
12 & The Adventures of Sherlock Holmes & Arthur Conan Doyle & The Red Adventure of the Copper Beeches & 1 & 12331 & 129293 & \textit{end} & Oct 14, 1892 \\
1 & Quietly Hostile & Samantha Irby & i like it! & 0 & 1783 & 95842 & \textit{beginning} & May 16, 2023 \\
2 & Quietly Hostile & Samantha Irby & the last normal day & 1 & 4188 & 95842 & \textit{beginning} & May 16, 2023 \\
3 & Quietly Hostile & Samantha Irby & david matthews’s greatest romantic hits & 1 & 4563 & 95842 & \textit{beginning} & May 16, 2023 \\
4 & Quietly Hostile & Samantha Irby & chub street diet & 1 & 4805 & 95842 & \textit{beginning} & May 16, 2023 \\
5 & Quietly Hostile & Samantha Irby & my firstborn dog & 1 & 4672 & 95842 & \textit{beginning} & May 16, 2023 \\
6 & Quietly Hostile & Samantha Irby & body horror! & 1 & 4798 & 95842 & \textit{beginning} & May 16, 2023 \\
7 & Quietly Hostile & Samantha Irby & two old nuns having amzing [sic] lesbian sex & 1 & 7138 & 95842 & \textit{beginning} & May 16, 2023 \\
8 & Quietly Hostile & Samantha Irby & qvc, ilysm & 1 & 5974 & 95842 & \textit{beginning} & May 16, 2023 \\
9 & Quietly Hostile & Samantha Irby & superfan!!!!!!! & 1 & 11397 & 95842 & \textit{middle} & May 16, 2023 \\
10 & Quietly Hostile & Samantha Irby & i like to get high at night and think about whales & 0 & 702 & 95842 & \textit{middle} & May 16, 2023 \\
11 & Quietly Hostile & Samantha Irby & oh, so you actually don’t wanna make a show about a horny fat bitch with diarrhea? okay! & 1 & 12351 & 95842 & \textit{middle} & May 16, 2023 \\
12 & Quietly Hostile & Samantha Irby & what if i died like elvis & 1 & 9068 & 95842 & \textit{middle} & May 16, 2023 \\
13 & Quietly Hostile & Samantha Irby & shit happens & 1 & 4960 & 95842 & \textit{end} & May 16, 2023 \\
14 & Quietly Hostile & Samantha Irby & food fight & 0 & 1508 & 95842 & \textit{end} & May 16, 2023 \\
15 & Quietly Hostile & Samantha Irby & o brother, where art thou? & 1 & 4865 & 95842 & \textit{end} & May 16, 2023 \\
16 & Quietly Hostile & Samantha Irby & how to look cool in front of teens? & 1 & 5723 & 95842 & \textit{end} & May 16, 2023 \\
17 & Quietly Hostile & Samantha Irby & we used to get dressed up to go to red lobster & 1 & 5558 & 95842 & \textit{end} & May 16, 2023 \\
18 & Quietly Hostile & Samantha Irby & please invite me to your party & 0 & 2049 & 95842 & \textit{end} & May 16, 2023 \\
1 & White Cat, Black Dog & Kelly Link & The White Cat's Divorce & 2 & 11697 & 92826 & \textit{beginning} & Mar 2, 2023 \\
2 & White Cat, Black Dog & Kelly Link & Prince Hat Underground & 2 & 19724 & 92826 & \textit{beginning} & Mar 2, 2023 \\
3 & White Cat, Black Dog & Kelly Link & The White Road & 2 & 11770 & 92826 & \textit{middle} & Mar 2, 2023 \\
4 & White Cat, Black Dog & Kelly Link & The Girl Who Did Not Know Fear & 1 & 8606 & 92826 & \textit{middle} & Mar 2, 2023 \\
5 & White Cat, Black Dog & Kelly Link & The Game of Smash and Recovery & 0 & 6869 & 92826 & \textit{middle} & Mar 2, 2023 \\
6 & White Cat, Black Dog & Kelly Link & The Lady and the Fox & 1 & 12504 & 92826 & \textit{middle} & Mar 2, 2023 \\
7 & White Cat, Black Dog & Kelly Link & Skinder's Veil & 2 & 21734 & 92826 & \textit{end} & Mar 2, 2023 \\
\bottomrule
\end{tabular}
}
\caption{List of stories included in the collections. We provide token counts for both the entire collection and individual story. The location was determined by dividing collection length into three parts, where the stories which begin in the first part are marked as \textit{beginning}, stories which begin in the second part are marked as \textit{middle}, and stories beginning in the third part are marked as \textit{end}.}
\label{tab:list_of_stories}
\end{table*}

\begin{table}[t]
\centering
\begin{tabular}{cc}
\hline
 & \textsc{Stories} \\
\toprule
\textsc{Mean} & 8,501.5 \\
\textsc{Max} & 21,734 \\
\textsc{Min} & 702 \\
\textsc{Std. Dev.} & 4,355.7 \\
\bottomrule
\end{tabular}
\caption{Statistics of stories in story collections (in tokens). Note that there is 7--18 stories per collection.}
\label{table:statistics_stories}
\end{table}

\begin{table*}[t]
\centering
\footnotesize
\resizebox{0.8\textwidth}{!}{%
\begin{tabular}{lcccccc}
\hline
\addlinespace
& \multicolumn{3}{c}{\textbf{Books } \includegraphics[height=1.1em]{figures/books-icon.png}} & \multicolumn{3}{c}{\textbf{Claim [Pairs] } \includegraphics[height=1em]{figures/annot-icon.png}} \\
\cmidrule(r){2-4} \cmidrule(lr){5-7}
 & {\renewcommand{\arraystretch}{1}\begin{tabular}[c]{@{}c@{}}\textbf{ 2024 }\\\textit{(n=33) }\vspace{.0em}\end{tabular}} & {\renewcommand{\arraystretch}{1}\begin{tabular}[c]{@{}c@{}}\textbf{ 2023 }\\
 \textit{(n=30) }\vspace{.0em}\end{tabular}} & {\renewcommand{\arraystretch}{1}\begin{tabular}[c]{@{}c@{}}\textbf{ Classics }\\\textit{(n=4) }\vspace{.0em}\end{tabular}} & 
 {\renewcommand{\arraystretch}{1}\begin{tabular}[c]{@{}c@{}}\textbf{ 2024 }\\
 \textit{(n=898 [449]) }\vspace{.0em}\end{tabular}} & {\renewcommand{\arraystretch}{1}\begin{tabular}[c]{@{}c@{}}\textbf{ 2023 }\\
 \textit{(n=978 [489]) }\vspace{.0em}\end{tabular}} & {\renewcommand{\arraystretch}{1}\begin{tabular}[c]{@{}c@{}}\textbf{ Classics }\\\textit{(n=126 [63]) }\vspace{.0em}\end{tabular}}\\ 
\hline
\addlinespace
\multicolumn{2}{c}{\textsc{Tokens} (\texttt{tiktoken})} & & &  & & \\
\cmidrule(r){1-2} 
Mean  & 129,526    & 123,908 & 139,002 & 23.51 & 22.75 &  24.71  \\
St. dev.  & 52,112    & 52,176    & 71,630 & 7.17 & 7.94 &  7.84 \\
Max      &  334,581  & 336,288   & 235,118 &  63 & 57 &  45 \\
Min       & 65,677 & 49,156   & 61,689 & 5 & 8 &  10 \\ 
\midrule
\multicolumn{2}{c}{\textsc{Words} (\texttt{whitespace})} & & & & & \\
\cmidrule(r){1-2}
Mean  & 99,753    & 96,117 & 110,229 & 18.44 & 17.85 &  20.06 \\
St. dev.  & 38,463    & 39,388   & 56,790 & 6.15 & 6.68 & 6.97 \\
Max      & 248,192  & 257,445   & 185,930 & 57 & 46 &  39  \\
Min      & 53,381 & 38,023 & 48,187 & 4 & 5 &  7 \\ 
\hline
\end{tabular}
}
\caption{Number of tokens and words across books and claims for books published in 2024, 2023, and classics. The number of claim pairs is indicated in [] square brackets.}
\label{app_tab:summary_stats_dataset_by_year}
\end{table*}

\subsection{Human annotation efforts}
\label{app_sec:claim_collection}

In this section of the appendix we provide additional details about the data collection pipeline.

\paragraph{Annotators:} The annotations were done by 18 annotators recruited on Upwork (all female) and 5 volunteer annotators, including three of the authors (four females and one male). All but four annotators are native English speakers from either the US or UK. All annotators hold university degrees, with one being a former college English professor and an author, and one holding a Ph.D. in English literature. As frequent readers, some annotators received advanced reader copies of the books, enabling the annotation of works, which were unpublished at the time of data collection.

\begin{figure*}[t]
    \centering
    \begin{subfigure}[b]{0.45\textwidth}
        \includegraphics[width=\linewidth]{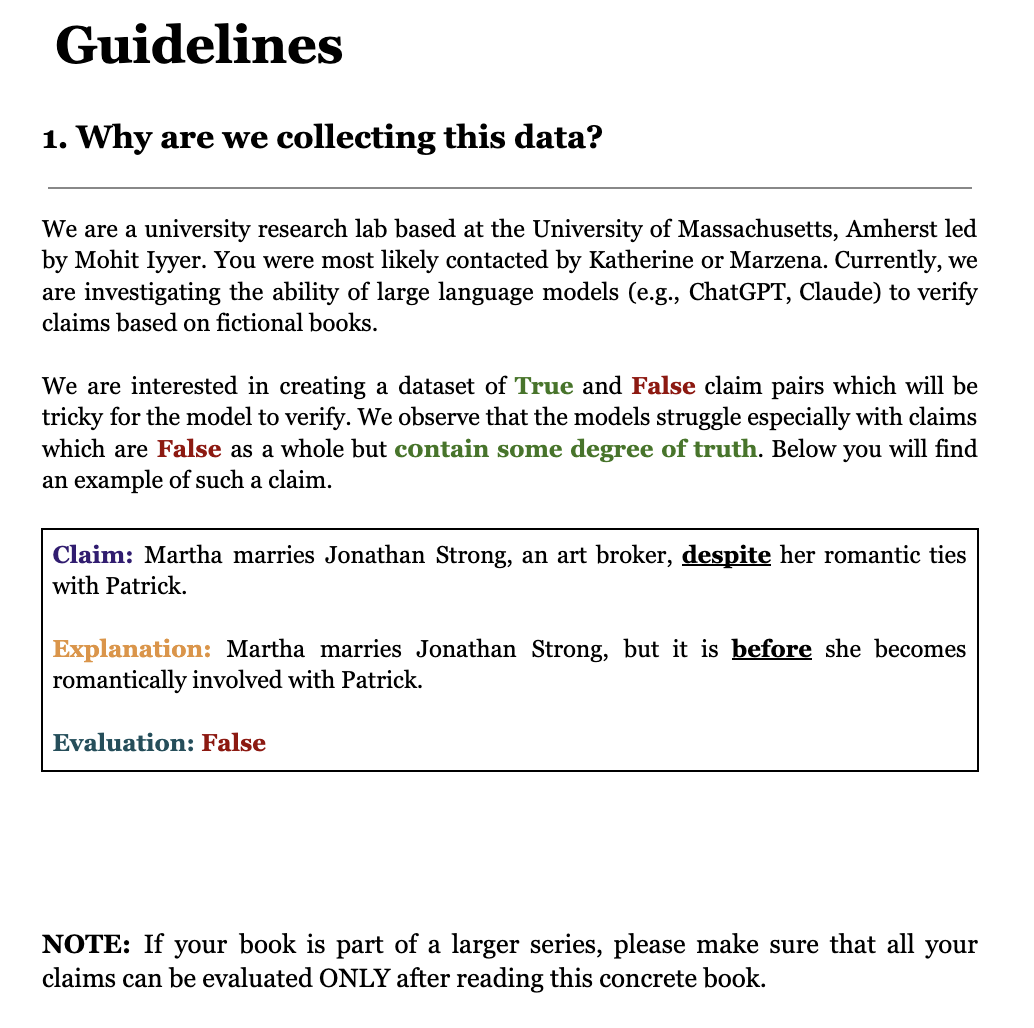}
        \caption{Guidelines: Page 1}
    \end{subfigure}
    \hfill
    \begin{subfigure}[b]{0.45\textwidth}
        \includegraphics[width=\linewidth]{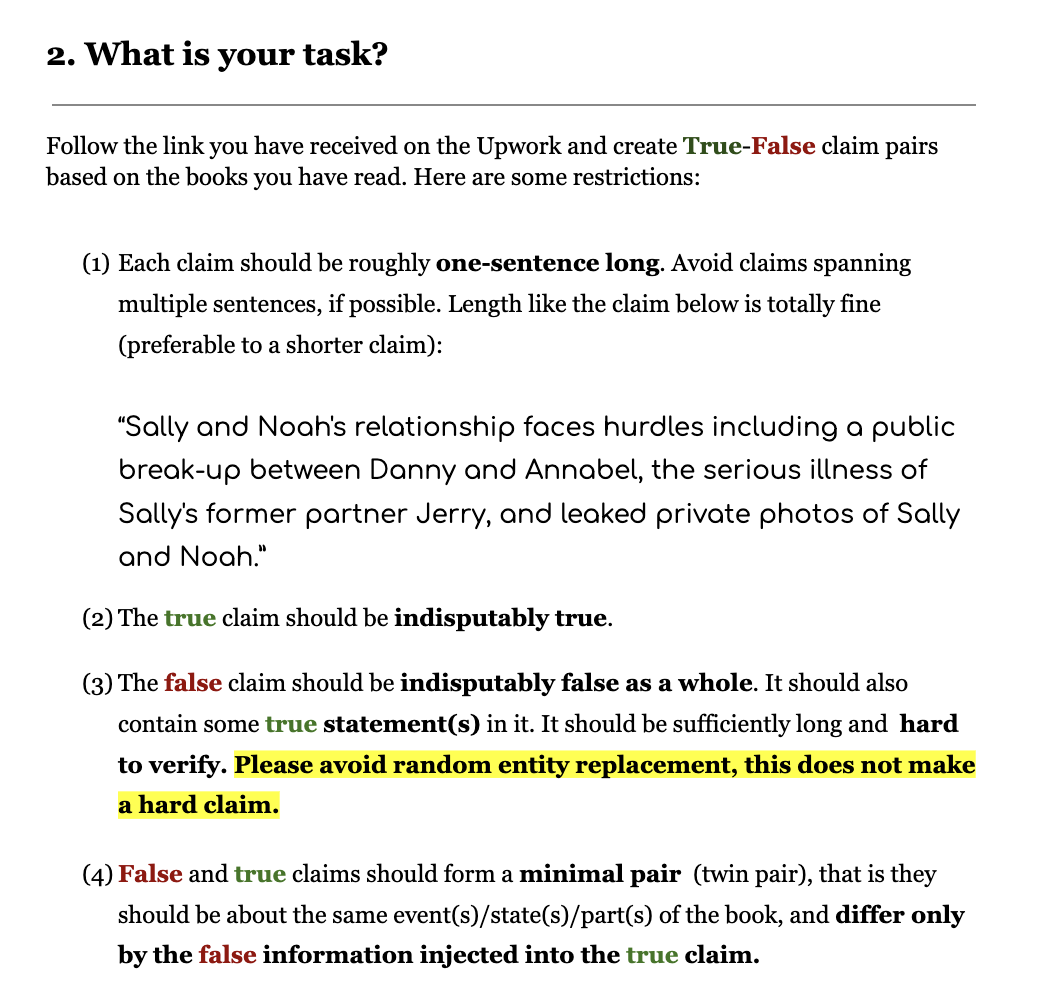}
        \caption{Guidelines: Page 2}
    \end{subfigure}
    \caption{Guidelines provided to the annotators for the annotation task. The annotators were also provided additional examples and guidance during the data collection process.}
    \label{fig:guidelines_for_annotators}
\end{figure*}

\begin{figure}[tbp]
  \includegraphics[width=1\linewidth]{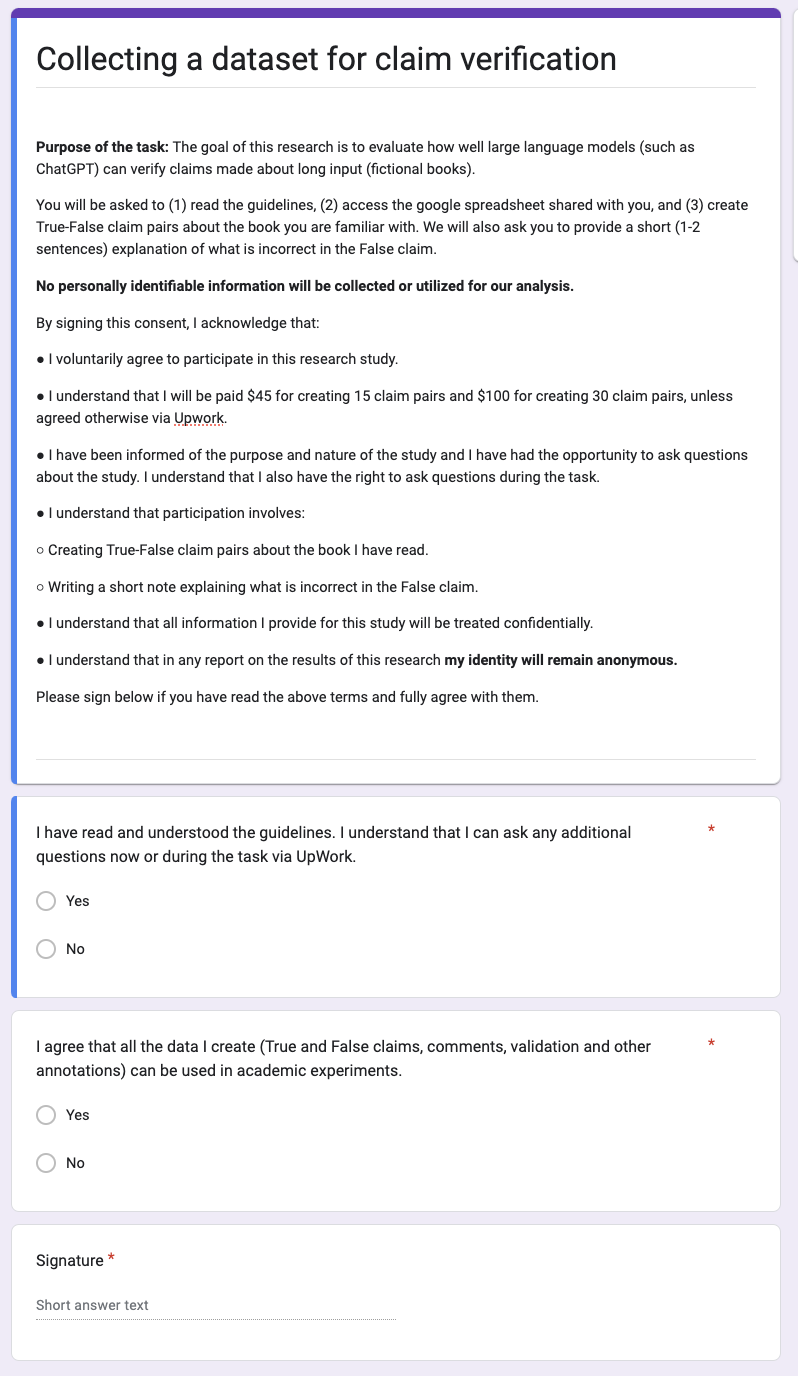} 
  \caption{Consent form which the annotators were asked to sign via \texttt{GoogleForms} before collecting the data.}
  \label{fig:consent_form}
\end{figure}

\paragraph{Collecting True/False claim pairs:} Before starting the annotation task, all annotators were required to read the guidelines  (\autoref{fig:guidelines_for_annotators}) and sign a consent form (\autoref{fig:consent_form}). The collected claims underwent a rigorous review process, being checked at least three times by two of the authors: initially during the writing stage and later independently during the final quality control phase. During this phase, we also proofread all claims and contacted annotators if anything was unclear. 

Collecting each pair of annotations typically required additional communication with the annotators, resulting in approximately 160 hours of work from each of the two authors involved in this process. Based on self-reported time and records in spreadsheets, we estimate that the annotators were able to produce between 6 and 10 claim pairs per hour.\footnote{The annotators we also asked to provide written justification as to why the true claim is true and the false claim is false relating to the events in the book.}

Overall, we collected about 15 claim pairs per book. While we initially aimed to collect more claim pairs per book, we observed that creating meaningful and challenging pairs becomes significantly more difficult beyond the first 10-15 pairs, though this number may vary slightly depending on the book. For some books, we collected fewer than 10 pairs due to the unavailability of annotators to create more pairs.

\paragraph{Advantages of a minimal pair design}
We employ minimal pairs for two main reasons. Firstly, it ensures data quality, allowing us to verify the false claim against its true counterpart easily and identify cases where both claims were too similar (i.e., the false claim could potentially be true) or subjective. Secondly, we require models to correctly label \textit{both} claims in a pair, which minimizes the chances of crediting them for correct predictions made "for the wrong reason" without proper utilization of the context.

Indeed, the results presented in the paper indicate that merely having a balanced set of labels is insufficient for accurately evaluating the models  (see \S\ref{sec:analysis} and examples in \autoref{fig:pair_failure}). For example, using balanced but unpaired data might lead us to conclude that \gpto{} achieves an overall accuracy of 76.7\% (77.5\% for True claims and 75.9\% for False claims; see \autoref{tab:model_acc_on_true_and_false}). However, these results are misleading because the model fails to validate \textit{both} claims correctly in approximately 20\% of cases, resulting in an actual accuracy of 55.8\%. This is significantly lower than the human accuracy of 96.9\% reported in \S\ref{sec:data_methods},  highlighting a substantial performance gap for the best model to bridge.

\paragraph{Quality control:} To ensure the quality of the annotations, we reannotated a subset of the data, consisting of 128 claims made about 6 books (76 claim pairs). For this task, we hired annotators who had read the same books.\footnote{We hired one new annotator per book which resulted in two annotations per book, the original annotator (author of the claims and labels) and the new annotator.} All but one of these annotators were also part of the original annotation team but had worked on different books previously.\footnote{One person who did not do the original annotation was a volunteer who have also read of the annotated books. This person read the book for this task specifically and performed their annotations right after finishing the book.} They were provided with \textit{the same} instructions as the prompt given to the model (\autoref{tab:prompt_template_claim_eval}), with the only modification being the phrase "based on the context provided" changed to "based on your book." All annotations were performed using \texttt{GoogleSpreadsheets}. 

Overall, the annotators agreed with the original annotator on 148 out of 152 labels (72 out of 76 claim pairs). We then reviewed the two pairs where the annotators disagreed. In one pair, the disagreement was due to a sentence-level detail about whether the main characters were engaged for two or three years. The characters were together for three years but engaged for only two, a detail mentioned in the book only once. The second annotator \textit{incorrectly} annotated the claim of three years as true and the two years claim as false leading to disagreement. In the second case, the claim was about "The Great Gatsby," where the initial annotator confused parts of the book with parts of the movie. Importantly, none of the core test books have movies made based on them as they were published within the last year, so this situation was specific to this classic novel. We have since reviewed all the classic claims again to ensure this issue does not recur.

\begin{table}[t]
\centering
\footnotesize
\resizebox{0.9\columnwidth}{!}{%
\begin{tabular}{lccc}
\toprule
\textsc{Scope} & \textsc{Pairs (\%)} & \textsc{True (\%)} & \textsc{False (\%)} \\
\midrule
\faBook \textsc{Global} & 47.9\% & 38.0\% & 47.1\% \\
\faFileTextO \textsc{Passage} & 39.7\% & 43.8\% & 39.7\% \\
\faAlignLeft \textsc{Sentence} & 12.4\% & 18.2\% & 13.2\% \\
\bottomrule
\end{tabular}
}
\caption{Percentage of pairs by scope in the annotated subset of data (\textit{global}, \textit{passage}, \textit{sentence}). Additionally, we report the percentage of \textit{claims} with specific labels for both True and False claims.}
\label{tab_app:evidence_scope_details}
\end{table}

\begin{figure}[tbp]
  \includegraphics[width=1\linewidth]{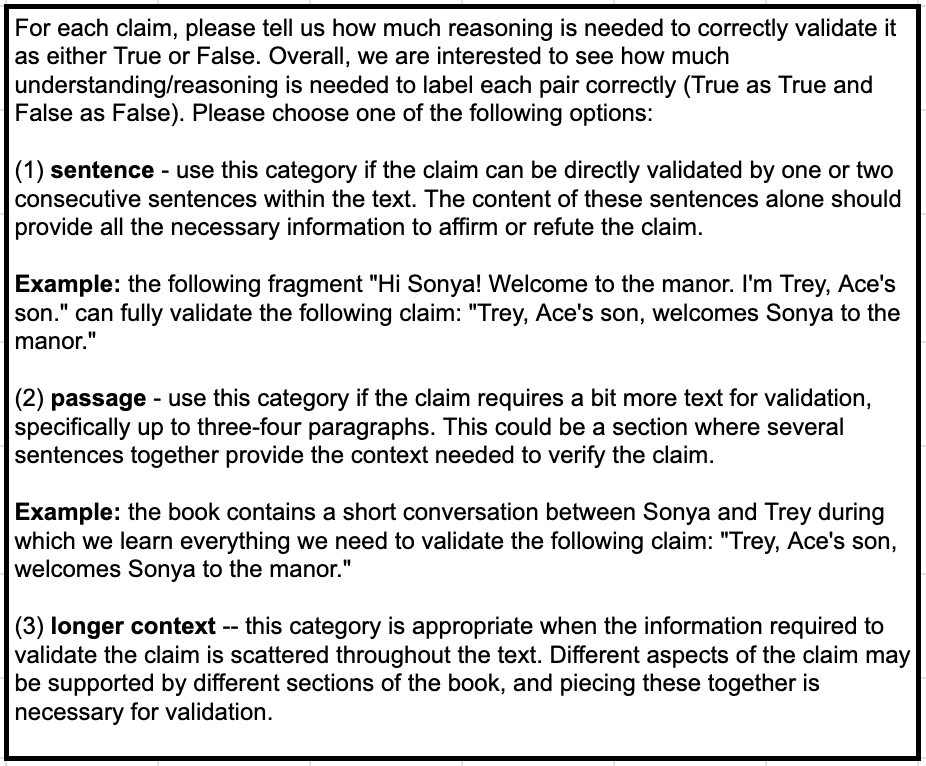} 
  \caption{Instructions given to the annotators for the annotation of claim scope.}
  \label{fig:scope_instructions_annotators}
\end{figure}

\paragraph{Evidence scope:} 
For this task, we asked four annotators to annotate the scope of the evidence for two of their books each, resulting in annotations for 121 claim pairs from 8 books (approximately 15 claim pairs per book; see \autoref{fig:scope_instructions_annotators} for instructions given to the annotators). The annotators first annotated the scope for each claim in the pair, and then the maximum scope was taken for the given pair.\footnote{For instance, if one claim in a pair is labeled as "passage" and the other as "global," the resulting scope for the entire pair would be "global." This is because we require the models to validate \textit{both} claims in the pair correctly, meaning the models would have to reason over both "passage" and "global" contexts for this specific pair. It is worth noting that the labels for both claims in the pair matched 87.6\% of the time.} Overall, 12.4\% of pairs were labeled as requiring only one to two sentences to validate, 39.7\% of pairs were labeled as requiring a longer passage, and 47.9\% were labeled as requiring global reasoning. The distribution of labels for specific claims is comparable (see \autoref{tab_app:evidence_scope_details}).

\paragraph{Note on evidence location:} Designing a realistic experiment to verify complex claims with evidence at different depths of a book is challenging. Unlike the "needle-in-a-haystack" experiment, where the "needle" can be easily placed in different parts of the (often unrelated) context, our task involves realistic claims about the book’s content, making it more problematic to control for various confounders. For instance, a claim about events at the beginning of the book might be inherently easier to verify. Similarly, it is difficult to assert with certainty that the only piece of evidence necessary to validate the claim is present at a specific depth in the book, as there could be corroborating evidence prior or later that aids verification (e.g., hints about the killer’s identity), which the human reader may (reasonably) not pay attention to. Hence, we do not attempt to annotate the evidence location. Instead, we use story collections, where we can precisely identify which story the claim pertains to and where the story is located within the collection. We present the results of this experiment both in the main body of this paper (\S\ref{sec:analysis}) and later in this appendix (\S\ref{app_sec:results_info}).

\section{Methods}
\label{app_sec:methods_info}
In this section we provide additional details about our evaluation methodology.

\paragraph{Models:} We evaluate 6 closed-source and 5 open-source long-context models (i.e., models with context window of at least 128k tokens): 

\begin{itemize}
    \item \faLock\ \textbf{Closed-source models:} \gpto{} \cite{openai-gpt4o}, \gptturbo{} \cite{openai2024gpt4}, \claude{} \cite{anthropic-claude},\sonnet{} \cite{anthropic-sonnet}, \geminipro{} \cite{geminiteam2024gemini}, \geminiflash{} \cite{geminiteam2024gemini}. All models were accessed via provider's API.\footnote{\claude's generations were done partially employing \texttt{anthropic} API and partially utilizing \texttt{vertex-ai} due to the rate limit. All \sonnet{} generations were done utilizing \texttt{vertex-ai}.}
    \item \faUnlock\ \textbf{Open-weight models:} \comr{} \cite{commandr}, \comrplus{} \cite{command-r-plus}, \gemma{} \cite{gemma}, \phimodel{} \cite{abdin2024phi3}, and \longllama{} \cite{longllama-tworkowski2023focused}. \comr{} and \comrplus{} we accessed via \texttt{cohere} API. Other models were run on up to three A100 80GB GPUs. For comparison, in our stories experiment, we also tested \mixtral{} (65k context window) and \qwen{} (32k context window). While their context windows are too short to process entire books, they can handle stories from our collections. Both models were run using the \texttt{together.ai} API.
    \item \faDatabase\ \textbf{Retrieval pipeline:} We also implement a retrieval pipeline (\bm{}), where we first retrieve the evidence (top 5, 25, and 50 paragraphs) and then prompt \gpto{} for evaluation.
\end{itemize}

\begin{table*}[t!]
\centering
\footnotesize
\resizebox{0.9\textwidth}{!}{%
\begin{tabular}{l c l c  c}
\toprule
\textsc{Model} & \textsc{Context} & \textsc{Avail.} & \textsc{Checkpoints}& \textsc{\# Param}  \\
\midrule
\href{https://platform.openai.com/docs/models}{\gpto} & 128k & \faLock & \texttt{gpt-4o-2024-05-13} & \includegraphics[height=1.1em]{figures/shrug_kaomoji.png} \\
\href{https://platform.openai.com/docs/models}{\gptturbo} & 128k & \faLock & \texttt{gpt-4-turbo-2024-04-09} & \includegraphics[height=1.1em]{figures/shrug_kaomoji.png} \\
\href{https://www.anthropic.com/news/claude-3-family}{\claude} & 200k & \faLock & \texttt{claude-3-opus-20240229} & \includegraphics[height=1.1em]{figures/shrug_kaomoji.png} \\
\href{https://www.anthropic.com/news/claude-3-5-sonnet}{\sonnet} & 200k & \faLock & \texttt{claude-3-5-sonnet@20240620} & \includegraphics[height=1.1em]{figures/shrug_kaomoji.png} \\
\href{https://cloud.google.com/vertex-ai/generative-ai/docs/model-reference/gemini}{\geminipro} & 1M & \faLock & \texttt{gemini-1.5-pro-preview-0514} & \includegraphics[height=1.1em]{figures/shrug_kaomoji.png}  \\
\href{https://cloud.google.com/vertex-ai/generative-ai/docs/model-reference/gemini}{\geminiflash} & 1M & \faLock & \texttt{gemini-1.5-flash-preview-0514} & \includegraphics[height=1.1em]{figures/shrug_kaomoji.png}  \\
\midrule
\href{https://huggingface.co/CohereForAI/c4ai-command-r}{\comr} & 128k & \faUnlock & \texttt{c4ai-command-r-v01} &35B  \\
\href{https://huggingface.co/CohereForAI/c4ai-command-r-plus}{\comrplus} & 128k & \faLock & \texttt{c4ai-command-r-plus} & 104B  \\
\href{https://huggingface.co/mustafaaljadery/gemma-2B-10M}{\gemma{}} & 10M & \faUnlock & \texttt{gemma-2b-10m} &2B \\
\href{https://huggingface.co/microsoft/Phi-3-mini-128k-instruct}{\phimodel{}} & 128k & \faUnlock & \texttt{Phi-3-mini-128k-instruct} &3.8B  \\
\href{https://huggingface.co/syzymon/long_llama_3b_instruct}{\longllama{}} & 256k & \faUnlock & \texttt{long\_llama\_3b\_instruct} &3B   \\
\midrule
\href{https://huggingface.co/mistral-community/Mixtral-8x22B-v0.1}{\mixtral{}} & 65k & \faUnlock & \texttt{Mixtral-8x22B-Instruct-v0.1} & 141B   \\
\href{https://huggingface.co/Qwen/Qwen2-72B-Instruct}{\qwen{}} & 32k\tablefootnote{This version can also be extended to 128k using \textsc{YaRN} \cite{peng2023yarnefficientcontextwindow}.} & \faUnlock & \texttt{Qwen2-72B-Instruct} & 72B   \\

\bottomrule
\end{tabular}
}
\caption{Evaluated models: the upper part displays all closed-source models, while the lower part lists all open-weight models. We also provide details of two shorter context models, which were tested on story-length inputs for comparison.}
\label{app_tab:models}
\end{table*}

\paragraph{Inference:} We generate the labels by prompting the models with the entire book and one claim at time for verification. All closed-source models were prompted using the provider's API, while open-weight models were run on up to three A100 80GB GPUs, excluding \comr{} and \comrplus{}, which were accessed using the \texttt{cohere} API. Additionally, we implement a \bm{} retrieval pipeline where we first retrieve the evidence from the book and then verify the claim based on the retrieved evidence with \gpto{}.\footnote{The estimated cost for running all data with each model is as follows: \gpto{} \$640 USD, \gptturbo{} \$1,280 USD, \claude{} \$3,350 USD, \sonnet{} \$670 USD, \geminipro{} 1,385 USD, \comrplus{} \$450 USD, \comr{} \$74 USD, \bm{} pipeline \$330 USD. Note that test runs resulted in small additional costs.}

\paragraph{Prompts:} All models were prompted with the prompt presented in \autoref{tab:prompt_template_claim_eval}. As we notice that the open-source models did not follow the instructions well, we also tested them with a simplified prompt as shown in \autoref{tab:prompt_template_claim_eval_simple}. Finally, for \bm{} we employed the prompt in \autoref{tab:prompt_template_claim_eval_bm25}. All generations are restricted to 800 tokens.\footnote{For some open-source models we noticed that the generation are plagued by repetitions and reduced this limit to 600 tokens.}

During a pilot study, we first tested three different types of prompts on a subset of data (10 books, 176 claim pairs):  
\begin{itemize}
    \item \textbf{Answer-only:} prompting the model for the answer only (\autoref{tab:prompt_template_claim_eval_devset_answer})
    \item \textbf{Answer-then-explanation:} prompting the model for the answer followed by an explanation (\autoref{tab:prompt_template_claim_eval_devset_answer_explain})
    \item \textbf{Explanation-then-answer:} prompting the model for the explanation followed by the answer (\autoref{tab:prompt_template_claim_eval})\footnote{We gradually refined this prompt after the first pilot run on a few examples showed that models sometimes generate unreasonably long explanations, leaving no space for the actual answer. To mitigate this issue, we added a request to provide the explanation "in at most one paragraph." Although we still observed some cases where the model's explanation was too long to generate the \texttt{<answer></answer>} tags, this occurred in only twice for \claude{} and once for \gptturbo{}. Importantly, the answer/label was always present in the explanation itself (early answer), so it was correctly extracted for that model.}
\end{itemize}
As testing these prompts on all models would be prohibitively expensive, we conducted these experiments with \gpto{} and \claude{}.\footnote{The total cost of this experiment, including iterative refinements to the prompt text, was \$3k USD.} We observed that \claude{}'s accuracy remained constant across all three setups (39.77\%), although the specific pairs it got correct varied. \gpto{} achieved the highest accuracy when prompted for the explanation followed by the answer (50\%), compared to 48.3\% for the answer followed by the explanation, and 49.43\% for the answer-only setup. For further experiments, we employ the \textit{explanation-then-answer} approach, as it yielded the best results despite the small differences between these three methods.

\begin{table}[t]
    \setlength{\tabcolsep}{4pt}
    \centering
    \resizebox{0.9\columnwidth}{!}{%
    \begin{tabular}{c p{9cm}}
    \toprule
         & \multicolumn{1}{c}{\bf Evaluation Template (Main)} \\
    \midrule
     \noalign{\vskip 1mm}
     & \texttt{You are provided with a context and a statement. Your task is to carefully read the context and then determine whether the statement is true or false. }\\
     \noalign{\vskip 2mm}
     & \texttt{Answer TRUE if the statement is true in its entirety based on the context provided.}\\
      & \texttt{Answer FALSE if any part of the statement is false based on the context provided.}\\
      \noalign{\vskip 2mm}
     & \texttt{\textless context\textgreater \textbf{book text}\textless/context\textgreater} \\
     \noalign{\vskip 2mm}
     & \texttt{\textless statement\textgreater \textbf{claim}\textless/statement\textgreater} \\
    \noalign{\vskip 2mm}
     & \texttt{\textless question\textgreater Based on the context provided, is the above statement TRUE or FALSE?\textless/question\textgreater} \\
    \noalign{\vskip 2mm}
     & \texttt{First provide an explanation of your decision-making process in at most one paragraph, and then provide your final answer. Use the following format:} \\
     & \texttt{\textless explanation\textgreater YOUR EXPLANATION\textless/explanation\textgreater} \\
     & \texttt{\textless answer\textgreater YOUR ANSWER\textless/answer\textgreater} \\
     \noalign{\vskip 1mm}
    \bottomrule
    \end{tabular}
    }
    \caption{Prompt template used for \texttt{Evaluation}. This prompt was employed to evaluate \textit{all} models. All open source models were also evaluated using prompt in \autoref{tab:prompt_template_claim_eval_simple}.} 
    \label{tab:prompt_template_claim_eval}
\end{table}

\begin{table}[t]
    \setlength{\tabcolsep}{4pt}
    \centering
    \resizebox{0.9\columnwidth}{!}{%
    \begin{tabular}{c p{9cm}}
    \toprule
         & \multicolumn{1}{c}{\bf Evaluation Template (Simplified)} \\
    \midrule
     \noalign{\vskip 1mm}
     & \texttt{You are provided with a context and a statement. Your task is to carefully read the context and then determine whether the statement is true or false. }\\
     \noalign{\vskip 2mm}
     & \texttt{Answer TRUE if the statement is true in its entirety based on the context provided.}\\
      & \texttt{Answer FALSE if any part of the statement is false based on the context provided.}\\
      \noalign{\vskip 2mm}
     & \texttt{\textless context\textgreater \textbf{book text}\textless/context\textgreater} \\
     \noalign{\vskip 2mm}
     & \texttt{\textless statement\textgreater \textbf{claim}\textless/statement\textgreater} \\
    \noalign{\vskip 2mm}
     & \texttt{\textless question\textgreater Based on the context provided, is the above statement TRUE or FALSE?\textless/question\textgreater} \\
     \noalign{\vskip 1mm}
    \bottomrule
    \end{tabular}
    }
    \caption{Simplified prompt template used for \texttt{Evaluation} of open-source models only.} 
    \label{tab:prompt_template_claim_eval_simple}
\end{table}
\begin{table}[t]
    \setlength{\tabcolsep}{4pt}
    \centering
    \resizebox{0.9\columnwidth}{!}{%
    \begin{tabular}{c p{9cm}}
    \toprule
         & \multicolumn{1}{c}{\bf Evaluation Template (BM25)} \\
    \midrule
     \noalign{\vskip 1mm}
     & \texttt{You are provided with excerpts of context and a statement. Your task is to carefully read the excerpts and then determine whether the statement is true or false.}\\
     \noalign{\vskip 2mm}
     & \texttt{Answer TRUE if the statement is true in its entirety based on the excerpts provided.}\\
      & \texttt{Answer FALSE if any part of the statement is false based on the excerpts provided.}\\
      \noalign{\vskip 2mm}
     & \texttt{\textless excerpt\_1\textgreater \textbf{excerpt\_1}\textless/excerpt\_1\textgreater} \\
     \noalign{\vskip 2mm}
      & \texttt{\textless excerpt\_2\textgreater \textbf{excerpt\_2}\textless/excerpt\_2\textgreater} \\
     \noalign{\vskip 2mm}
     & \texttt{\textless excerpt\_(i)\textgreater \textbf{...}\textless/excerpt\_(i)\textgreater} \\
     \noalign{\vskip 2mm}
      & \texttt{\textless excerpt\_k\textgreater \textbf{excerpt\_k}\textless/excerpt\_k\textgreater} \\
     \noalign{\vskip 2mm}
     & \texttt{\textless statement\textgreater \textbf{claim}\textless/statement\textgreater} \\
    \noalign{\vskip 2mm}
     & \texttt{\textless question\textgreater Based on the excerpts provided, is the above statement TRUE or FALSE?\textless/question\textgreater} \\
    \noalign{\vskip 2mm}
     & \texttt{First provide an explanation of your decision-making process in at most one paragraph, and then provide your final answer. Use the following format:} \\
     & \texttt{\textless explanation\textgreater YOUR EXPLANATION\textless/explanation\textgreater} \\
     & \texttt{\textless answer\textgreater YOUR ANSWER\textless/answer\textgreater} \\
     \noalign{\vskip 1mm}
    \bottomrule
    \end{tabular}
    }
    \caption{Prompt used for the \bm{} \texttt{Evaluation} pipeline.} 
    \label{tab:prompt_template_claim_eval_bm25}
\end{table}
\begin{table}[t]
    \setlength{\tabcolsep}{4pt}
    \centering
    \resizebox{0.9\columnwidth}{!}{%
    \begin{tabular}{c p{9cm}}
    \toprule
         & \multicolumn{1}{c}{\bf Evaluation Template (Answer-only)} \\
    \midrule
     \noalign{\vskip 1mm}
     & \texttt{You are provided with a context and a statement. Your task is to carefully read the context and then determine whether the statement is true or false. }\\
     \noalign{\vskip 2mm}
     & \texttt{Answer TRUE if the statement is true in its entirety based on the context provided.}\\
      & \texttt{Answer FALSE if any part of the statement is false based on the context provided.}\\
      \noalign{\vskip 2mm}
     & \texttt{\textless context\textgreater \textbf{book text}\textless/context\textgreater} \\
     \noalign{\vskip 2mm}
     & \texttt{\textless statement\textgreater \textbf{claim}\textless/statement\textgreater} \\
    \noalign{\vskip 2mm}
     & \texttt{\textless question\textgreater Based on the context provided, is the above statement TRUE or FALSE?\textless/question\textgreater} \\
    \noalign{\vskip 2mm}
     & \texttt{First provide an explanation of your decision-making process in at most one paragraph, and then provide your final answer. Use the following format:} \\
     & \texttt{\textless answer\textgreater YOUR ANSWER\textless/answer\textgreater} \\
     \noalign{\vskip 1mm}
    \bottomrule
    \end{tabular}
    }
    \caption{Prompt used for the initial pilot study when the model is prompted to return only the answer.} 
    \label{tab:prompt_template_claim_eval_devset_answer}
\end{table}
\begin{table}[t]
    \setlength{\tabcolsep}{4pt}
    \centering
    \resizebox{0.9\columnwidth}{!}{%
    \begin{tabular}{c p{9cm}}
    \toprule
         & \multicolumn{1}{c}{\bf Evaluation Template (Answer and Explanation)} \\
    \midrule
     \noalign{\vskip 1mm}
     & \texttt{You are provided with a context and a statement. Your task is to carefully read the context and then determine whether the statement is true or false. }\\
     \noalign{\vskip 2mm}
     & \texttt{Answer TRUE if the statement is true in its entirety based on the context provided.}\\
      & \texttt{Answer FALSE if any part of the statement is false based on the context provided.}\\
      \noalign{\vskip 2mm}
     & \texttt{\textless context\textgreater \textbf{book text}\textless/context\textgreater} \\
     \noalign{\vskip 2mm}
     & \texttt{\textless statement\textgreater \textbf{claim}\textless/statement\textgreater} \\
    \noalign{\vskip 2mm}
     & \texttt{\textless question\textgreater Based on the context provided, is the above statement TRUE or FALSE?\textless/question\textgreater} \\
    \noalign{\vskip 2mm}
     & \texttt{First provide an explanation of your decision-making process in at most one paragraph, and then provide your final answer. Use the following format:} \\
     & \texttt{\textless answer\textgreater YOUR ANSWER\textless/answer\textgreater} \\
       & \texttt{\textless explanation\textgreater YOUR EXPLANATION\textless/explanation\textgreater} \\
     \noalign{\vskip 1mm}
    \bottomrule
    \end{tabular}
    }
    \caption{Prompt used for the initial pilot study when the model is prompted to return the answer and then the explanation.} 
    \label{tab:prompt_template_claim_eval_devset_answer_explain}
\end{table}
\paragraph{Label extraction:} As we prompt the models to generate answers following a structured template, we first attempt to extract the answer from the first encountered \texttt{<answer></answer>} tags.\footnote{In cases where the text between the tags is longer than simply true/false, we follow the replacement pipeline and finally extract the label from the text between the tags.} If this is not possible (e.g., the model did not generate the tags or produced output longer than a simple true/false within the tags), we apply the following steps: 
\begin{enumerate}
    \item Replace any occurrence of ``true or false'' and the text of the claim itself with an empty string.\footnote{"True or false" often occurs when the model repeats the original question. We additionally replace the claim itself, as it can also contain words like true/false (e.g., "true friend").}
    \item Replace ``not true'' with ``false.''
    \item Extract ``true'' if only ``true'' is present, ``false'' if only ``false'' is present, and the first occurrence if both are present in the generation.
\end{enumerate}
If neither ``true'' nor ``false'' is present, we count it as an automatic failure.

\begin{table}[t]
\centering
\footnotesize
\resizebox{0.9\columnwidth}{!}{%
\begin{tabular}{lcc}
\toprule
\textsc{Model} & \textsc{Percentage} & \textsc{Count (Present/Total)} \\
\midrule
\gpto & 1.1\% & 13/1234 \\
\gptturbo & 53.2\% & 657/1234 \\
\claude & 68.4\% & 1281/1874 \\
\sonnet & 55.7\% & 1044/1874 \\
\geminipro & 44.0\% & 453/1029 \\
\geminiflash & 24.3\% & 250/1030 \\
\bottomrule
\end{tabular}
}
\caption{Percentage and count of explanations where the first sentence contains an early response by the model. Open-source models are excluded as they typically do not generate structured responses or explanations.}
\label{tab:early_answer_percentage}
\end{table}

\paragraph{Early response:} We employ the \textit{explanation-then-answer} template to allow the model to reason about the answer before providing its final choice. However, we observe that models sometimes still return the answer first, within the \texttt{<explanation></explanation>} tags, which is in line with the observations in \citet{levy2024flenqa}.\footnote{We identify instances of an early response by segmenting the text between the \texttt{<explanation></explanation>} tags into sentences. In the first sentence, we remove occurrences of "true or false," "to determine if the statement is true," and similar phrases. We then check if the first sentence contains "true" or "false." Note that this excludes phrases like "this statement is incorrect." Additionally, we acknowledge that phrases such as "true nature," which may appear in the first sentence, may slightly affect this percentage as they are irrelevant to the model's answer.} Overall, \claude\ and \sonnet\ are the most affected, with 68.4\% and 55.7\% of explanations containing either "true" or "false" in the first sentence, respectively (see \autoref{tab:early_answer_percentage}).

\paragraph{Generation issues:} We encounter several issues during model prompting. Most notably, \geminipro{} and \geminiflash{} return a "prohibited content" API error for about 48.6\% of claims, significantly reducing the number of claims (and claim pairs) processed by these models despite their large claimed context window. We also observe that open-source models struggle to follow the assigned template. Only \comr{} returns the answer with proper \texttt{<answer></answer>} tags 100\% of the time. This compares to \comrplus{} at 66.3\%, \phimodel{} at 62.7\%, and \gemma{} at 52.5\%. Additionally, we notice that generations by \gemma{} are plagued with code output, often not including any instances of "true" or "false," while \longllama{} returns empty outputs for the prompt in \autoref{tab:prompt_template_claim_eval} (i.e., the non-simplified version of the prompt).

\section{Results}
\label{app_sec:results_info}

In this section of the appendix we provide more results of our analysis. We also provide more examples of claims where the models have generated incorrect label or incorrect justification in \autoref{tab:generation_example_more}.

\begin{table*}[h!]
    \centering
    \footnotesize
    \renewcommand{\arraystretch}{1.2}
    \resizebox{0.9\textwidth}{!}{%
    \begin{tabular}{p{0.33\textwidth} p{0.5\textwidth} p{0.45\textwidth}}
        \toprule
         \textsc{\textbf{Claim \& Gold Label}} & \textsc{\textbf{Model's Label \& Justification}} & \textsc{\textbf{Human Explanation}} \\
        \midrule
        \textit{Wyatt apologizes to Ernest because he believes
        McTavish left him a negative review.} \newline 
        
        \textit{Everyone on the Train is a Suspect} \newline 
        \makebox[110pt][r]{by Benjamin Stevenson}  \newline 

        \makebox[250pt][r]{\texttt{{\color{teal}\textbf{\Large--True}}} \newline} \newline 

        & \textit{Wyatt's apology to Ernest is mentioned in the context, but it is not because he believes McTavish left him a negative review. \textbf{\color{purple}Instead, Wyatt's apology is vague and not directly linked to the review.} Ernest himself is confused about the reason for the apology, and it is later revealed that Simone, Ernest's agent, was behind the negative review as part of a coded message to McTavish. Therefore, the statement that Wyatt apologizes to Ernest because he believes McTavish left him a negative review is incorrect.} \newline
        \newline
        --\gpto{}: \texttt{{\color{purple}\textbf{\Large False}}}
        
        & When Wyatt apologizes, Ernest does not initially understand the reason behind it. However, it is \textit{heavily} implied shortly after that Wyatt, McTavish's publisher, was apologizing for the bad review, which at that time everyone, including Wyatt thought was written by McTavish himself. \\
        \textit{Emily Sparrow was pregnant with John's child and disappeared in 1966.} \newline 
        
        \textit{How to Solve Your Own Murder} \newline
        \makebox[90pt][r]{by Kristen Perrin} \newline

        \makebox[250pt][r]{\texttt{{\color{teal}\textbf{\Large --True}}} \newline} \newline 
        
        & \textit{The text states that Emily Sparrow was pregnant, but \textbf{\color{purple}it is never explicitly stated who the father of the child is}. While it is implied that the father could be John, it is also implied that it could be Rutherford Gravesdown (Ford).} 
        \newline
        \newline
        --\geminipro{}: \texttt{{\color{purple}\textbf{\Large False}}}
        
        & Throughout the book, the reader is taken back to 1966, where it is revealed that Emily was pregnant, and the father's identity was uncertain—either Lord Gravesdown or John. The baby, Laura, was later adopted and had a daughter, Annie. In the latter part of the book, set in the present day, Annie meets John, who confesses that he is her grandfather, thus confirming that he was the father of Emily's baby.  \\
        \textit{None of the writers on the train went to the Gorge excursion as they had to stay to participate in a festival event.} \newline
        
        \textit{Everyone on the Train is a Suspect} \newline 
        \makebox[110pt][r]{by Benjamin Stevenson} \newline

        \makebox[250pt][r]{\texttt{{\color{purple}\textbf{\Large --False}}} \newline} \newline 
        
        & \textit{The statement is true based on the context. The passage states that while the non-festival tourists went on a scheduled day trip sailing down Katherine Gorge, "In front of our carriages were forty or so black fold-out chairs set up in the red dirt for the festival attendees." \textbf{\color{purple}This indicates that the writers stayed behind for a festival event} while the other passengers went on the Gorge excursion.} 
        \newline
        \newline
        --\claude{}: \texttt{{\color{teal}\textbf{\Large True}}}
        
        & While all writers invited to the festival stayed for the event, Juliet, who was also an accomplished writer and, in fact, we later learn she was originally invited as well, went on the trip. \\

        \textit{Lost Bride Manor is haunted by various spirits, including the benevolent ghost of Clover, Sonya's grandmother and Collin's mother, who communicates through music.} \newline
         
        \textit{Inheritance} \newline 
        \makebox[110pt][r]{by Nora Roberts} \newline 

        \makebox[250pt][r]{\texttt{{\color{teal}\textbf{\Large --True}}} \newline} \newline
        
        & \textit{The statement is mostly true, but contains one key inaccuracy. The context does indicate that Lost Bride Manor is haunted by various spirits, including Clover, who communicates through music. Clover is described as Sonya's grandmother and Collin's mother, which is correct. However, the statement characterizes Clover as "benevolent", but the context does not provide enough information to definitively label her as such. \textbf{\color{purple}While Clover seems to be helpful at times, playing music and aiding Sonya, the text does not explicitly state that she is entirely benevolent.} Given this small but significant discrepancy, the statement cannot be considered entirely true.} 
        \newline
        \newline
        --\sonnet{}: \texttt{{\color{purple}\textbf{\Large False}}}
        
        & Clover is a ghost haunting the manor where Sonya moves in. From the first day, Clover, who we learn is the ghost of Sonya's grandmother, helps Sonya by playing music to cheer her up, adding to her courage, and keeping her company. Clover also tries to protect Sonya from the evil spirit that also haunts the manor. Although it is not directly stated in the text, Clover's kind and well-meaning nature towards Sonya is evident through her actions. \\
        \midrule  
        \textit{Ernest, the narrator, drops numerous hints about the killer's identity from the start, but any hints of how the first victim dies aren't revealed until chapter 11.} \newline 
        
        \textit{Everyone on the Train is a Suspect} \newline 
        \makebox[110pt][r]{by Benjamin Stevenson} \newline

        \makebox[250pt][r]{\texttt{{\color{purple}\textbf{\Large --False}}} \newline} \newline 
        
         & \textit{To determine the accuracy of the statement, I reviewed the context for any early hints about the killer's identity and the method of the first victim's death. The prologue and early chapters contain several clues about the characters' motives and potential involvement in the murder. Specifically, \textbf{\color{purple}the prologue includes a detailed description of a potential murder scene, mentioning "the bloodied handprint on the bathroom door" and "the crimson, dripping tip of the murder weapon."} This indicates that hints about the method of the first victim's death are indeed revealed before chapter 11.}
        \newline
        \newline
        --\gpto{}: \texttt{{\color{purple}\textbf{\Large False}}}       \newline
         & It is true that Ernest, the narrator, drops some hints about how the first victim dies from the start. However, the passage cited by the model is irrelevant. It is taken from the Prologue, where Ernest, a mystery novel writer, talks about elements that typically appear in a mystery novel's prologue. He mentions a "bloodied handprint" and "the crimson, dripping tip of the murder weapon," which are common in mystery novels prologes. However, in this novel, the first victim was poisoned, and Ernest hints at this by saying that the inspiration for this book "came from a drink" with the victim. \\
      
         \textit{Levi promises Grace that he will kill anyone who tries to hurt her, and Grace calls him to help her dispose of Norman's body.} \newline 
         
        \textit{Safe and Sound} \newline 
        \makebox[110pt][r]{by Laura McHugh} \newline

        \makebox[250pt][r]{\texttt{{\color{purple}\textbf{\Large --False}}} \newline} \newline
        
         & \textit{In the context, Levi does promise Grace that he would kill anyone who tried to hurt her again. However, when Grace kills Norman,…she calls Tyson, Levi's brother. The passage states: "She…wanted to make it look like she couldn't have survived. So she could just go missing."… \textbf{\color{purple}It then says "Tyson grabbed me, clamped his hands around my neck, and didn't let go. `I dumped a body for you.'"} This shows that it was Tyson, not Levi, who Grace called to help dispose of Norman's body.}
        \newline
        \newline
        --\gpto{}: \texttt{{\color{purple}\textbf{\Large False}}}        \newline
         & The model's justification begins correctly, but the first passage cited is unrelated to the claim and clearly a lie told by Tyson in the context of the story. Additionally, the text "Tyson grabbed me, clamped his hands around my neck, and didn't let go." never appears in the book. Tyson does say "I dumped a body for you." to Grace---which proves the claim is false---but he does not choke her before saying so or at any other point in the book.\\

         \textit{While on Sigrid's boat, Seventeen admits to Kat that he had the same underwater dream that she had.} \newline 
         
        \textit{Assassin Eighteen} \newline 
        \makebox[110pt][r]{by John Brownlow} 

        \makebox[250pt][r]{\texttt{{\color{teal}\textbf{\Large --True}}} \newline} \newline
        
         & \textit{The statement is true based on the context provided. In Part 8, \textbf{\color{purple}after Kat is rescued from the water by Rachel and Vilmos in the Zodiac}, she describes her near-death experience underwater where she felt she and Seventeen were entwined as one. Seventeen then admits "I had the same dream", confirming he experienced the same vision while he was drowning. }
        \newline
        \newline
        --\claude{}: \texttt{{\color{teal}\textbf{\Large True}}}
         & It is true that Seventeen admits he had the same dream in Part 8 of the book. However, Kat was not rescued by Rachel and Vilmos in the Zodiac during this scene. That happens much later in the book, towards the end of Part 11. This conversation happened when Kat and Seventeen were rescued by Sigrid. \\

         \textit{Sonya befriends Trey's sister Anna, a potter, and his cousin Owen Poole, who becomes involved in solving the mystery of the lost brides.} \newline 
        
        \textit{Inheritance} \newline 
        \makebox[110pt][r]{by Nora Roberts} \newline 
        
         \makebox[250pt][r]{\texttt{{\color{purple}\textbf{\Large --False}}} \newline} \newline 
         & \textit{The statement is mostly true, but contains one key inaccuracy. The context shows that Sonya does befriend Trey's sister Anna, who is a potter, and helps redesign her website. Sonya also meets and interacts with Owen Poole, \textbf{\color{purple}Trey's cousin, who becomes involved in investigating the haunted manor and the mystery of the lost brides. However, the statement is incorrect in saying that Owen "becomes involved in solving the mystery of the lost brides."} While Owen does help investigate strange occurrences at the manor and even enters the mirror with Sonya at the end, he is not described as actively trying to solve the mystery of the lost brides specifically. Sonya is the main character focused on that mystery. Therefore, while most of the statement is accurate, this one detail makes it not entirely true.}
        \newline
        \newline
        --\sonnet{}: \texttt{{\color{purple}\textbf{\Large False}}}
         & It is true that Sonya befriends Trey's sister, Anna, and Owen. However, Owen is not Trey's cousin; he is Sonya's cousin, whom she did not know about. Owen and Trey then try to help Sonya solve the mystery of the lost brides. \\
         
        \bottomrule
    \end{tabular}
    }
   \caption{\textbf{Top half:} Examples of claims where the model generated incorrect labels, along with the model-generated label and justification, and human-written explanations for the correct label of each claim. \textbf{Bottom half:} Examples of claims correctly labeled by the model but with wrong justifications. The explanations were provided by the authors, based on examples from books they read. Tags have been removed from the generated content for clarity.}
    \label{tab:generation_example_more}
\end{table*}

\begin{figure*}[tbp]
  \includegraphics[width=1\linewidth]{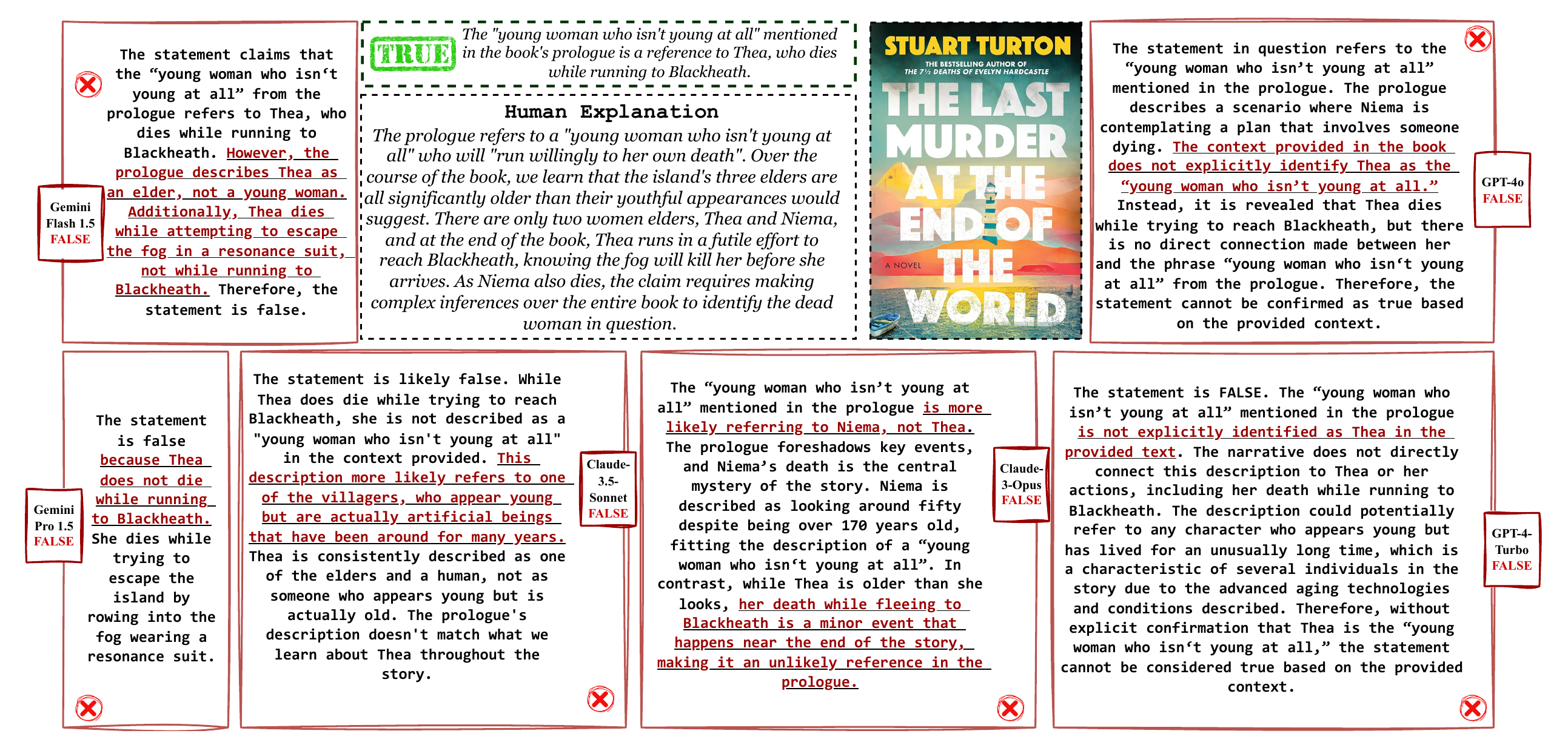} 
  \caption{Example of a True claim for which all models predicted incorrect label. We provide the claim along with the annotator's explanation for clarity.}
  \label{fig:all_models_failed}
\end{figure*}

\autoref{tab:pair_accuracy_models_no_classics} reports the accuracy of each model excluding the classic novels. Overall, \gpto{} still performs the best from all the models, although its accuracy drops slightly from 55.8\% to 55.1\% when classic novels are excluded.

\begin{table*}[t]
\centering
\resizebox{0.75\textwidth}{!}{%
\begin{tabular}{lcc}
\toprule
\textsc{Model} & \textsc{Pair ACC}\textsubscript{(correct/total)} & \textsc{Common Set ACC} \\
\midrule
\gpto & \textbf{55.3}\textsubscript{(333/602)} & \textbf{57.5}\textsubscript{(195/339)} \\
\gptturbo & 39.5\textsubscript{(238/602)} & 38.9\textsubscript{(132/339)} \\
\claude & 49.4\textsubscript{(439/889)} & 49.9\textsubscript{(169/339)} \\
\sonnet & 42.0\textsubscript{(373/889)} & 41.3\textsubscript{(140/339)} \\
\geminipro & 48.1\textsubscript{(224/466)} & 48.4\textsubscript{(164/339)} \\
\geminiflash & 34.5\textsubscript{(161/467)} & 35.1\textsubscript{(119/339)}  \\
\midrule
\comr{} & 18.8\textsubscript{(81/430)} & \textit{n/a} \\
\comr{}\textsubscript{simple} & 22.1\textsubscript{(95/430)} & \textit{n/a} \\
\comrplus & 17.4\textsubscript{(75/430)} & \textit{n/a} \\
\comrplus{}\textsubscript{simple} & 13.3\textsubscript{(57/430)} & \textit{n/a} \\
\phimodel{} & 9.7\textsubscript{(23/237)} & \textit{n/a} \\
\phimodel{}\textsubscript{simple} & 14.2\textsubscript{(45/316)} & \textit{n/a} \\
\gemma{} & 4.2\textsubscript{(39/938)} & \textit{n/a} \\
\gemma{}\textsubscript{simple} & 7.7\textsubscript{(72/938)} & \textit{n/a} \\
\longllama{}\textsubscript{simple} & 5.1\textsubscript{(45/889)} & \textit{n/a} \\
\midrule
\bm{} (\textit{k=5}) & 27.8\textsubscript{(261/938)} & 28.1\textsubscript{(95/338)} \\
\bm{} (\textit{k=25}) & 44.0\textsubscript{(413/938)} & 45.9\textsubscript{(155/338)} \\
\bm{} (\textit{k=50}) & 49.5\textsubscript{(464/938)} & 50.3\textsubscript{(170/338)} \\
\midrule
\textsc{Random} & 25.0\textsubscript{(234/938)} & 25.0\textsubscript{(85/339)} \\
\bottomrule
\end{tabular}
}
\caption{Model accuracy on claim pairs for all data excluding \textit{classic} novels (see \autoref{tab:prompt_template_claim_eval} for the prompt; and \autoref{tab:prompt_template_claim_eval_bm25} for the prompt employed with \textsc{BM25}). ``\textsc{Common set}'' refers to claim pairs shared among the models. The subscript ``\textsc{simple}'' refers to the calls done with simplified prompt (\autoref{tab:prompt_template_claim_eval_simple}).}
\label{tab:pair_accuracy_models_no_classics}
\end{table*}

\paragraph{Are newer books harder to verify?} Our data includes (1) classic books, which were almost certainly in the models' training data, (2) books published in 2023, which might have been included in the training data, and (3) books published in 2024, which were likely not included in the training data.\footnote{Given the time required to collect, clean the data, and train the model, we hypothesize that at least books published in May/June 2024 were not included in the models' training data.} Hence, we look at models' performance by the publication year (see \autoref{app_table:model_acc_by_pubyear}). While we do not observe a large difference in performance for books published in 2023 vs. 2024, \gpto{} and \geminiflash{}'s accuracy drops slightly for the newer books, from 56.7\% to 53.5\% and from 37.4\% to 30.7\%, respectively. On the other hand, the performance of \claude{}, \sonnet{}, \geminipro{}, and \gptturbo{} seems to be higher for books published in 2024 than in 2023. Overall, these differences may be due to the inherent difficulty of the claims for a given model rather than the publication year itself. We do observe that both \gpto{} and \gptturbo{} perform much better on "The Great Gatsby" (the only classic book, which fits their context window) at 73.3\% and 66.7\%, respectively. However, no definitive claims can be made as the set of claims is too small and related to one book only.

\begin{table*}[t]
\centering
\begin{tabular}{lccc}
\toprule
\textsc{Model} & \textsc{Classics}$_{\text{(correct/total)}}$ & \textsc{2023}$_{\text{(correct/total)}}$ & \textsc{2024}$_{\text{(correct/total)}}$ \\
\midrule
\gpto & 73.3$_{\text{(11/15)}}$ &           56.7$_{\text{(194/342)}}$ & 53.5$_{\text{(139/260)}}$ \\
\gptturbo & 66.7$_{\text{(10/15)}}$ &       39.5$_{\text{(135/342)}}$ & 39.6$_{\text{(103/260)}}$ \\
\claude & 50.0$_{\text{(24/48)}}$ &         48.8$_{\text{(227/465)}}$ & 50.0$_{\text{(212/424)}}$ \\
\sonnet & 22.9$_{\text{(11/48)}}$ &         41.9$_{\text{(195/465)}}$ & 42.0$_{\text{(178/424)}}$ \\
\geminipro & 47.9$_{\text{(23/48)}}$ &      45.7$_{\text{(121/265)}}$ & 51.2$_{\text{(103/201)}}$ \\
\geminiflash & 31.2$_{\text{(15/48)}}$ &    37.4$_{\text{(99/265)}}$ & 30.7$_{\text{(62/202)}}$ \\
\midrule
\comr & 40.0$_{\text{(6/15)}}$ & 18.5$_{\text{(51/275)}}$ & 19.4$_{\text{(30/155)}}$ \\
\comr$_{\textit{simple}}$ & 33.3$_{\text{(5/15)}}$ & 21.5$_{\text{(59/275)}}$ & 23.2$_{\text{(36/155)}}$ \\
\comrplus & 13.3$_{\text{(2/15)}}$ &        18.5$_{\text{(51/275)}}$ & 15.5$_{\text{(24/155)}}$ \\
\comrplus{}$_{\textit{simple}}$ & 26.7$_{\text{(4/15)}}$ & 14.5$_{\text{(40/275)}}$ & 11.0$_{\text{(17/155)}}$ \\
\phimodel{} & 10.0$_{\text{(1/10)}}$ & 8.4$_{\text{(12/143)}}$ & 10.6$_{\text{(10/94)}}$ \\
\phimodel$_{\textit{simple}}$ & 20.0$_{\text{(3/15)}}$ & 13.1$_{\text{(25/191)}}$ & 16.0$_{\text{(20/125)}}$ \\
\gemma & 0.0$_{\text{(0/63)}}$ & 3.7$_{\text{(18/489)}}$ & 4.7$_{\text{(21/449)}}$ \\
\gemma$_{\textit{simple}}$ & 4.8$_{\text{(3/63)}}$ & 9.8$_{\text{(48/489)}}$ & 5.3$_{\text{(24/449)}}$ \\
\longllama$_{\textit{simple}}$ & 2.1$_{\text{(1/48)}}$ & 6.5$_{\text{(30/465)}}$ & 3.5$_{\text{(15/424)}}$ \\
\midrule
\bm{} (\textit{k=5}) & 33.3$_{\text{(21/63)}}$ & 28.8$_{\text{(141/489)}}$ & 26.7$_{\text{(120/449)}}$ \\
\bm{} (\textit{k=25}) & 44.4$_{\text{(28/63)}}$ & 44.6$_{\text{(218/489)}}$ & 43.4$_{\text{(195/449)}}$ \\
\bm{} (\textit{k=50}) & 52.4$_{\text{(33/63)}}$ & 47.2$_{\text{(231/489)}}$ & 51.9$_{\text{(233/449)}}$ \\
\bottomrule
\end{tabular}
\caption{Models accuracy on claim pairs by the publication year.}
\label{app_table:model_acc_by_pubyear}
\end{table*}

\begin{figure}[tbp]
  \includegraphics[width=1\linewidth]{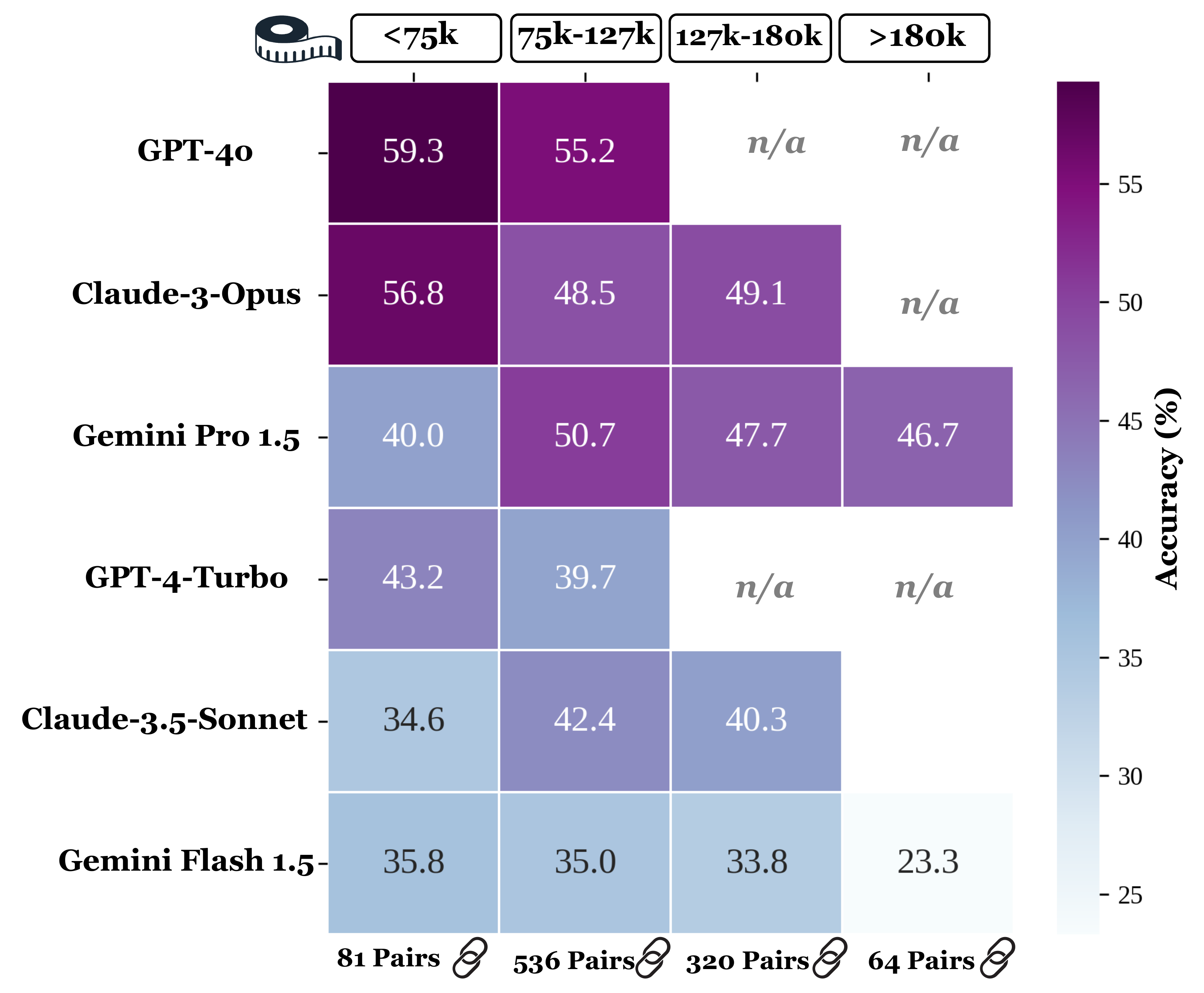} 
  \caption{Model performance across different book lengths. Token counts are provided as per \texttt{tiktoken}. Length buckets were determined by taking one standard deviation from the mean on both sides, with additional buckets for values above and below this range. The number of valid pairs in each bucket is provided below.}
  \label{fig:length_acc_heatmap}
\end{figure}

\paragraph{It is unclear if claims about longer books are harder to verify:} We categorized \name\ books into four buckets based on length: (1) up to 75k tokens, (2) 75k to 127k tokens, (3) 127k to 180k tokens, and (4) above 180k tokens.\footnote{We selected these numbers based on the mean length of our books (127k tokens), adjusted by adding or subtracting one standard deviation. Additional buckets were included beyond these limits on both ends.} We observed a slight drop in performance between the first and second buckets for \claude{} (56.8\% to 48.5\%), \gpto{} (59.3\% to 55.2\%), and \gptturbo{} (43.2\% to 39.7\%). However, these differences are small and could be influenced by factors such as the number of claims in each bucket or the complexity of the book itself (i.e., longer books tend to have more complex narratives). No such difference was observed for \geminiflash{} and \sonnet{}, which performance is similar or slightly better, albeit still low, for longer books. \geminipro{} performed worst for the shortest books (40.0\%), with performance for longer books varying from 46.7\% to 50.7\%. See \autoref{app_table:model_acc_by_lenght} and \autoref{fig:length_acc_heatmap} for the details. We also report Pearson correlation between the book length and model's accuracy for each model (see \autoref{tab:corr_pearson_len}).

\begin{table*}[h!]
\centering
\begin{tabular}{lllll}
\toprule
\textsc{Model} & \textsc{Below 75k} & \textsc{75k-127k} & \textsc{127k-180k} & \textsc{Above 180k} \\
\midrule
\gpto & 59.3$_{\text{(48/81)}}$ & 			55.2$_{\text{(296/536)}}$ & \textit{n/a} & \textit{n/a} \\
\gptturbo & 43.2$_{\text{(35/81)}}$ & 		39.7$_{\text{(213/536)}}$ & \textit{n/a} & \textit{n/a} \\
\claude & 56.8$_{\text{(46/81)}}$ & 		48.5$_{\text{(260/536)}}$ & 49.1$_{\text{(157/320)}}$ & \textit{n/a} \\
\sonnet & 34.6$_{\text{(28/81)}}$ & 		42.4$_{\text{(227/536)}}$ & 40.3$_{\text{(129/320)}}$ & \textit{n/a} \\
\geminipro & 40.0$_{\text{(32/80)}}$ & 		50.7$_{\text{(139/274)}}$ & 47.7$_{\text{(62/130)}}$ & 46.7$_{\text{(14/30)}}$ \\
\geminiflash & 35.8$_{\text{(29/81)}}$ & 	35.0$_{\text{(96/274)}}$ & 33.8$_{\text{(44/130)}}$ & 23.3$_{\text{(7/30)}}$ \\
\midrule
\comr & 					25.9$_{\text{(21/81)}}$ & 18.1$_{\text{(66/364)}}$ & \textit{n/a} & \textit{n/a} \\
\comr$_{\textit{simple}}$ & 25.9$_{\text{(21/81)}}$ & 21.7$_{\text{(79/364)}}$ & \textit{n/a} & \textit{n/a} \\
\comrplus & 				27.2$_{\text{(22/81)}}$ & 15.1$_{\text{(55/364)}}$ & \textit{n/a} & \textit{n/a} \\
\comrplus$_{\textit{simple}}$ & 27.2$_{\text{(22/81)}}$ & 	10.7$_{\text{(39/364)}}$ & \textit{n/a} & \textit{n/a} \\
\longllama$_{\textit{simple}}$ & 9.9$_{\text{(8/81)}}$ & 	5.4$_{\text{(29/536)}}$ & 2.8$_{\text{(9/320)}}$ & \textit{n/a} \\
\phimodel & 					13.9$_{\text{(5/36)}}$ & 	8.5$_{\text{(18/211)}}$ & \textit{n/a} & \textit{n/a} \\
\phimodel$_{\textit{simple}}$ & 13.6$_{\text{(11/81)}}$ & 	14.8$_{\text{(37/250)}}$ & \textit{n/a} & \textit{n/a} \\
\gemma & 						4.9$_{\text{(4/81)}}$ & 4.1$_{\text{(22/536)}}$ & 3.4$_{\text{(11/320)}}$ & 3.1$_{\text{(2/64)}}$ \\
\gemma$_{\textit{simple}}$ & 	9.9$_{\text{(8/81)}}$ & 7.6$_{\text{(41/536)}}$ & 6.9$_{\text{(22/320)}}$ & 6.2$_{\text{(4/64)}}$ \\
\midrule
\bm{} (\textit{k=5}) & 			29.6$_{\text{(24/81)}}$ & 28.0$_{\text{(150/536)}}$ & 26.6$_{\text{(85/320)}}$ & 35.9$_{\text{(23/64)}}$ \\
\bm{} (\textit{k=25}) & 		49.4$_{\text{(40/81)}}$ & 42.7$_{\text{(229/536)}}$ & 41.6$_{\text{(133/320)}}$ & 60.9$_{\text{(39/64)}}$ \\
\bm{} (\textit{k=50}) & 		51.9$_{\text{(42/81)}}$ & 47.9$_{\text{(257/536)}}$ & 48.1$_{\text{(154/320)}}$ & 68.8$_{\text{(44/64)}}$ \\
\bottomrule
\end{tabular}
\caption{Models accuracy on claim pairs by the book length in tokens (\texttt{tiktoken}).}
\label{app_table:model_acc_by_lenght}
\end{table*}

\begin{table*}[t]
\centering
\footnotesize
\begin{tabular}{l c c}
\toprule
\textsc{Model} & \textsc{Correlation} & \textsc{\textit{p}-value} \\
\midrule
\gpto{} & -0.02 & 0.9150 \\
\gptturbo{} & 0.05 & 0.7722 \\
\claude{} & -0.09 & 0.4682 \\
\sonnet{} & 0.03 & 0.8252 \\
\geminipro{} & 0.03 & 0.8773 \\
\geminiflash{} & -0.13 & 0.4698 \\
\midrule
\comr{} & -0.38 & 0.0400\textsuperscript{*} \\
\comr{}\textsubscript{simple} & -0.22 & 0.2329 \\
\comrplus{} & -0.16 & 0.3865 \\
\comrplus{}\textsubscript{simple} & -0.53 & 0.0029\textsuperscript{**} \\
\phimodel{} & -0.57 & 0.0091\textsuperscript{**} \\
\phimodel{}\textsubscript{simple} & 0.12 & 0.5972 \\
\gemma{} & -0.11 & 0.3959 \\
\gemma{}\textsubscript{simple} & -0.09 & 0.4858 \\
\longllama\textsubscript{simple}& -0.14 & 0.2760 \\
\midrule
\bm{} (\textit{k=5}) & 0.11 & 0.3774 \\
\bm{} (\textit{k=25}) & 0.21 & 0.0849 \\
\bm{} (\textit{k=25}) & 0.27 & 0.0263\textsuperscript{*} \\
\hline
\end{tabular}
\caption{Pearson correlation between length and accuracy for different models. Significant correlations are marked with an asterisk.}
\label{tab:corr_pearson_len}
\end{table*}

\begin{figure}[tbp]
  \includegraphics[width=1\linewidth]{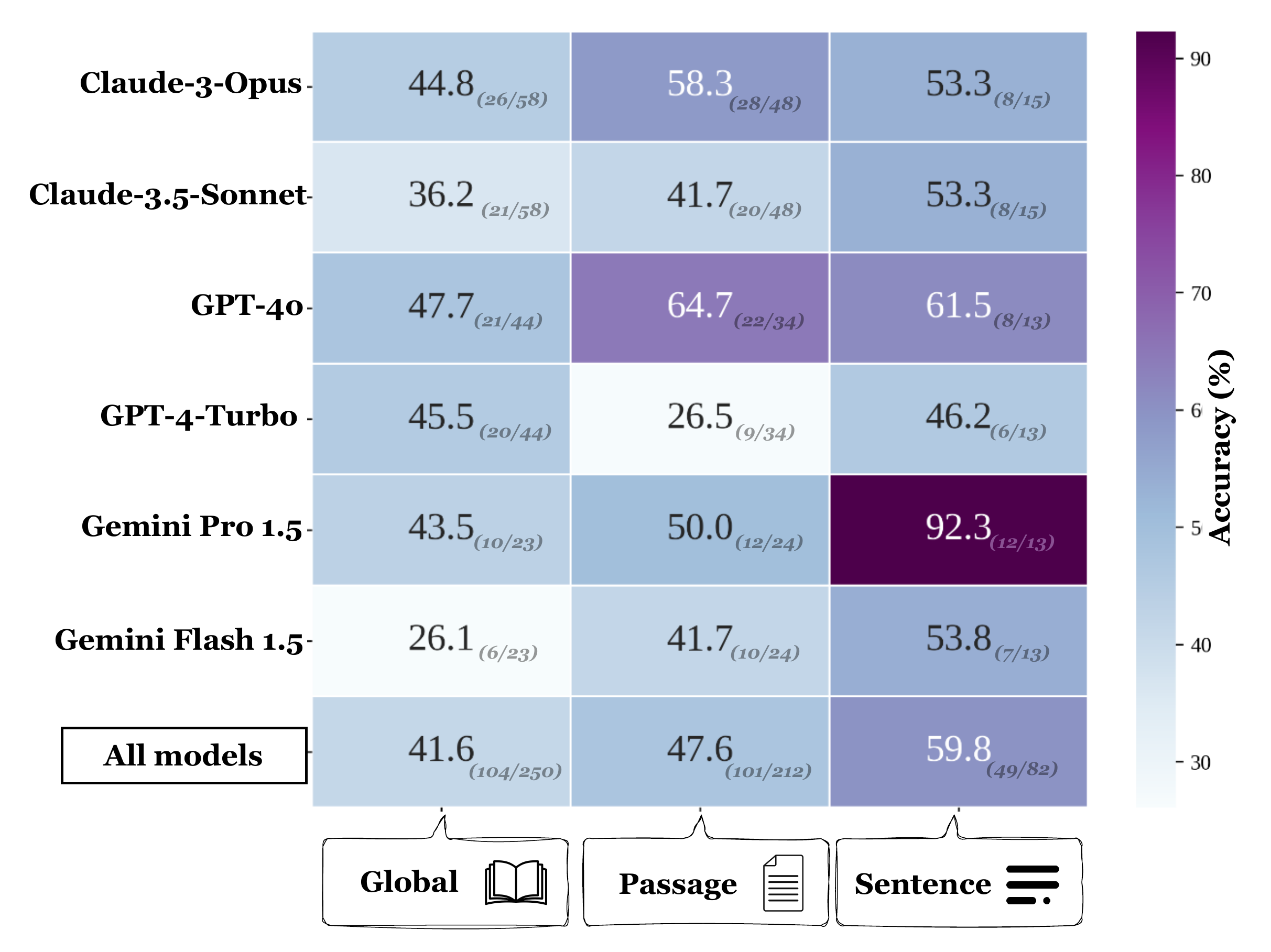} 
    \caption{Performance of different closed-source models based on the scope of evidence. The count of correctly identified pairs and the total count of pairs are provided in brackets for reference.}
  \label{fig:scope_annot_acc_by_model}
\end{figure}

\paragraph{Impact of evidence scope:} We report the performance of all models on the subset of data annotated for evidence scope in  \autoref{tab_app:evidence_scope_all_models}.  Note that the number of examples varies between models due to the restrictions of the models' context windows and \geminipro{}/\geminiflash{} API errors. Additionally, the performance of closed-source models is visualized in  \autoref{fig:scope_annot_acc_by_model}.

\begin{table*}[t]
\centering
\begin{tabular}{l l l l}
\hline
\textsc{Model} & \textsc{Sentence}\textsubscript{(correct/total)} & \textsc{Passage}\textsubscript{(correct/total)} & \textsc{Global}\textsubscript{(correct/total)} \\
\toprule
\gpto & 61.5 \textsubscript{(8/13)} & 64.7 \textsubscript{(22/34)} & 47.7 \textsubscript{(21/44)} \\
\gptturbo & 46.2 \textsubscript{(6/13)} & 26.5 \textsubscript{(9/34)} & 45.5 \textsubscript{(20/44)} \\
\claude & 53.3 \textsubscript{(8/15)} & 58.3 \textsubscript{(28/48)} & 44.8 \textsubscript{(26/58)} \\
\sonnet & 53.3 \textsubscript{(8/15)} & 41.7 \textsubscript{(20/48)} & 36.2 \textsubscript{(21/58)} \\
\geminipro & 92.3 \textsubscript{(12/13)} & 50.0 \textsubscript{(12/24)} & 43.5 \textsubscript{(10/23)} \\
\geminiflash & 53.8 \textsubscript{(7/13)} & 41.7 \textsubscript{(10/24)} & 26.1 \textsubscript{(6/23)} \\
\midrule
\comr & 25.0 \textsubscript{(2/8)} & 23.1 \textsubscript{(6/26)} & 3.7 \textsubscript{(1/27)} \\
\comr\textsubscript{simple} & 37.5 \textsubscript{(3/8)} & 11.5 \textsubscript{(3/26)} & 22.2 \textsubscript{(6/27)} \\
\comrplus & 25.0 \textsubscript{(2/8)} & 19.2 \textsubscript{(5/26)} & 18.5 \textsubscript{(5/27)} \\
\comrplus\textsubscript{simple} & 12.5 \textsubscript{(1/8)} & 7.7 \textsubscript{(2/26)} & 11.1 \textsubscript{(3/27)} \\
\phimodel & 0.0 \textsubscript{(0/4)} & 12.5 \textsubscript{(2/16)} & 0.0 \textsubscript{(0/16)} \\
\phimodel\textsubscript{simple} & 0.0 \textsubscript{(0/6)} & 15.8 \textsubscript{(3/19)} & 0.0 \textsubscript{(0/21)} \\
\gemma{}\textsubscript{simple} & 6.7 \textsubscript{(1/15)} & 2.1 \textsubscript{(1/48)} & 1.7 \textsubscript{(1/58)} \\
\gemma & 0.0 \textsubscript{(0/15)} & 8.3 \textsubscript{(4/48)} & 1.7 \textsubscript{(1/58)} \\
\longllama\textsubscript{simple} & 6.7 \textsubscript{(1/15)} & 0.0 \textsubscript{(0/48)} & 3.4 \textsubscript{(2/58)} \\
\midrule
\bm{} (\textit{k=5}) & 46.7 \textsubscript{(7/15)} & 22.9 \textsubscript{(11/48)} & 25.9 \textsubscript{(15/58)} \\
\bm{} (\textit{k=25}) & 66.7 \textsubscript{(10/15)} & 43.8 \textsubscript{(21/48)} & 29.3 \textsubscript{(17/58)} \\
\bm{} (\textit{k=50}) & 73.3 \textsubscript{(11/15)} & 45.8 \textsubscript{(22/48)} & 41.4 \textsubscript{(24/58)} \\
\bottomrule
\end{tabular}
\caption{Models' accuracy (\%) by the scope of evidence on the annotated subset of data. Counts are provided as subscripts.}
\label{tab_app:evidence_scope_all_models}
\end{table*}

\paragraph{Impact of irrelevant context:} As mentioned previously, creating \textit{complex} claims with evidence at different book depths is challenging, especially when aiming to create global claims that require the model to reason over a longer context. Instead, we utilize short stories to investigate whether claims made about the same story are harder to identify when the model is prompted with the entire collection versus just the story itself. The accuracies of all models for the whole collection versus individual stories are presented in \autoref{app_tab:acc_on_stories_and_collections_with_true_false}. The story-only setup is indicated with a subscript "story." To see these results on common set of claim pairs, that is claim pairs fully processed by all models, see \autoref{table:stories_common_set}. Additionally, we report the performance by story location, either the beginning, middle or the end of the collection, relative to the performance when prompted with that story only (see \autoref{app_tab:stories_by_depth}).\footnote{Note that we do not provide an in-depth analysis and discussion of accuracy relative to the story's location due to the small number of annotations at different depths.}

\begin{table*}[t]
\centering
\footnotesize
\begin{tabular}{lccc}
\hline
\textsc{Model} & \textsc{ACC (\textbf{\color{violet}Pair})}\textsubscript{(correct/total)} & \textsc{ACC (\textbf{\color{teal}True})}\textsubscript{(correct/total)} & \textsc{ACC (\textbf{\color{purple}False})}\textsubscript{(correct/total)} \\
\hline
\gpto & 56.9 \textsubscript{(29/51)} & 82.4 \textsubscript{(42/51)} & 72.5 \textsubscript{(37/51)} \\
\gptturbo & 39.2 \textsubscript{(20/51)} & 58.8 \textsubscript{(30/51)} & 76.5 \textsubscript{(39/51)} \\
\claude & 37.7 \textsubscript{(26/69)} & 75.4 \textsubscript{(52/69)} & 62.3 \textsubscript{(43/69)} \\
\sonnet & 39.1 \textsubscript{(27/69)} & 52.2 \textsubscript{(36/69)} & 84.1 \textsubscript{(58/69)} \\
\geminipro & 52.3 \textsubscript{(23/44)} & 68.2 \textsubscript{(30/44)} & 84.1 \textsubscript{(37/44)} \\
\geminiflash & 40.9 \textsubscript{(18/44)} & 61.4 \textsubscript{(27/44)} & 77.3 \textsubscript{(34/44)} \\
\midrule
\gpto{}\textsubscript{story} & 65.2 \textsubscript{(45/69)} & 76.8 \textsubscript{(53/69)} & 88.4 \textsubscript{(61/69)} \\
\gptturbo{}\textsubscript{story} & 50.7 \textsubscript{(35/69)} & 59.4 \textsubscript{(41/69)} & 87.0 \textsubscript{(60/69)} \\
\claude{}\textsubscript{story} & 63.8 \textsubscript{(44/69)} & 79.7 \textsubscript{(55/69)} & 81.2 \textsubscript{(56/69)} \\
\sonnet{}\textsubscript{story} & 50.7 \textsubscript{(35/69)} & 62.3 \textsubscript{(43/69)} & 88.4 \textsubscript{(61/69)} \\
\geminipro{}\textsubscript{story} & 51.5 \textsubscript{(35/68)} & 58.8 \textsubscript{(40/68)} & 77.9 \textsubscript{(53/68)} \\
\geminiflash{}\textsubscript{story} & 42.0 \textsubscript{(29/69)} & 50.7 \textsubscript{(35/69)} & 75.4 \textsubscript{(52/69)} \\
\midrule
\comrplus & 11.8 \textsubscript{(6/51)} & 68.6 \textsubscript{(35/51)} & 31.4 \textsubscript{(16/51)} \\
\comrplus{}\textsubscript{simple} & 19.6 \textsubscript{(10/51)} & 74.5 \textsubscript{(38/51)} & 39.2 \textsubscript{(20/51)} \\
\comr & 23.5 \textsubscript{(12/51)} & 80.4 \textsubscript{(41/51)} & 39.2 \textsubscript{(20/51)} \\
\comr{}\textsubscript{simple} & 31.4 \textsubscript{(16/51)} & 82.4 \textsubscript{(42/51)} & 45.1 \textsubscript{(23/51)} \\
\phimodel & 7.9 \textsubscript{(3/38)} & 45.0 \textsubscript{(18/40)} & 35.9 \textsubscript{(14/39)} \\
\phimodel{}\textsubscript{simple} & 9.8 \textsubscript{(5/51)} & 80.4 \textsubscript{(41/51)} & 29.4 \textsubscript{(15/51)} \\
\midrule
\comr{}\textsubscript{story} & 39.1 \textsubscript{(27/69)} & 81.2 \textsubscript{(56/69)} & 53.6 \textsubscript{(37/69)} \\
\comrplus{}\textsubscript{story} & 43.5 \textsubscript{(30/69)} & 81.2 \textsubscript{(56/69)} & 59.4 \textsubscript{(41/69)} \\
\phimodel{}\textsubscript{story} & 15.4 \textsubscript{(10/65)} & 26.9 \textsubscript{(18/67)} & 83.3 \textsubscript{(55/66)} \\
\phimodel{}\textsubscript{simple-story} & 24.2 \textsubscript{(16/66)} & 62.1 \textsubscript{(41/66)} & 56.7 \textsubscript{(38/67)} \\
\midrule
\mixtral{}\textsubscript{story} & 56.6 \textsubscript{(39/69)} & 75.4 \textsubscript{(52/69)} & 81.2 \textsubscript{(56/69)} \\
\qwen{}\textsubscript{story} & 60.9 \textsubscript{(42/69)} & 79.7 \textsubscript{(55/69)} & 81.2 \textsubscript{(56/69)} \\
\midrule
\faDatabase\ \bm{} (\textit{k=5})  & 37.7 \textsubscript{(26/69)} & 44.9 \textsubscript{(31/69)} & 91.3 \textsubscript{(63/69)} \\
\faDatabase\ \bm{} (\textit{k=25})  & 53.6 \textsubscript{(37/69)} & 62.3 \textsubscript{(43/69)} & 85.5 \textsubscript{(59/69)} \\
\faDatabase\ \bm{} (\textit{k=50})  & 56.5 \textsubscript{(39/69)} & 71.0 \textsubscript{(49/69)} & 82.6 \textsubscript{(57/69)} \\
\midrule
\textsc{Random} & 25.0 \textsubscript{(17/69)} & 50.0 \textsubscript{(35/69)} & 50.0 \textsubscript{(35/69)} \\
\hline
\end{tabular}
\caption{\textbf{\color{violet}Pairwise} accuracy, accuracy on \textbf{\color{teal}True}, and accuracy on \textbf{\color{purple}False} for claims made about collections of stories (80k-129k tokens) versus individual stories from the collection (700-21k tokens, average 8.5k). All texts fitting within the model's context window were listed. For the 128k token models, the accuracy for the entire collection excludes one book with 129k tokens, which is included in the accuracy for individual stories, as none of the stories exceeded the context window. In the case of \geminipro\ and \geminiflash, two collections were not processed due to being identified as copyrighted content ("prohibited content" API error). However, both models processed the stories from these collections, likely due to the length difference, though some claims were still refused due to disruptive content. We also provide accuracy on stories for \mixtral{} and \qwen{}, which have context window of 65k and 32k respectively, for comparison. Subscript "story" denotes results obtained with prompting the models with individual stories as the context.}
\label{app_tab:acc_on_stories_and_collections_with_true_false}
\end{table*}
\begin{table*}[h!]
\centering
\begin{tabular}{>{\arraybackslash}m{4cm}>{\centering\arraybackslash}m{3cm}>{\centering\arraybackslash}m{3cm}>{\centering\arraybackslash}m{3cm}}
\toprule
\textsc{Model} & \textsc{ACC \textbf{\color{violet}Pair}}\textsubscript{(correct/total)} & \textsc{ACC \textbf{\color{teal}True}}\textsubscript{(correct/total)} & \textsc{ACC \textbf{\color{purple}False}}\textsubscript{(correct/total)} \\
\midrule
\gpto & 48.0 \textsubscript{(12/25)} & 76.0\textsubscript{(19/25)} & 68.0\textsubscript{17/25} \\
\gptturbo & 44.0 \textsubscript{(11/25)} & 64.0\textsubscript{16/25} & 76.0\textsubscript{19/25} \\
\claude & 32.0 \textsubscript{(8/25)} & 88.0\textsubscript{22/25} & 44.0\textsubscript{11/25} \\
\sonnet & 52.0\textsubscript{13/25} & 68.0\textsubscript{17/25} & 80.0\textsubscript{20/25} \\
\geminipro & 56.0\textsubscript{14/25} & 80.0\textsubscript{20/25} & 76.0\textsubscript{19/25} \\
\geminiflash & 48.0\textsubscript{12/25} & 76.0\textsubscript{19/25} & 68.0\textsubscript{17/25} \\
\midrule
\gpto\textsubscript{story} & 72.0\textsubscript{18/25} & 84.0\textsubscript{21/25} & 88.0\textsubscript{22/25} \\
\gptturbo\textsubscript{story}& 56.0\textsubscript{14/25} & 68.0\textsubscript{17/25} & 84.0\textsubscript{21/25} \\
\claude\textsubscript{story} & 72.0\textsubscript{18/25} & 84.0\textsubscript{21/25} & 84.0\textsubscript{21/25} \\
\sonnet\textsubscript{story} & 60.0\textsubscript{15/25} & 68.0\textsubscript{17/25} & 92.0\textsubscript{23/25} \\
\geminipro\textsubscript{story} & 64.0\textsubscript{16/25} & 72.0\textsubscript{18/25} & 88.0\textsubscript{22/25} \\
\geminiflash\textsubscript{story} & 48.0\textsubscript{12/25} & 68.0\textsubscript{17/25} & 76.0\textsubscript{19/25} \\
\midrule
\mixtral\textsubscript{story} & 64.0\textsubscript{16/25} & 80.0\textsubscript{20/25} & 84.0\textsubscript{21/25} \\
\qwen\textsubscript{story} & 64.0\textsubscript{16/25} & 80.0\textsubscript{20/25} & 84.0\textsubscript{21/25} \\
\midrule
\faDatabase\ \bm{} (\textit{k=5}) & 52.0\textsubscript{13/25} & 64.0\textsubscript{16/25} & 88.0\textsubscript{22/25} \\
\faDatabase\ \bm{} (\textit{k=5}) & 68.0\textsubscript{17/25} & 76.0\textsubscript{19/25} & 92.0\textsubscript{23/25} \\
\faDatabase\ \bm{} (\textit{k=5}) & 72.0\textsubscript{18/25} & 88.0\textsubscript{22/25} & 84.0\textsubscript{21/25} \\
\bottomrule
\end{tabular}
\caption{\textbf{\color{violet}Pairwise} accuracy, accuracy on \textbf{\color{teal}True}, and accuracy on \textbf{\color{purple}False} for claims made about collections of stories (80k-129k tokens) versus individual stories from the collection (700-21k tokens, average 8.5k) on \textbf{common set} of claim pairs (i.e., pairs which were processed by all the models in all shown configurations). We also provide accuracies for \mixtral{} and \qwen{} on the same set of claims. These models were prompted with individual stories as the context which makes their results comparable with other models marked with subscript "story."}
\label{table:stories_common_set}
\end{table*}


\begin{table*}[t]
\centering
\begin{tabularx}{\textwidth}{l *{3}{>{\centering\arraybackslash}X}} 
\toprule
&   \multicolumn{3}{c}{\textsc{\faMapMarker\ Story Location within Collection}} \\
\cmidrule(r){2-4}
\textsc{Model} &\faBattery[0]\ \textsc{Beginning} & \faBattery[2]\ \textsc{Middle} & \faBattery[4]\ \textsc{End} \\
\midrule
\faBook\ \gpto & 64.0$_{(16/25)}$ & 50.0$_{(7/14)}$ & 50.0$_{(6/12)}$ \\
\faFileTextO\ \gpto\textsubscript{\textbf{\color{purple}story}} & 76.0$_{(19/25)}$ & 64.3$_{(9/14)}$ & 66.7$_{(8/12)}$ \\
\midrule
\faBook\ \gptturbo & 36.0$_{(9/25)}$ & 35.7$_{(5/14)}$ & 50.0$_{(6/12)}$ \\
\faFileTextO\ \gptturbo\textsubscript{\textbf{\color{purple}story}} & 60.0$_{(15/25)}$ & 42.9$_{(6/14)}$ & 58.3$_{(7/12)}$ \\
\midrule
\faBook\  \claude & 38.2$_{(13/34)}$ & 31.6$_{(6/19)}$ & 43.8$_{(7/16)}$ \\
\faFileTextO\  \claude\textsubscript{\textbf{\color{purple}story}} & 73.5$_{(25/34)}$ & 68.4$_{(13/19)}$ & 37.5$_{(6/16)}$ \\
\midrule
\faBook\ \sonnet & 38.2$_{(13/34)}$ & 31.6$_{(6/19)}$ & 50.0$_{(8/16)}$ \\
\faFileTextO\  \sonnet\textsubscript{\textbf{\color{purple}story}} & 61.8$_{(21/34)}$ & 47.4$_{(9/19)}$ & 31.2$_{(5/16)}$ \\
\midrule
\faBook\ \geminipro & 52.2$_{(12/23)}$ & 70.0$_{(7/10)}$ & 40.0$_{(4/10)}$ \\
\faFileTextO\   \geminipro\textsubscript{\textbf{\color{purple}story}} & 56.5$_{(13/23)}$ & 70.0$_{(7/10)}$ & 40.0$_{(4/10)}$ \\
\midrule
\faBook\ \geminiflash & 39.1$_{(9/23)}$ & 27.3$_{(3/11)}$ & 60.0$_{(6/10)}$ \\
\faFileTextO\   \geminiflash\textsubscript{\textbf{\color{purple}story}} & 34.8$_{(8/23)}$ & 63.6$_{(7/11)}$ & 30.0$_{(3/10)}$ \\
\midrule
\faBook\ \comr & 20.0$_{(5/25)}$ & 21.4$_{(3/14)}$ & 33.3$_{(4/12)}$ \\
\faFileTextO\  \comr\textsubscript{\textbf{\color{purple}story}} & 36.0$_{(9/25)}$ & 35.7$_{(5/14)}$ & 33.3$_{(4/12)}$ \\
\midrule
\faBook\ \comrplus & 20.0$_{(5/25)}$ & 0.0$_{(0/14)}$ & 8.3$_{(1/12)}$ \\
\faFileTextO\  \comrplus\textsubscript{\textbf{\color{purple}story}} & 48.0$_{(12/25)}$ & 50.0$_{(7/14)}$ & 41.7$_{(5/12)}$ \\
\midrule
\faBook\ \phimodel & 6.2$_{(1/16)}$ & 8.3$_{(1/12)}$ & 14.3$_{(1/7)}$ \\
\faFileTextO\  \phimodel\textsubscript{\textbf{\color{purple}story}} & 18.8$_{(3/16)}$ & 33.3$_{(4/12)}$ & 0.0$_{(0/7)}$ \\
\midrule
\faBook\ \phimodel\textsubscript{simple} & 12.5$_{(3/24)}$ & 15.4$_{(2/13)}$ & 0.0$_{(0/12)}$ \\
\faFileTextO\  \phimodel\textsubscript{simple-\textbf{\color{purple}story}}& 37.5$_{(9/24)}$ & 30.8$_{(4/13)}$ & 25.0$_{(3/12)}$ \\
\midrule
\faFileTextO\  \mixtral\textsubscript{\textbf{\color{purple}story}}& 61.8$_{(21/34)}$ & 52.6$_{(10/19)}$ & 50.0$_{(8/16)}$ \\
\faFileTextO\  \qwen\textsubscript{\textbf{\color{purple}story}}& 61.8$_{(21/34)}$ & 63.2$_{(12/19)}$ & 56.2$_{(9/16)}$ \\
\bottomrule
\end{tabularx}
\caption{Models' performance on stories at different depths of the story collection. Each collection is divided into three parts based on the number of tokens: (1) beginning (first third), (2) middle (second third), and (3) end (last third). The subscript "story" (marked with \faFileTextO{}) refers to outputs obtained by prompting with the story (rather than the collection) as the context, for comparison. We also provide results for \mixtral{} and \qwen{} for comparison.}
\label{app_tab:stories_by_depth}
\end{table*}

\paragraph{\claude{} cites evidence more often than other models:} We employ a simple heuristic by looking for quotation marks in the evidence to identify how often models cite excerpts from the source. Qualitatively, we observed that \claude{} cites the source more often than other closed-source models, and the citations we verified were always present in the source. This quantitative analysis confirms our observation, as 49.6\% of \claude{}'s responses contain quotation marks, followed by 18.6\% for both \gpto{} and \gptturbo{}, 11.5\% for \geminipro{}, and 10.8\% for \geminiflash{}. We also examined whether responses containing citations are more likely to be correct but did not find such a relation (see \autoref{tab:quotations_in_ans}).

\begin{table*}[h]
\centering
\footnotesize
\resizebox{0.9\textwidth}{!}{%
\begin{tabular}{lccc}
\toprule
 & \multicolumn{3}{c}{\textsc{Quotations (\%)}} \\
\cmidrule(lr){2-4}
 \textsc{Model}   & \textsc{All responses} & \textsc{Correct responses} & \textsc{Incorrect responses} \\
\midrule
\gpto{}       & 18.4\% \textsubscript{227/1234}     & 18.8\% \textsubscript{178/946}          & 17.0\% \textsubscript{49/288} \\

\gptturbo{}    & 18.8\% \textsubscript{232/1234}      & 19.5\% \textsubscript{164/839}          & 17.2\% \textsubscript{68/395} \\

\claude{}      & 49.8\% \textsubscript{934/1874}      & 49.8\% \textsubscript{685/1376}         & 50.0\% \textsubscript{249/498} \\

\sonnet{}      & 59.6\% \textsubscript{1116/1874}      & 62.2\% \textsubscript{808/1300}         & 53.7\% \textsubscript{308/574} \\

\geminipro{}   & 11.6\% \textsubscript{119/1029}      & 11.4\% \textsubscript{84/738}           & 12.0\% \textsubscript{35/291} \\

\geminiflash{} & 10.8\% \textsubscript{111/1030}      & 10.5\% \textsubscript{70/668}           & 11.3\% \textsubscript{41/362} \\
\bottomrule
\end{tabular}
}
\caption{Percentage of responses with identified quotations by model. Separate values are reported for claims labeled correctly and incorrectly. Note that the percentages are reported by claim, as each explanation is generated at the claim level. The counts (quotations/total) are provided in subscript. We do not report these numbers for open-weights models as the generations often do not follow the requested output format.}
\label{tab:quotations_in_ans}
\end{table*}

\paragraph{Effect of genre:} We classify the books into three genres: historical, contemporary, and speculative. \autoref{app_tab:genre_all_models} presents results by genre for all models.

\begin{table*}[t]
\centering
\begin{tabular}{lccc}
\toprule
 & \multicolumn{3}{c}{\textsc{\faBook\ Genre}} \\
\cmidrule(lr){2-4}
\textsc{Model} & \textsc{Historical} & \textsc{Contemporary} & \textsc{Speculative} \\
\midrule
\gpto & 70.3\textsubscript{(26/37)} & 59.0\textsubscript{(229/388)} & 44.2\textsubscript{(72/163)} \\
\gptturbo & 70.3\textsubscript{(26/37)} & 39.9\textsubscript{(155/388)} & 34.4\textsubscript{(56/163)} \\
\claude & 63.5\textsubscript{(33/52)} & 51.1\textsubscript{(290/567)} & 42.9\textsubscript{(124/289)} \\
\sonnet & 42.3\textsubscript{(22/52)} & 42.2\textsubscript{(239/567)} & 38.1\textsubscript{(110/289)} \\
\geminipro & 53.8\textsubscript{(28/52)} & 49.2\textsubscript{(127/258)} & 44.4\textsubscript{(84/189)} \\
\geminiflash & 46.2\textsubscript{(24/52)} & 37.2\textsubscript{(96/258)} & 27.4\textsubscript{(52/190)} \\
\midrule
\comr & 30.3\textsubscript{(10/33)} & 17.8\textsubscript{(46/259)} & 20.1\textsubscript{(28/139)} \\
\comr\textsubscript{simple} & 21.2\textsubscript{(7/33)} & 22.8\textsubscript{(59/259)} & 21.6\textsubscript{(30/139)} \\
\comrplus & 18.2\textsubscript{(6/33)} & 21.6\textsubscript{(56/259)} & 10.1\textsubscript{(14/139)} \\
\comrplus\textsubscript{simple} & 15.2\textsubscript{(5/33)} & 13.1\textsubscript{(34/259)} & 15.1\textsubscript{(21/139)} \\
\phimodel & 10.0\textsubscript{(1/10)} & 9.5\textsubscript{(14/147)} & 9.0\textsubscript{(8/89)} \\
\phimodel\textsubscript{simple} & 20.0\textsubscript{(3/15)} & 15.8\textsubscript{(29/183)} & 12.6\textsubscript{(15/119)} \\
\gemma & 1.5\textsubscript{(1/67)} & 3.2\textsubscript{(18/567)} & 4.4\textsubscript{(15/338)} \\
\gemma\textsubscript{simple} & 4.5\textsubscript{(3/67)} & 7.2\textsubscript{(41/567)} & 7.7\textsubscript{(26/338)} \\
\longllama\textsubscript{simple} & 0.0\textsubscript{(0/52)} & 5.1\textsubscript{(29/567)} & 4.2\textsubscript{(12/289)} \\
\bm{} (\textit{k=5}) & 35.8\textsubscript{(24/67)} & 28.6\textsubscript{(162/567)} & 26.0\textsubscript{(88/338)} \\
\bm{} (\textit{k=25}) & 50.7\textsubscript{(34/67)} & 42.9\textsubscript{(243/567)} & 44.4\textsubscript{(150/338)} \\
\bm{} (\textit{k=50})& 56.7\textsubscript{(38/67)} & 50.8\textsubscript{(288/567)} & 45.9\textsubscript{(155/338)} \\
\bottomrule
\end{tabular}
\caption{Model performance by genre: \textit{historical} (pre-WWII), \textit{contemporary} (post-WWII), and \textit{speculative} (fantasy/SF/ghosts).}
\label{app_tab:genre_all_models}
\end{table*}

\begin{figure*}[tbp]
  \includegraphics[width=1\linewidth]{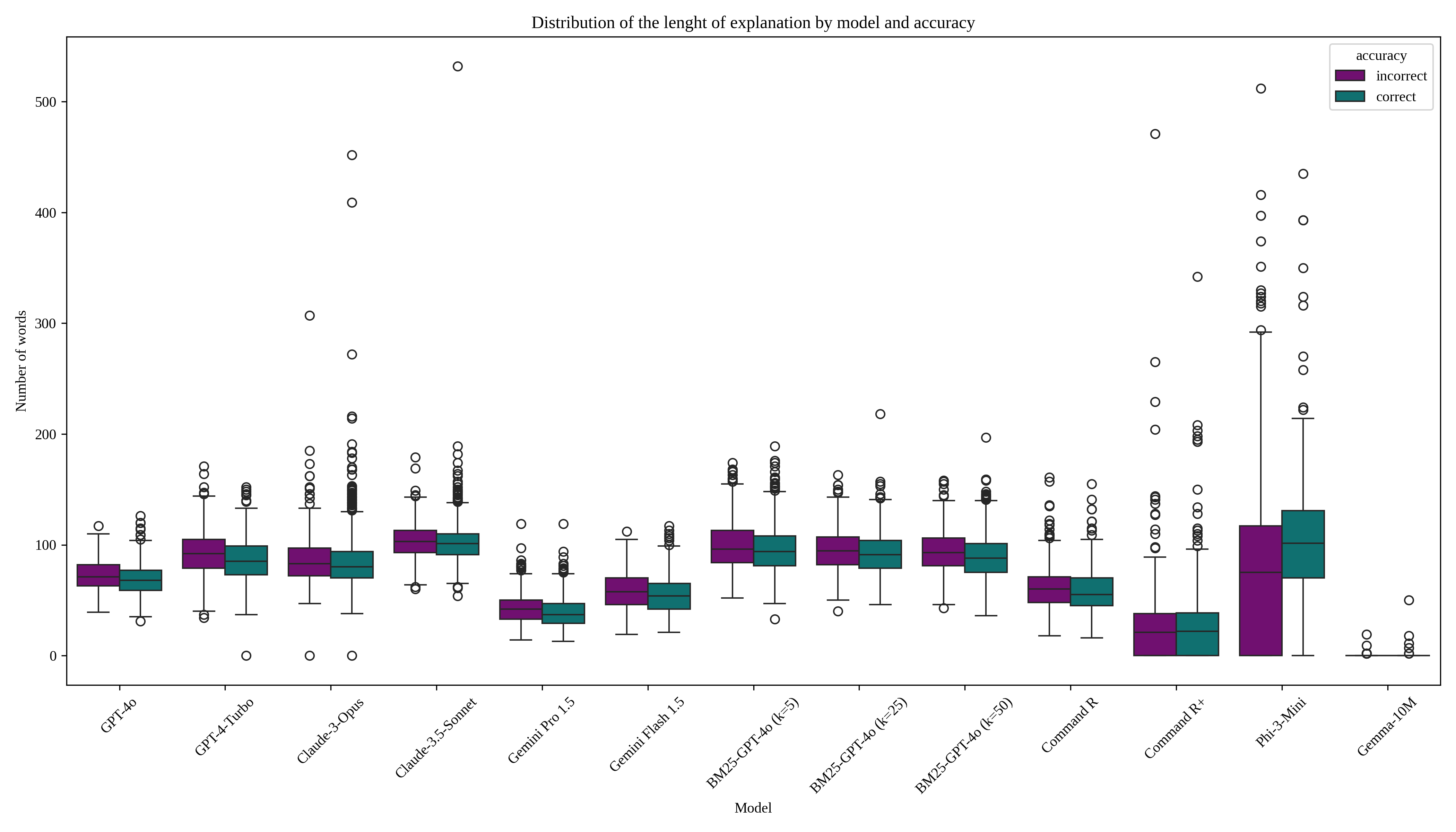} 
  \caption{Boxplots of the length of justification (in words) provided for each model for correct (green) and incorrect (purple) predictions.}
  \label{fig:boxplot_len}
\end{figure*}

\paragraph{Length of the justifications:} \autoref{fig:boxplot_len} provides lengths of the justification provided by each model for the main prompt template (\autoref{tab:prompt_template_claim_eval}). We report the lengths in  words by accuracy for each model. Note that \gemma{} never produces real explanation, rather sometimes repeats part of the original prompt, i.e., \texttt{<explanation>YOUR ANSWER</explanation>}.

\begin{figure*}[h!]
  \includegraphics[width=1\linewidth]{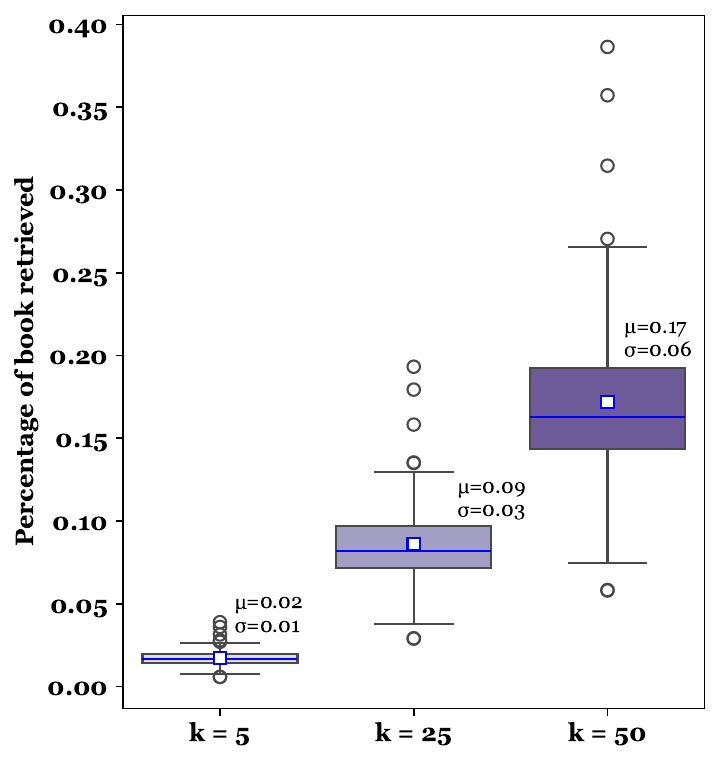} 
  \caption{Average percentage of the book retrieved by BM25 for varying values of top k. For example, k = 5 means that the top 5 ranked excerpts according to BM25 were fed to GPT-4o as context for a claim.}
  \label{fig:percentage_retrieved_bm25}
\end{figure*}

\paragraph{Length of the retrieved chunks:} \autoref{fig:percentage_retrieved_bm25} shows the average percentage of the book retrieved by \textsc{\small BM25} for varying values of \textit{k}, i.e., for top 5, 25, and 50 chunks. 

\begin{figure*}[h!]
  \includegraphics[width=1\linewidth]{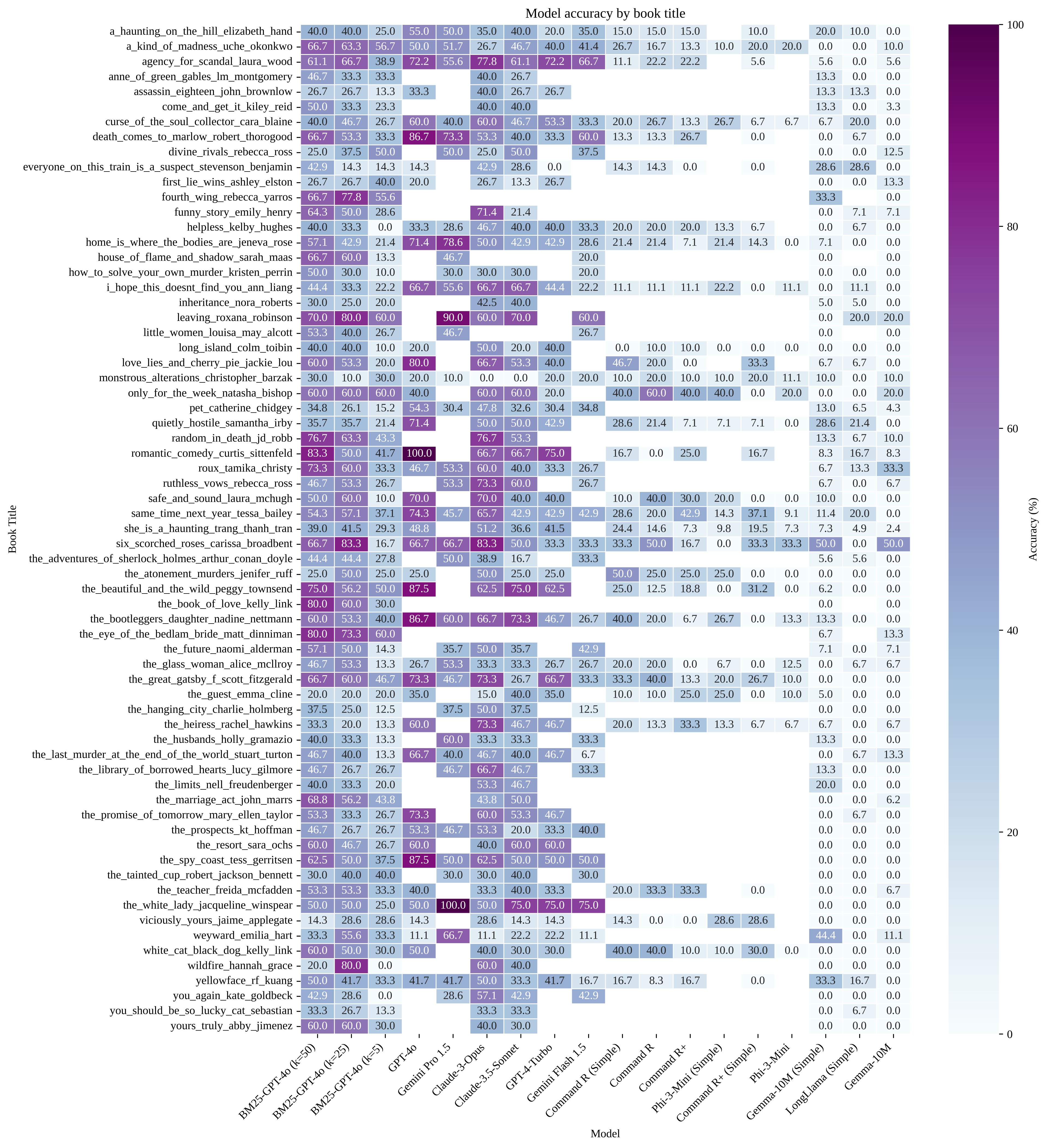} 
  \caption{Heatmap of model performance by book title. Empty cells indicate cases where the model did not process the book (\geminipro{} and \geminiflash{}) or could not process the book because of its context window.}
  \label{fig:model_by_book_performance}
\end{figure*}

\paragraph{Performance by book:} We report the performance of each model by the book title in \autoref{fig:model_by_book_performance}. Empty cells indicate cases where the model did not process the book \geminipro{} and \geminiflash{}) or could not process the book because of its context window. See \autoref{tab:list_of_novels} for the number of claim pairs written for each book.

\paragraph{Statistical analysis:} We conducted statistical analysis by fitting generalized linear mixed-effects models using the \texttt{glmer()} function in \texttt{R} \cite{glmer2015}.\footnote{All models were fitted using the closed-source models data only, as all open-weights models performed below random.} The response variable was pairwise \textbf{accuracy}, a binary categorical variable ("correct" or "incorrect"). Pair IDs and annotators were modeled as random effects, with various predictors (fixed effects) included in different models:

\begin{enumerate}
\item \textbf{Model} - a 6-level categorical variable (\gpto{}, \gptturbo{}, \claude{}, \sonnet{}, \geminipro{}, \geminiflash{}), with analysis restricted to pairs processed by all models. \gpto{} was set as the reference level (intercept). See \autoref{app_tab:stats_accuracy_model_glmer} and \autoref{app_tab:stats_accuracy_model_glmer_posthoc} for the results.;
\item \textbf{Length group} - a 4-level categorical variable ("below 75k", "75k-127k", "127k-180k", "above 180k"). "Below 75k" category was set as the reference level (intercept). See \autoref{app_tab:stats_accuracy_length_group_glmer} and \autoref{app_tab:stats_accuracy_length_group_glmer_posthoc} for the results;
\item \textbf{Year} - a 3-level categorical variable ("classics", "2023", "2024") with "classics" set as the reference level. See \autoref{app_tab:stats_accuracy_year_glmer} for the results;
\item \textbf{Genre} - a 3-level categorical variable ("historical", "contemporary", "speculative") with "speculative" set as the reference level. See \autoref{app_tab:stats_accuracy_genre_glmer} and \autoref{app_tab:stats_accuracy_genre_glmer_posthoc} for the results;
\item \textbf{Scope} - a 3-level categorical variable ("sentence", "passage", "global") with "sentence" set as the reference level. See \autoref{app_tab:stats_accuracy_scope_glmer} and \autoref{app_tab:stats_accuracy_scope_glmer_posthoc} for the results.
\end{enumerate}

All models were fitted using the \texttt{bobyqa} optimizer with a \texttt{binomial} link function. We chose mixed-effects models for two main reasons: (1) to account for repeated measures, as each model validates multiple pairs and each pair is validated by multiple models, and (2) to flexibly model pair IDs and annotators as random effects, partially controlling for the inherent difficulty of the pairs unrelated to the predictors.

For these mixed-effects models, we report two types of R\textsuperscript{2} values:
\begin{itemize}
\item \textit{Marginal} R\textsuperscript{2}, which indicates the proportion of variance explained by the fixed effects (predictors) alone.
\item \textit{Conditional} R\textsuperscript{2}, which represents the proportion of variance explained by both the fixed and random effects \cite{Nakagawa2017}.
\end{itemize}

We further conducted a post-hoc analysis using the \texttt{emmeans} package \cite{emmeans2023} in R with Tukey adjustments for multiple comparisons. To obtain probabilities, we first converted log-odds to odds ratios by exponentiating the estimates, and then converted the odds ratios to probabilities which are reported in the post-hoc tables.

\begin{table*}[t]
\centering
\begin{tabular}{lcccl}
\toprule
\multicolumn{5}{c}{\texttt{model <- glmer(accuracy $\sim$ model + (1|id) + (1|annotator), common\_set\_data)}} \\
 \midrule
\textsc{Predictors} & \textsc{Odds Ratios} & \textsc{CI (95\%)} & \textsc{\textit{p}-value} & \\
\midrule
(Intercept) & 1.73 & 1.18 – 2.55 & 0.005 & ** \\
\gptturbo{} & 0.33 & 0.23 – 0.48 & <0.001 & *** \\
\claude{} & 0.64 & 0.45 – 0.91 & 0.014 & * \\
\sonnet{} & 0.34 & 0.24 – 0.50 & <0.001 & *** \\
\geminipro{} & 0.55 & 0.38 – 0.78 & 0.001 & *** \\
\geminiflash{} & 0.24 & 0.17 – 0.35 & <0.001 & *** \\
\midrule
\textsc{Random Effects} & & \\
\midrule
$\sigma^2$ (\textit{residual variance}) & 3.29 & & \\
$\tau_{00}$ (id) (\textit{variance of random intercepts}) & 2.54 & & \\
$\tau_{00}$ (annotator) (\textit{variance of random intercepts}) & 0.19 & & \\
\textsc{ICC} & 0.45 & & \\
\textsc{N (id)} & 354 & & \\
\textsc{N (annotator)} & 15 & & \\
\textsc{Observations} & 2124 & & \\
\midrule
\textsc{R\textsuperscript{2} (\textit{marginal})} & 0.036 & & \\
\textsc{R\textsuperscript{2} (\textit{conditional})} & 0.473 & & \\
\bottomrule
\end{tabular}
\caption{Summary of generalized linear mixed model with \textbf{model} as the predictor of \textbf{accuracy}: \texttt{model <- glmer(accuracy $\sim$ model + (1|id) + (1|annotator), data)}. \gpto{} was set as the reference level (intercept). Note that while we observe significant differences between the models' performance, the marginal R\textsuperscript{2} is low, suggesting that model type alone does not explain the majority of the variance in the data. See \autoref{app_tab:stats_accuracy_model_glmer_posthoc} for post-hoc analysis.}
\label{app_tab:stats_accuracy_model_glmer}
\end{table*}

\begin{table*}[t]
\centering
\resizebox{0.9\linewidth}{!}{%
\begin{tabular}{lcccccl}
\toprule
\textsc{Contrast} & \textsc{Estimate} & \textsc{SE} & \textsc{Odds Ratio} & \textsc{Probability} & \textsc{\textit{p}-value} & \\
\midrule
\gpto{} - \claude{}              & 0.4480 & 0.1820 & 1.5652 & 0.6102 & 0.1355 & \\
\gpto{} - \sonnet{}              & 1.0654 & 0.1849 & 2.9021 & 0.7437 & <0.001 & ***\\
\gpto{} - \geminiflash{}         & 1.4229 & 0.1888 & 4.1492 & 0.8058 & <0.001 & *** \\
\gpto{} - \geminipro{}           & 0.6013 & 0.1822 & 1.8244 & 0.6459 & 0.0124 & *\\
\gpto{} - \gptturbo{}            & 1.1005 & 0.1852 & 3.0056 & 0.7503 & <0.001 & ***\\
\claude{}  - \claude{}           & 0.6174 & 0.1820 & 1.8541 & 0.6496 & 0.0090 & **\\
\claude{}  - \geminiflash{}      & 0.9749 & 0.1855 & 2.6509 & 0.7261 & <0.001 & ***\\
\claude{}  - \geminipro{}        & 0.1533 & 0.1801 & 1.1656 & 0.5382 & 0.9578 &\\
\claude{}  - \gptturbo{}         & 0.6524 & 0.1822 & 1.9202 & 0.6576 & 0.0046 & ** \\
\sonnet{} - \geminiflash{}       & 0.3575 & 0.1852 & 1.4297 & 0.5884 & 0.3833 & \\
\sonnet{} - \geminipro{}         & -0.4642 & 0.1816 & 0.6287 & 0.3860 & 0.1084 & \\
\sonnet{} - \gptturbo{}          & 0.0350 & 0.1826 & 1.0356 & 0.5088 & 1.0000 & \\
\geminiflash{} - \geminipro{}    & -0.8216 & 0.1849 & 0.4397 & 0.3054 & <0.001 & *** \\
\geminiflash{} - \gptturbo{}     & -0.3224 & 0.1853 & 0.7244 & 0.4201 & 0.5050 & \\
\geminipro{} - \gptturbo{}       & 0.4992 & 0.1819 & 1.6474 & 0.6223 & 0.0667 & .\\
\bottomrule
\end{tabular}
}
\caption{Post-hoc comparisons of models for \textbf{accuracy} (\autoref{app_tab:stats_accuracy_model_glmer}) using Tukey adjustments for multiple comparisons. The probability values refer to the likelihood that the first model in each contrast is more accurate than the second model (i.e., a value of 0.5 suggests that both models are comparable in terms of accuracy).}
\label{app_tab:stats_accuracy_model_glmer_posthoc}
\end{table*}

\begin{table*}[t]
\centering
\begin{tabular}{lcccl}
\toprule
\multicolumn{5}{c}{\texttt{model <- glmer(accuracy $\sim$ length\_group + (1|pairID) + (1|annotator), data)}} \\
 \midrule
\textsc{Predictors} & \textsc{Odds Ratios} & \textsc{CI (95\%)} & \textsc{\textit{p}-value} & \\
\midrule
(Intercept) & 0.56 & 0.34 – 0.94 & 0.028 & * \\
\textsc{75k--127k} & 1.45 & 0.87 – 2.43 & 0.154 & \\
\textsc{127k--180k} & 1.26 & 0.74 – 2.14 & 0.391 & \\
\textsc{above 180k} & 0.75 & 0.28 – 1.96 & 0.551 & \\
\midrule
\textsc{Random Effects} & & \\
\midrule
$\sigma^2$ (\textit{residual variance}) & 3.29 & & \\
$\tau_{00}$ (id) (\textit{variance of random intercepts}) & 2.15 & & \\
$\tau_{00}$ (annotator) (\textit{variance of random intercepts}) & 0.24 & & \\
\textsc{ICC} & 0.42 & & \\
\textsc{N (id)} & 967 & & \\
\textsc{N (annotator)} & 22 & & \\
\textsc{Observations} & 4137 & & \\
\midrule
\textsc{R\textsuperscript{2} (\textit{marginal})} & 0.004 & & \\
\textsc{R\textsuperscript{2} (\textit{conditional})} & 0.423 & & \\
\bottomrule
\end{tabular}
\caption{Summary of generalized linear mixed model with \textbf{length group} as the predictor of \textbf{accuracy}: \texttt{model <- glmer(accuracy $\sim$ length\_group + (1|pairID) + (1|annotator), data)}. The "below 75k" group was set as the reference level (intercept). Note that the marginal R\textsuperscript{2} is very low, indicating that length group alone does not explain the majority of the variance in the data. See \autoref{app_tab:stats_accuracy_length_group_glmer_posthoc} for post-hoc analysis.}
\label{app_tab:stats_accuracy_length_group_glmer}
\end{table*}


\begin{table*}[t]
\centering
\resizebox{0.9\linewidth}{!}{%
\begin{tabular}{lcccccl}
\toprule
\textsc{Contrast} & \textsc{Estimate} & \textsc{SE} & \textsc{Odds Ratio} & \textsc{Probability} & \textsc{\textit{p}-value} & \\
\midrule
\textsc{Below 75k - 127k-180k} & -0.2312 & 0.2698 & 0.7936 &  0.4424 & 0.8267 & \\
\textsc{Below 75k - 75k-127k} & -0.3739 & 0.2620 & 0.6880 & 0.4076 & 0.4823 & \\
\textsc{Below 75k - Above 180k} & 0.2938 & 0.4925 & 1.3416 & 0.5729 & 0.9331 & \\
\textsc{127k-180k - 75k-127k} & -0.1427 & 0.1652 & 0.8670 & 0.4644 & 0.8235 & \\
\textsc{127k-180k - Above 180k} & 0.5251 & 0.4461 & 1.6906 & 0.6283 & 0.6415 & \\
\textsc{75k-127k - Above 180k} & 0.6677 & 0.4464 & 1.9498 & 0.6610 & 0.4399 & \\
\bottomrule
\end{tabular}
}
\caption{Post-hoc comparisons of length groups for \textbf{accuracy} (\autoref{app_tab:stats_accuracy_length_group_glmer}) using Tukey adjustments for multiple comparisons. The probability values refer to the likelihood that the verification for the first group in each contrast is more accurate than the verification for the second group.}
\label{app_tab:stats_accuracy_length_group_glmer_posthoc}
\end{table*}


\begin{table*}[t]
\centering
\begin{tabular}{lcccl}
\toprule
\multicolumn{5}{c}{\texttt{model <- glmer(accuracy $\sim$ year + (1|pairID) + (1|annotator), data)}} \\
 \midrule
\textsc{Predictors} & \textsc{Odds Ratios} & \textsc{CI (95\%)} & \textsc{\textit{p}-value} & \\
\midrule
(Intercept) & 0.65 & 0.35 – 1.20 & 0.170 &  \\
\textsc{2023} & 1.17 & 0.64 – 2.15 & 0.321 & \\
\textsc{2024} & 1.17 & 0.63 – 2.16 & 0.344 & \\
\midrule
\textsc{Random Effects} & & \\
\midrule
$\sigma^2$ (\textit{residual variance}) & 3.29 & & \\
$\tau_{00}$ (id) (\textit{variance of random intercepts}) & 2.16 & & \\
$\tau_{00}$ (annotator) (\textit{variance of random intercepts}) & 0.22 & & \\
\textsc{ICC} & 0.42 & & \\
\textsc{N (id)} & 967 & & \\
\textsc{N (annotator)} & 22 & & \\
\textsc{Observations} & 4137 & & \\
\midrule
\textsc{R\textsuperscript{2} (\textit{marginal})} & 0.000 & & \\
\textsc{R\textsuperscript{2} (\textit{conditional})} & 0.420 & & \\
\bottomrule
\end{tabular}
\caption{Summary of generalized linear mixed model with \textbf{year} as the predictor of \textbf{accuracy}: \texttt{model <- glmer(accuracy $\sim$ year + (1|pairID) + (1|annotator), data)}. The "classics" year group was set as the reference level (intercept). Note that the marginal R\textsuperscript{2} is 0.000, indicating that year alone does not explain the majority of the variance in the data.}
\label{app_tab:stats_accuracy_year_glmer}
\end{table*}

\begin{table*}[t]
\centering
\begin{tabular}{lcccl}
\toprule
\multicolumn{5}{c}{\texttt{model <- glmer(accuracy $\sim$ genre + (1|id) + (1|annotator), data = filtered\_data)}} \\
\midrule
\textsc{Predictors} & \textsc{Odds Ratios} & \textsc{CI (95\%)} & \textsc{\textit{p}-value} & \\
\midrule
(Intercept) & 0.55 & 0.40 – 0.74 & <0.001 & *** \\
\textsc{Contemporary} & 1.53 & 1.11 – 2.11 & 0.010 & *\\
\textsc{Historical} & 2.18 & 1.20 – 3.96 & 0.010 & * \\
\midrule
\textsc{Random Effects} & & & \\
\midrule
$\sigma^2$ (\textit{residual variance}) & 3.29 & & \\
$\tau_{00}$ (id) (\textit{variance of random intercepts}) & 2.10 & & \\
$\tau_{00}$ (annotator) (\textit{variance of random intercepts}) & 0.16 & & \\
\textsc{ICC} & 0.41 & & \\
\textsc{N (id)} & 938 & & \\
\textsc{N (annotator)} & 22 & & \\
\textsc{Observations} & 3991 & & \\
\midrule
\textsc{R\textsuperscript{2} (\textit{marginal})} & 0.010 & & \\
\textsc{R\textsuperscript{2} (\textit{conditional})} & 0.412 & & \\
\bottomrule
\end{tabular}
\caption{Summary of generalized linear mixed model with \textbf{genre} as the predictor of \textbf{accuracy}: \texttt{model <- glmer(accuracy $\sim$ genre + (1|id) + (1|annotator), data = filtered\_data)}. The "speculative" genre was set as the reference level (intercept). We also excluded two books which felt into "historical and contemporary" and "essays" categories. Note that the marginal R\textsuperscript{2} is low, indicating that genre alone does not explain the majority of the variance in the data. The post-hoc analysis for this model is presented in \autoref{app_tab:stats_accuracy_genre_glmer_posthoc}.}
\label{app_tab:stats_accuracy_genre_glmer}
\end{table*}

\begin{table*}[ht]
\centering
\resizebox{0.8\linewidth}{!}{%
\begin{tabular}{lcccccl}
\toprule
\textsc{Contrast} & \textsc{Estimate} & \textsc{SE} & \textsc{Odds Ratio} & \textsc{Probability} & \textsc{\textit{p}-value} & \\
\midrule
\textsc{Speculative} - \textsc{Contemporary} & -0.4234 & 0.1648 & 0.6548 & 0.3957 & 0.0275 & * \\
\textsc{Speculative} - \textsc{Historical} & -0.7802 & 0.3042 & 0.4583 & 0.3143 & 0.0278 & * \\
\textsc{Contemporary} - \textsc{Historical} & -0.3568 & 0.2874 & 0.6999 & 0.4117 & 0.4287 & \\
\bottomrule
\end{tabular}
}
\caption{Post-hoc comparisons of genres for \textbf{accuracy} (\autoref{app_tab:stats_accuracy_genre_glmer}) using Tukey adjustments for multiple comparisons. The probability values refer to the likelihood that the validation of books of the first genre in each contrast is more accurate than the validation of books of the second genre (i.e., a value of 0.5 suggests that both genres are comparable in terms of accuracy).}
\label{app_tab:stats_accuracy_genre_glmer_posthoc}
\end{table*}

\begin{table*}[ht]
\centering
\begin{tabular}{lcccl}
\toprule
\multicolumn{5}{c}{\texttt{model <- glmer(accuracy $\sim$ scope + (1|id) + (1|annotator), data = filtered\_data)}} \\
\midrule
\textsc{Predictors} & \textsc{Odds Ratios} & \textsc{CI (95\%)} & \textsc{\textit{p}-value} & \\
\midrule
(Intercept) & 2.12 & 0.74 – 6.11 & 0.163 & \\
\textsc{Passage} & 0.35 & 0.14 – 0.87 & 0.024 & *\\
\textsc{Global} & 0.32 & 0.13 – 0.78 & 0.012 & * \\
\midrule
\textsc{Random Effects} & & & \\
\midrule
$\sigma^2$ (\textit{residual variance}) & 3.29 & & \\
$\tau_{00}$ (id) (\textit{variance of random intercepts}) & 1.19 & & \\
$\tau_{00}$ (annotator) (\textit{variance of random intercepts}) & 0.53 & & \\
\textsc{ICC} & 0.34 & & \\
\textsc{N (id)} & 121 & & \\
\textsc{N (annotator)} & 4 & & \\
\textsc{Observations} & 544 & & \\
\midrule
\textsc{R\textsuperscript{2} (\textit{marginal})} & 0.030 & & \\
\textsc{R\textsuperscript{2} (\textit{conditional})} & 0.363 & & \\
\bottomrule
\end{tabular}
\caption{Summary of generalized linear mixed model with \textbf{scope} as the predictor of \textbf{accuracy}: \texttt{model <- glmer(accuracy $\sim$ scope + (1|id) + (1|annotator), data = filtered\_data)}. The "sentence" scope was set as the reference level (intercept). Note that the marginal R\textsuperscript{2} is low, indicating that scope alone does not explain the majority of the variance in the data. See \autoref{app_tab:stats_accuracy_scope_glmer_posthoc} for the post-hoc analysis.}
\label{app_tab:stats_accuracy_scope_glmer}
\end{table*}

\begin{table*}[ht]
\centering
\resizebox{0.8\linewidth}{!}{%
\begin{tabular}{lcccccl}
\toprule
\textsc{Contrast} & \textsc{Estimate} & \textsc{SE} & \textsc{Odds Ratio} & \textsc{Probability} & \textsc{\textit{p}-value} & \\
\midrule
\textsc{Sentence} - \textsc{Global} & 1.1322 & 0.4506 & 3.1026 & 0.7563 & 0.0321 & * \\
\textsc{Sentence} - \textsc{Passage} & 1.0488 & 0.4642 & 2.8542 & 0.7405 & 0.0617 & . \\
\textsc{Global}  - \textsc{Passage} & -0.0834 & 0.3139 & 0.9199 & 0.4792 & 0.9618 & \\
\bottomrule
\end{tabular}
}
\caption{Post-hoc comparisons of scopes for \textbf{accuracy} (\autoref{app_tab:stats_accuracy_scope_glmer}) using Tukey adjustments for multiple comparisons. The probability values refer to the likelihood that the validation of claim pairs in the first scope in each contrast is more accurate than the validation of claim pairs in the second scope (i.e., a value of 0.5 suggests that both scopes are comparable in terms of accuracy).}
\label{app_tab:stats_accuracy_scope_glmer_posthoc}
\end{table*}

\end{document}